\documentclass[10pt,twocolumn,letterpaper]{article}

\usepackage{iccv}
\usepackage{times}
\usepackage{epsfig}
\usepackage{graphicx}
\usepackage{amsmath}
\usepackage{amssymb}
\usepackage{mathtools}
\usepackage{multirow}
\usepackage{algorithm}
\usepackage{algorithmic}
\usepackage{bm}
\usepackage{float}
\usepackage{xcolor}
\usepackage{caption}
\usepackage{subcaption}
\usepackage{booktabs}
\DeclarePairedDelimiter\norm{\lVert}{\rVert}

\newcommand{\para}[1]{\vspace{4pt}\noindent\textbf{#1}}

\usepackage[pagebackref=true,breaklinks=true,colorlinks,bookmarks=false]{hyperref}
\iccvfinalcopy 

\def\iccvPaperID{9291} 

\ificcvfinal\fi

\begin{document}

\title{Likelihood-Based Diverse Sampling for Trajectory Forecasting}

\author{%
Yecheng Jason Ma \\
University of Pennsylvania \\ 
{\tt\small jasonyma@seas.upenn.edu} 
\and  
Jeevana Priya Inala\\ 
MIT CSAIL \\ 
{\tt\small jinala@csail.mit.edu} \\ 
\and 
Dinesh Jayaraman, Osbert Bastani  \\ 
University of Pennsylvania \\ 
{\tt\small \{dineshj, obastani\}@seas.upenn.edu}\\
}

\maketitle
\ificcvfinal\thispagestyle{empty}\fi

\begin{abstract}
 Forecasting complex vehicle and pedestrian multi-modal distributions requires powerful probabilistic approaches. Normalizing flows (NF) have recently emerged as an attractive tool to model such distributions. However, a key drawback is that independent samples drawn from a flow model often do not adequately capture all the modes in the underlying distribution. We propose \textbf{L}ikelihood-Based \textbf{D}iverse \textbf{S}ampling (LDS),
a method for improving the quality and the diversity of trajectory samples from a pre-trained flow model. Rather than producing individual samples, LDS produces a set of trajectories in one shot. Given a pre-trained forecasting flow model, we train LDS using gradients from the model, to optimize an objective function that rewards high likelihood for individual trajectories in the predicted set, together with high spatial separation among trajectories. LDS outperforms state-of-art post-hoc neural diverse forecasting methods for various pre-trained flow models as well as conditional variational autoencoder (CVAE) models. Crucially, it can also be used for transductive trajectory forecasting, where the diverse forecasts are trained on-the-fly on unlabeled test examples.
LDS is easy to implement, and we show that it offers a simple plug-in improvement over baselines on two challenging benchmarks. Code is at: \url{https://github.com/JasonMa2016/LDS}
\end{abstract}

\section{Introduction}
A key challenge facing self-driving cars is accurately forecasting the future trajectories of other vehicles. These future trajectories are often diverse and multi-modal, requiring a forecasting model to predict not a single ground truth future but the full range of plausible futures \cite{liang2020garden}. 

\begin{figure}
\centering
\resizebox{\columnwidth}{!}{
\includegraphics[width=\columnwidth]{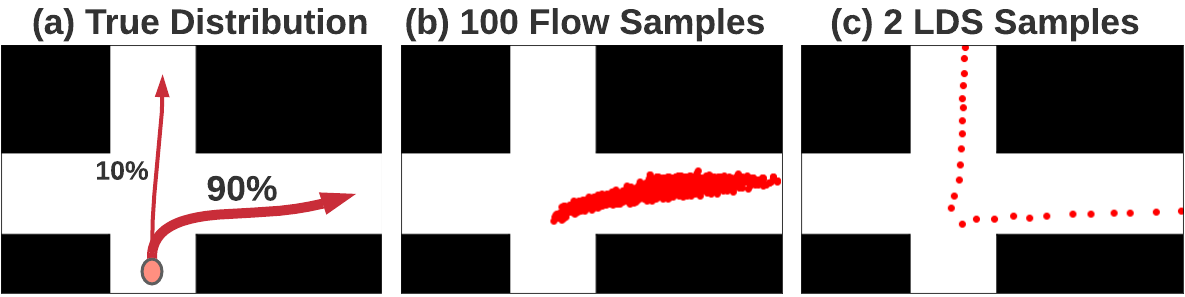}}
\caption{(a) At this intersection, 90\% of cars turn right, and 10\% drive straight in our training dataset. (b) A normalizing flow trajectory predictor trained on this data, sampled 100 times i.i.d., does not produce any straight trajectories. (c) With our LDS sampler plugged in, the same predictor generates both straight and right-turn trajectories with just 2 samples. Details are in Appendix \ref{appendix:toy-example}.}
\label{fig:toy-example}
\end{figure}

With the increasing abundance of driving data~\cite{caesar2020nuscenes, chang2019argoverse}, a promising approach is to learn a deep generative model from data to predict the probability distribution over future trajectories~\cite{lee2017desire, gupta2018social, ivanovic2019trajectron, tang2019multiple, rhinehart2018r2p2, rhinehart2019precog, salzmann2020trajectron++, park2020diverse}. However, due to natural biases, sampling i.i.d.~from a deep generative model's prior distribution may fail to cover all modes in the trajectory distribution, especially given the uneven distribution of real-world traffic maneuvers. Consider the scenario in Figure \ref{fig:toy-example}(a) and a normalizing flow (NF) \cite{rezende2015variational} forecasting model \cite{rhinehart2018deep, rhinehart2019precog}. The i.i.d.~forecasts from the flow model in Figure \ref{fig:toy-example}(b) successfully capture the major mode corresponding to turning right; however, for driving safely at this intersection, we must also anticipate the minor mode corresponding to vehicles driving straight.

\begin{figure*}
\centering
\resizebox{0.8\textwidth}{!}{
\includegraphics[]{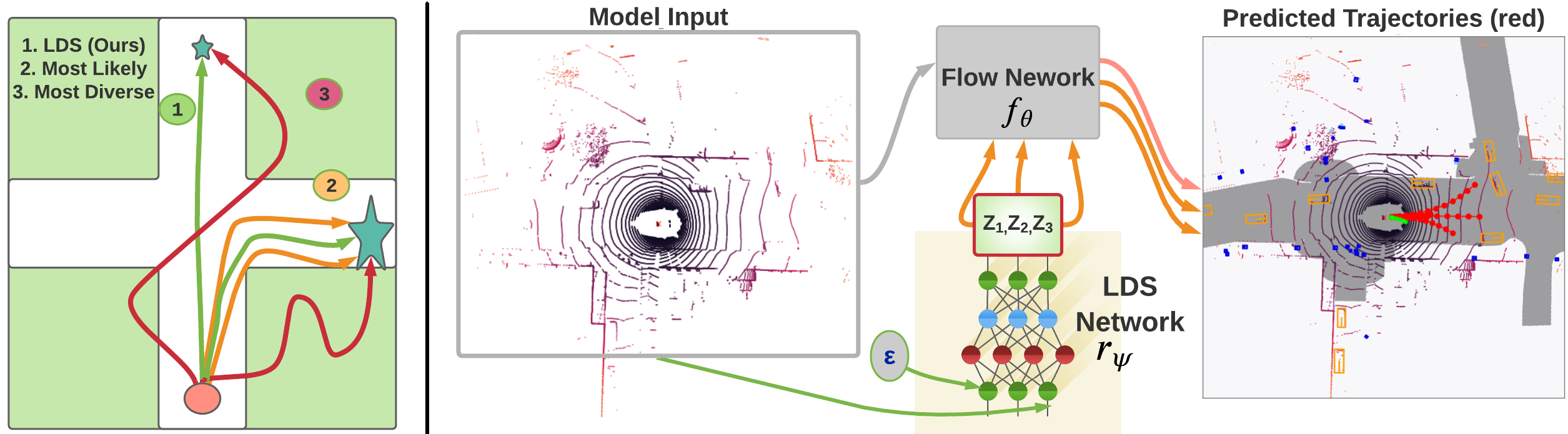}}
\caption{\textbf{Left}: 
Fitted on data at this intersection where most vehicles turn right and a small fraction head straight,
\textbf{(1)}: LDS predicts a set of paths that cover both modes and are realistic, \textbf{(2)}: Optimizing for only model likelihood generates samples that are realistic but miss the minor mode (small star) at the top, and \textbf{(3)}: Optimizing for only diversity generates samples that may cover both modes but are not realistic. \textbf{Right}: LDS architecture overview. LDS replaces standard i.i.d.~sampling in the flow model with a learned joint distribution over a set of samples in the latent space. These samples allow the pre-trained flow model to output diverse and realistic trajectories.}
\label{fig:LDS-architecture}
\end{figure*}

We propose a general, \textit{post-hoc} approach, called
\textbf{L}ikelihood-Based \textbf{D}iverse \textbf{S}ampling (LDS),
for enhancing the quality and the diversity of samples from a pre-trained generative model. The key idea is that rather than drawing i.i.d.~samples from the generative model, LDS learns a sampling distribution over an entire \emph{set} of trajectories, which jointly maximizes two objectives: (i) the likelihood of the trajectories according to the model, and (ii) a robust goal-based diversity objective that encourages high final spatial separation among trajectories. Intuitively, these two objectives together encourage a set of forecasts to cover modes in the underlying trajectory distribution. The result of running LDS on our toy example is shown in Figure \ref{fig:toy-example}(c); it correctly discovers the minor mode of heading straight and distributes samples over both modes. Figure \ref{fig:LDS-architecture} provides an overview of the LDS objective and architecture. Because our technique leverages trajectory likelihood under the learned underlying generative model, it is naturally suited for NF-based models, which compute the exact likelihood of their generated samples. To the best of our knowledge, our method is the first diverse sampling technique tailored for NF models. We note, however, that our technique can also be applied to other likelihood-based models---e.g., we can handle VAEs by using the evidence lower-bound for the likelihood.

A key advantage of LDS is that it can be leveraged in the setting of \textit{transductive learning} \cite{vapnik1998statistical}. In particular, since LDS does not require paired trajectories (history and future) for training, it can directly tailor the sampler to improve predictions for a given novel test instance, even without any prior training. By specializing to this test instance, transductive learning can outperform supervised learning which cannot perform test-time adaptation. To the best of our knowledge, the tranductive setting has not been previously considered for trajectory forecasting, but we believe it most closely mirrors the forecasting task autonomous vehicles face in the real world. 

LDS is simple to implement, requiring fewer than 30 lines of code. We evaluate LDS on nuScenes \cite{caesar2020nuscenes} and Forking Paths \cite{liang2020garden}, two challenging single-future and multi-future \cite{liang2020garden} benchmarks. Our experiments demonstrate that LDS provides a reliable performance boost when plugged into various existing NF/CVAE forecasting models and outperforms competing approaches. These results further improve when LDS is applied for transductive forecasting. 

\section{Related Work}
\label{sec:related-work}
\noindent \textbf{Multi-Modal Forecasting}. There are various approaches to the multi-modal forecasting problem. One approach is to pre-define trajectory primitives that serve as candidate outputs and formulate the forecasting problem as a mix of classification and regression \cite{deo2018convolutional, chai2019multipath,cui2019multimodal, phan2020covernet}. However, these approaches require extra labeled information; furthermore, they often only output a deterministic set of predictions for a given input. We instead build on deep generative models \cite{gupta2018social, lee2017desire, tang2019multiple, ivanovic2019trajectron, salzmann2020trajectron++, rhinehart2018deep, park2020diverse} that directly model multi-modal densities. In particular, normalizing flows \cite{rezende2015variational, papamakarios2019normalizing} have become a popular choice due to their comparative ease of optimization \cite{rhinehart2018r2p2, park2020diverse, rhinehart2018deep, filos2020can}, as well as their flexibility for downstream tasks such as goal-conditioned planning \cite{rhinehart2019precog}. Our method is designed as a plug-in improvement for sampling better and more diverse predictions from any model within this family. Our method is also compatible with VAE \cite{kingma2013auto}-based forecasting  models \cite{lee2017desire, yuan2019diverse, salzmann2020trajectron++, mangalam2020not}. Our diversity loss is formulated using predicted trajectories' endpoints. Similar ideas have been explored recently \cite{mangalam2020not, zhao2020tnt, choi2020shared}, but our method is a post-hoc approach and infers plausible endpoints in an unsupervised manner while previous works are end-to-end architectures and require ground truth endpoints to guide model training.

\para{Post-Hoc Neural Diversity Sampling}. Several prior works learn to sample from pre-trained generative models~\cite{batra2012diverse, gong2014diverse, elfeki2019gdpp, mangalam2020not}. Our approach is most closely related to two recent neural post-hoc diverse sampling methods for forecasting, DSF \cite{yuan2019diverse} and DLow \cite{yuan2020dlow}. DSF uses a neural network to parameterize a deterministic point process (DPP) \cite{kulesza2012determinantal} from which samples are drawn. The DPP kernel chooses samples using a threshold function based on the Euclidean distance of the latent samples from its prior distribution mean. DSF inherits the disadvantages of slow training and inference from its DPP. Computational issues aside, DSF fails to scale to high-dimensional latent spaces where Euclidean distance is not an informative metric of closeness. In comparison, DLow uses a modified ground truth trajectory reconstruction loss and KL divergence to shape the latent samples and an exponential kernel function to induce sample diversity. However, it restricts the architecture of its sampling network to be a single-layer linear network with respect to the latent samples to admit tractable KL-constraint
computation in its objective. This limits the expressiveness of the learned sampling distribution. In addition, because its objective requires ground-truth trajectory futures for training, it cannot be used in the transductive setup we introduce in our experiments. Both DSF and DLow also introduce additional kernel-related hyperparameters that are difficult to optimize. Compared to DSF and DLow, LDS permits multi-layered sampling architectures and high-dimensional latent spaces, exploits trajectory likelihood under the generative model to permit flexible application including transductive forecasting, introduces few hyperparameters, and performs consistently better across our experiments.

\section{Problem Setup}

Consider the problem of predicting the trajectory of an agent whose 2D position at time $t$ is denoted as $\mathbf{S}_t =(x_t,y_t)$. 
We denote the current time step as $t=0$, and the future aggregated state as $\mathbf{S} \coloneqq \mathbf{S}_{1:T} \in \mathbb{R}^{T\times 2}$. At time $t=0$, the agent has access to observation $\mathbf{o}$, which may include contextual features such as Lidar scans, physical attributes of the vehicle/pedestrian agent (e.g. velocity, yaw), and the state histories of all agents in the scene. The goal of trajectory forecasting is to predict $\mathbf{S}$ given $\mathbf{o}$, $p(\mathbf{S}|\mathbf{o})$. We denote the training dataset as $\mathcal{D} =\{(\mathbf{o}, \mathbf{S})\}$. 

Our approach assumes as given, a flow model $f_{\theta}$ that has been pre-trained to learn the distribution $p_{\theta}(\mathbf{S}|\mathbf{o}; \mathcal{D})$. At a high level, assuming a multivariate Gaussian base sampling distribution $\mathbf{Z} \sim P_{\mathbf{Z}} \equiv \mathcal{N}(0, \mathbf{I})$, $f_{\theta}$ is a bijective mapping between $\mathbf{Z}$ and $\mathbf{S}$, captured by the following forward and inverse computations of $f_{\theta}$: 
\begin{equation}
\label{eq:flow-relationship}
\mathbf{S} =  f_{\theta}(\mathbf{Z};\textbf{o}) \sim p_{\theta}(\mathbf{S} \mid \mathbf{o}), \quad \mathbf{Z} =  f_{\theta}^{-1}(\mathbf{S};\mathbf{o}) \sim P_{\mathbf{Z}} 
\end{equation}
To draw one trajectory sample $\mathbf{S}$, we sample $\mathbf{Z} \sim P_{\mathbf{Z}}$ and compute $\mathbf{S} = f_{\theta}(\mathbf{Z};\mathbf{o})$. Furthermore, the exact likelihood of a trajectory $\mathbf{S}$ is given by the change of variables rule:
\begin{equation} 
\begin{aligned}
\log p_{{\theta}}(\mathbf{S} | \mathbf{o})
&=  \log \left( p(\mathbf{Z}) \cdot \left| \text{det}\; \frac{\text{d} f_{\theta}}{\text{d} \mathbf{Z}}\Bigr\vert_{\mathbf{Z}= f_{\theta}^{-1}(\mathbf{S}; \mathbf{o})} \right|1^{-1} \right), \\
\end{aligned}
\label{eq:exact-likelihood}
\end{equation}
where the bijective property and standard architectural choices for $f_\theta$ permit easy computation of the determinant. We refer readers to Appendix \ref{appendix:flow} for a more detailed introduction to flow-based trajectory forecasting. 

\section{Diversity Sampling for Flow} 
\label{sec:LDS}

In stochastic settings, it is often necessary to use $K>1$ trajectory predictions rather than just one, to ensure that the samples cover the full range of possible stochastic futures; we assume the number of predictions $K$ is a given hyperparameter. However, as Figure~\ref{fig:toy-example} shows, simply drawing $K$ i.i.d.~samples from the flow model $f_\theta$ may undersample from minor modes and fail to capture all potential outcomes. We propose an alternative strategy, which we call \textbf{L}ikelihood-Based \textbf{D}iverse \textbf{S}ampling (LDS), that learns a joint distribution over $K$ samples $\{\mathbf{Z}_1,...,\mathbf{Z}_K\}$ in the latent space of $f_\theta$. In doing so, it aims to improve the diversity of the trajectories $f_{\theta}(\mathbf{Z}_1),...,f_{\theta}(\mathbf{Z}_K)$ while maintaining their plausibility according to the flow model\footnote{Though our technical discussion focuses on NF models, we emphasize that LDS can also be applied to CVAE models by replacing NLL with ELBO; we provide details on LDS-CVAE in Appendix \ref{appendix:LDS-cvae-objective}.}.

In particular, LDS trains a neural network $r_{\psi}$ to transform a Gaussian distribution $\epsilon \sim \mathcal{N}(0, \mathbf{I})$ into a distribution over a set $\mathcal{Z} \coloneqq \{\mathbf{Z}_1,..., \mathbf{Z}_K\} = r_{\psi}(\epsilon; \mathbf{o})$ of latent vectors given an observation $\mathbf{o}$. This set in turn induces a distribution over trajectories $\mathcal{S} \coloneqq \{\mathbf{S}_1,...,\mathbf{S}_K\}$, where $\mathbf{S}_k=f_{\theta}(\mathbf{Z}_k;\mathbf{o})$ for each $k$. Since the distribution is defined over multisets of samples, the individual samples $\mathbf{S}_k$ are no longer independent. Informally, they should be anti-correlated to ensure they cover different modes. We train $r_{\psi}$ to minimize the following loss function:
\begin{equation}
\label{eq:LDS-full-loss}
\text{L}_{\text{LDS}}(\psi) \coloneqq \text{NLL}(\psi) - \lambda_d \text{L}_{\text{d}}(\psi),
\end{equation}
which combines the negative log likelihood (NLL) loss from the flow model and a goal diversity loss $\text{L}_\text{d}$. Figure \ref{fig:LDS-architecture} (Left) provides intuition for these two terms, and we explain them in detail below.

\para{Likelihood Objective.} The NLL term is defined as:
\begin{align}
\text{NLL}(\psi) & \coloneqq -\sum_{k=1}^K \log p_{\theta} (\mathbf{S}_k|\mathbf{o}_k),
\label{eq:LDS-likelihood-loss}
\end{align}
where $\{\mathbf{S}_1,...,\mathbf{S}_k\} = f_{\theta}(r_{\psi}(\epsilon; \mathbf{o}))$ and $\log p_{\theta} (\mathbf{S}_k|\mathbf{o})$ is computed as in Equation~\eqref{eq:exact-likelihood}. This NLL term incentivizes LDS to output a set of forecasts that all have high likelihood according to the flow model $f_\theta$. This term incentivizes $f_\theta$ to maximize the likelihood of the training trajectories $\mathcal{D}$ by selecting trajectories that are plausible and likely to occur. By itself, it does not incentivize diversity among forecasts; they may easily concentrate around the major mode as in the ``most likely'' trajectories in Figure~\ref{fig:LDS-architecture} (Left).

\para{Diversity Objective.} To combat this tendency, we introduce the \emph{minimum} pairwise squared $\text{L}_2$ distance between the predicted trajectory endpoints:
\begin{equation}
\label{eq:LDS-diversity-loss}
\text{L}_{\text{d}}(\psi) \coloneqq \min_{i \neq j \in K} \norm{f_{\theta}(\mathbf{Z}_i)_{T} - f_{\theta}(\mathbf{Z}_j)_{T}}_2^2.
\end{equation}
The minimum formulation strongly incentivizes LDS to distribute its samples among different modes in the distribution, since any two predictions that end up too close to each other would significantly decrease $\text{L}_{\text{d}}$. In our experiments, we have observed \textit{mean}-based diversity formulation suffers from network ``cheating" behavior, in which all but one prediction collapse to a single trajectory and the left-out trajectory is distant, resulting in comparatively high \textit{average} diversity. Our formulation is robust to this degeneracy as only the pairwise minimum will be considered. While many other notions of distances between trajectories are compatible with our framework, we measure the distances at the last time step alone, since spatial separation between trajectory endpoints is a good measure of trajectory diversity in our applications \cite{mangalam2020not, zhao2020tnt, choi2020shared}. Finally, to train $r_{\psi}$, LDS minimizes $\text{L}_{\text{LDS}}$ using stochastic gradient descent; see Algorithm \ref{algo:LDS-full}. 

\begin{algorithm}[t]
\centering
\caption{Batch LDS Training}
\label{algo:LDS-full}
\begin{algorithmic}[1]
\STATE {\bfseries Input:} Flow $f_{\theta}$, Observation Batch $\{\mathbf{o}\}$
\STATE Initialize LDS model $r_{\psi}$
\FOR{$\mathbf{o}_i \in \{\mathbf{o}\}$}
\STATE Sample $\mathbf{\epsilon} \sim \mathcal{N}(0, \mathbf{I})$
\STATE Compute  $\mathbf{Z}_1,..., \mathbf{Z}_K = r_{\psi}(\mathbf{\epsilon};\mathbf{o}_i)$
\STATE Generate predictions $f_{\theta}(\mathbf{Z}_1),.., f_{\theta}(\mathbf{Z}_K)$
\STATE Compute losses using Equations \ref{eq:LDS-likelihood-loss} \& \ref{eq:LDS-diversity-loss}
\ENDFOR 
\STATE Perform stochastic gradient descent on $\psi$ to minimize $\text{L}_{\text{LDS}}$ (Equation \ref{eq:LDS-full-loss})
\STATE {\bfseries Output:} Trained LDS model $r_{\psi}$
\end{algorithmic}
\end{algorithm}

\para{Implementation Details.} LDS $r_{\psi}$ is implemented as a 3-layer feed-forward neural network with $K$ heads, each of which corresponds to the latent $z_k$ for a single output in the predicted set. We assume access to the input embedding layers of the pre-trained flow model, which embeds high-dimensional (visual) input $\textbf{o}$ into lower dimensional feature vectors. These feature vectors are taken as input to LDS. Additionally, we fix the dimensions of the input Gaussian noise $\epsilon$ to be the same as the trajectory output $\mathbf{S}$ from $f_\theta$ (i.e. $T \times 2$). Furthermore, to prevent the diversity loss from diverging, we clip it to a positive value. Additional details are in Appendix \ref{appendix:nuscenes-LDS-details} \& \ref{appendix:forking-paths-LDS-details}.

\section{Transductive Trajectory Forecasting}
\label{sec:transductive-learning}

Transductive learning refers to ``test-time training" on unlabeled test instances \cite{vapnik1998statistical, joachims2003transductive}. For diversity sampling, it amounts to the following: \textit{given a new observation \textbf{o}, compute a set of diverse trajectories that best captures the possible stochastic futures}. That is, whereas supervised learning focuses on \textit{average} accuracy across all test trajectories, transductive learning focuses on the vehicle's current situation. This setting closely captures the forecasting problem autonomous vehicles face in practice; however, existing end-to-end forecasting models typically lack the capability for the transductive setting because their training procedures require data with ground truth labels.

In contrast, LDS's objective function does not depend on access to any ground truth future trajectories, and relies only on the pre-trained generative model and unlabeled inputs $\mathbf{o}$. In this sense, the LDS sampler is trained without supervision. It can therefore be used transductively and adapt on-the-fly to a given new observation (e.g. a new traffic scene the vehicle enters). Formally, given an unlabeled input $\mathbf{o}$, we can train a LDS model $r_{\psi}$ tailored to $\mathbf{o}$. We call this variant \textbf{LDS-TD-NN}. Compared to vanilla ``batch'' LDS, LDS-TD-NN does not need to be trained offline with a large dataset, making it also suitable for settings where little or no training data is available ---e.g., the training set for the pre-trained flow model is unavailable.

In the transductive setting, LDS could eschew the neural network and directly optimize the latent space samples $\textbf{Z}_1,...,\textbf{Z}_K$. We call this \textit{particle} variant \textbf{LDS-TD-P}. Note that, unlike the neural variant, LDS-TD-P cannot take the observation $\mathbf{o}$ as input. LDS-TD-NN and LDS-TD-P are summarized in Algorithm \ref{algo:LDS-td} and \ref{algo:LDS-td-p} in Appendix \ref{appendix:LDS-pseudocode}.

\section{Experiments}
\label{sec:experiments}

Our experiments aim to address the following questions:
\textbf{(1)} Does LDS boost performance across different pre-trained flow models? 
\textbf{(2)} Can LDS be applied to pre-trained VAE models as well?,
\textbf{(3)} How does LDS compare against (a) other learning-based diverse sampling methods and (b) existing end-to-end multi-modal forecasting models?,
\textbf{(4)} Is LDS effective in the transductive learning setting?
\textbf{(5)} Which components of LDS are most important for performance?
We address questions 1-4 via experiments on two qualitatively different datasets, nuScenes \cite{caesar2020nuscenes} and Forking Paths \cite{liang2020garden} (Section \ref{sec:quantitative-results} and \ref{sec:qualitative-results}). For each one, we use a distinct pre-trained flow model with architecture and input representation tailored to that dataset. We address question 5 via an ablation study in Section \ref{sec:ablation-studies}. 

\subsection{Datasets and Models} 
We begin by describing our datasets, models, and evaluation metrics; see Appendix \ref{appendix:nuscenes-details} \& \ref{appendix:forking-paths-details} for details on model architectures, hyperparameters, and training procedures. 

\para{NuScenes.} NuScenes is a multi-purpose autonomous vehicle dataset with a large trajectory forecasting dataset \cite{caesar2020nuscenes}. Following prior work \cite{cui2019multimodal, phan2020covernet, filos2020can}, the predictor takes as input the current observation (e.g., Lidar scan) and attributes (e.g., velocity) of a vehicle, and forecasts this vehicle's trajectory over the next $6$ seconds (i.e., $12$ frames).

\para{LDS Models.} We train an autoregressive affine flow model (\textbf{AF}) proposed by \cite{rhinehart2018deep, rhinehart2019precog} as our underlying flow model for trajectory prediction. On top of \textbf{AF}, we train three variants of LDS. The first is $\textbf{LDS-AF}$, the batch version in Algorithm \ref{algo:LDS-full}. The latter two are the transductive neural and particle-based variants $\textbf{LDS-AF-TD-NN}$ and $\textbf{LDS-AF-TD-P}$ discussed in Section \ref{sec:transductive-learning}. To further illustrate the generality of LDS, we also consider $\textbf{LDS-CVAE}$, a LDS applied to a CVAE model. Here, we replace the exact likelihood in Equation \ref{eq:LDS-likelihood-loss} with ELBO.

\noindent\textbf{Baselines.} For neural diverse sampling approaches, we consider DLow and DSF and apply them to both CVAE and AF to form DLow-\{AF, CVAE\} and DSF-\{AF, CVAE\}. In addition to neural diverse sampling baselines, we include three end-to-end multi-modal forecasting models: $\textbf{CoverNet}$ \cite{phan2020covernet}, $\textbf{MultiPath}$ \cite{chai2019multipath}, and $\textbf{MTP}$ \cite{cui2019multimodal}. For the first two, we directly report results published in~\cite{phan2020covernet}; for $\textbf{MTP}$, we re-train the model using nuScenes' official implementation to provide an end-to-end model for evaluation metrics not included in~\cite{phan2020covernet}. We choose these end-to-end models for comparison since they use the same inputs as our AF and CVAE backbones, and we aim to test whether post-hoc sampling methods applied to simple backbone models are competitive with specialized end-to-end models. In Appendix \ref{appendix:nuscenes-additional-results}, we compare to Trajectron++ \cite{salzmann2020trajectron++}, which uses an experimental setup different from other prior approaches.

\para{Forking Paths.} One limitation of most trajectory forecasting datasets such as nuScenes is that there is only a single ground-truth future trajectory for each training sample. To evaluate each forecasting model's ability to predict diverse, plausible future trajectories, it is critical to evaluate the model against multiple ground-truth trajectories in a multi-future dataset. This approach directly evaluates whether a model captures the intrinsic stochasticity in future trajectories. Therefore, we additionally evaluate LDS on the recent Forking Paths (FP) dataset \cite{liang2020garden}. FP recreates scenes from real-world pedestrian trajectory datasets \cite{oh2011large, awad2018trecvid} in the CARLA simulator \cite{dosovitskiy2017carla}, and asks multiple human annotators to annotate future trajectories in the simulator, thereby creating multiple ground truth future pedestrian trajectories for each scene. The flow model takes as input the trajectory of a pedestrian over the past 3 seconds (i.e., 12 frames), and its goal is to predict their trajectory over the next 5 seconds (i.e., 20 frames).

\para{LDS Models.} For this dataset, we focus on flow models and use the recently introduced Cross-Agent Attention Model Normalizing Flow (\textbf{CAM-NF}) \cite{park2020diverse} as our underlying flow model for trajectory prediction; see Appendix \ref{appendix:forking-paths-details} for CAM-NF details. Compared to AF used for nuScenes experiments, CAM-NF is an already performant generative model, and an additional goal of this experiment is to investigate whether LDS can be useful for already performant generative models. As before, we train \textbf{LDS} and \textbf{LDS-TD-NN} on top of CAM-NF.

\para{Baselines.} We compare to \textbf{DSF} and \textbf{DLow} applied to CAM-NF. All other baseline results are taken directly from \cite{liang2020garden} (rows with $*$ in Table \ref{table:LDS-ablation-loss-function} (Left)), including \textbf{Social-LSTM} \cite{alahi2016social}, \textbf{Social-GAN} \cite{gupta2018social}, \textbf{Next} \cite{liang2019peeking}, and \textbf{Multiverse} \cite{liang2020garden}, as well as simple \textbf{Linear} and \textbf{LSTM} networks. 

\para{Training and Evaluation.} We follow the procedure in \cite{liang2020garden}. We first train CAM-NF using VIRAT/ActEV \cite{oh2011large, awad2018trecvid}, the real-world datasets from which FP extracts simulated pedestrian scenes. Then, we train \{DLow, DSF, LDS\} on top of the pre-trained CAM-NF model using the training dataset (VIRAT/ActEV). Finally, we evaluate all models on the test set FP using $K=20$ samples against the multiple ground-truth futures. For LDS-TD-NN, we directly train and evaluate $r_\psi$ on FP using small minibatches as described in Algorithm \ref{algo:LDS-td} in Appendix \ref{appendix:LDS-pseudocode}. 

An important challenge is that trajectories in FP have different (typically longer) lengths compared to trajectories in VIRAT/ActEV, since the human annotators provided trajectories of varying durations; this complicates the forecasting problem on the FP test set.

\subsection{Evaluation Metrics}

We report minimum average displacement error $\textbf{minADE}_\text{K}$ and final displacement error $\textbf{minFDE}_\text{K}$ of $K$ prediction samples $\hat{\mathbf{S}}_{k}$ compared to the ground truth trajectories $\mathbf{S}_1,...,\mathbf{S}_J$ \cite{tang2019multiple, chai2019multipath, liang2020garden}:
$$
\begin{aligned} 
&\text{minADE}_K(\hat{\mathbf{S}}, \mathbf{S}) = \frac{\sum_{j=1}^J \min_{i \in K} \sum_{t=1}^T \norm{\hat{\mathbf{S}}_{i,t} - \mathbf{S}_t}^2}{T \times J},\\ 
&\text{minFDE}_K(\hat{\mathbf{S}}, \mathbf{S}) =\frac{\sum_{j=1}^J \min_{i \in K} \norm{\hat{\mathbf{S}}_{i,T} - \mathbf{S}_T}^2}{J}
\end{aligned}
$$    
These metrics are widely used in stochastic prediction tasks \cite{tang2019multiple, gupta2018social} and tend to reward predicted sets of trajectories that are both diverse and realistic. In multi-future datasets ($J>1$) such as Forking Paths, these metrics are standalone sufficient to evaluate both the diversity and the plausibility of model predictions, because a set of predictions that does not adequately cover all futures will naturally incur high errors. In single-future datasets ($J=1$) such as nuScenes, however, they do not explicitly penalize a predicted set of trajectories that simply repeats trajectories close to the single ground truth. To explicitly measure prediction diversity on nuScenes, we also report the minimum average self-distance $\textbf{minASD}_\text{K}$ and minimum final self-distance $\textbf{minFSD}_\text{K}$ between pairs of predictions samples:
$$
\begin{aligned}
\text{minASD}_{\text{K}}(\hat{\mathbf{S}}) &= \min_{i \neq j \in K} \frac{1}{T} \sum_{t=1}^T \norm{\hat{\mathbf{S}}_{i,t} - \hat{\mathbf{S}}_{j,t}}^2 \\
\text{minFSD}_{\text{K}}(\hat{\mathbf{S}}) &= \min_{i \neq j \in K} \norm{\hat{\mathbf{S}}_{i,T} - \hat{\mathbf{S}}_{j,T}}^2
\end{aligned}
$$
These metrics evaluate the lower bound diversity among a predicted set of trajectories, and they tend to decrease as $K$ increases since the predictions become more ``crowded'' around the modes already covered. Note that minFSD is identical to the diversity term in the LDS objective (Equation~\eqref{eq:LDS-diversity-loss}). Several prior works have reported the \textit{average} ASD (meanASD) and FSD (meanFSD) instead \cite{yuan2019diverse, yuan2020dlow}; however, we observe that minASD is a superior metric since it is more robust to outliers among the predictions (see Appendix \ref{appendix:additional-ablation-results} for an illustrative example). For completeness, we also report meanASD and meanFSD in Appendix \ref{appendix:nuscenes-additional-results}; our findings are consistent with those presented here. Finally, since $\text{minFDE}_\text{K}$, $\text{minASD}_{\text{K}}$, and $\text{minFSD}_{\text{K}}$ were not reported in previous work, we only report them for the models we implement.

All compared models, except vanilla CVAE and NF, take the number $K$ of modes/number of samples in the prediction set as a hyperparameter. We report results for each metric using the corresponding model configurations---e.g., when measuring $\text{minASD}_{\text{5}}$, we use $K=5$ for all models.

\begin{table*}
\centering
\resizebox{\textwidth}{!}{
\begin{tabular}{ccccccc|cccc}
\toprule
Method    & Modes    & $\text{minADE}_{1} (\downarrow)$ & $\text{minADE}_{5} (\downarrow)$ & $\text{minADE}_{10} (\downarrow)$ & $\text{minFDE}_{5} (\downarrow)$ & $\text{minFDE}_{10} (\downarrow)$ & $\text{minASD}_5$ $(\uparrow)$ & $\text{minFSD}_5$ $(\uparrow)$  & $\text{minASD}_\text{10}$ $(\uparrow)$ & $\text{minFSD}_\text{10}$ $(\uparrow)$ \\
\midrule
\textbf{MultiPath}$^*$ \cite{chai2019multipath} & 64 & 5.05 & 2.32 & 1.96 & -- & -- & --& --& --& --\\ 
\textbf{CoverNet}$^*$ \cite{phan2020covernet}& 232 & 4.73 & \textbf{2.14} & \textbf{1.72} & -- &--& --& --& --& --\\ 
\textbf{MTP} \cite{cui2019multimodal} & 5, 10 & 4.68 $\pm$ 1.04 & 2.61 $\pm$ 0.17 & 1.84 $\pm$ 0.04 & 5.80 $\pm$ 0.49 & 3.72 $\pm$ 0.07 & 1.74 $\pm$ 0.32 & 4.31 $\pm$ 1.60 & 0.97 $\pm$ 0.15 & 2.43 $\pm$ 0.34\\
\midrule 

\textbf{CVAE} & N/A & 4.20 $\pm$ 0.03& 2.71 $\pm$ 0.03  & 2.08 $\pm$ 0.02 & 6.20 $\pm$ 0.05& 4.58 $\pm$ 0.05 & 1.28 $\pm$ 0.03 & 2.99 $\pm$ 0.07 & 0.57 $\pm$ 0.02 & 1.30 $\pm$ 0.04\\
\textbf{DSF-CVAE} \cite{yuan2019diverse} &5, 10& \multicolumn{1}{c}{--} & 2.54 $\pm$ 0.21  & 2.02 $\pm$ 0.11 & 5.77 $\pm$ 0.51& 4.44 $\pm$ 0.27 & 1.38 $\pm$ 0.22 &  3.33 $\pm$ 0.58 & 0.78 $\pm$ 0.04 &1.85 $\pm$ 0.13 \\
\textbf{DLow-CVAE} \cite{yuan2020dlow} &5, 10 & \multicolumn{1}{c}{--} & 2.23 $\pm$ 0.13& \textbf{1.75} $\pm$ 0.03& 5.00 $\pm$ 0.29& \textbf{3.71} $\pm$ 0.08 & 2.64 $\pm$ 0.25 & 6.38 $\pm$ 0.65 &  1.18 $\pm$ 0.16 & 2.73 $\pm$ 0.43\\
\textbf{LDS-CVAE} (Ours) &5, 10 & \multicolumn{1}{c}{--} & 
\textbf{2.16} $\pm$ 0.03 & \textbf{1.75} $\pm$ 0.05 & \textbf{4.82} $\pm$ 0.06 & \textbf{3.71} $\pm$ 0.14 & \textbf{3.02} $\pm$ 0.23 & \textbf{7.46} $\pm$ 0.44 & \textbf{1.74} $\pm$ 0.46 & \textbf{4.07} $\pm$ 1.10 \\
\midrule
\textbf{AF}  & N/A & 4.01 $\pm$ 0.05 & 2.86 $\pm$ 0.01 & 2.19 $\pm$ 0.03 & 6.26 $\pm$ 0.05 & 4.49 $\pm$ 0.07 & 1.58 $\pm$ 0.02 & 3.75 $\pm$ 0.04 & 0.70 $\pm$ 0.01 & 1.63 $\pm$ 0.02\\
\textbf{DSF-AF} &5, 10& \multicolumn{1}{c}{--} & 2.61 $\pm$ 0.12 & 2.23 $\pm$ 0.10 & 5.91 $\pm$ 0.33 & 4.80 $\pm$ 0.23 & 0.87 $\pm$ 0.13 & 2.14 $\pm$ 0.41 & 0.44 $\pm$ 0.05 &1.11 $\pm$ 0.10\\
\textbf{DLow-AF}  & 5, 10 & \multicolumn{1}{c}{--} & \textbf{2.11} $\pm$ 0.01 & 1.78 $\pm$ 0.05 & \textbf{4.70} $\pm$ 0.03 & 3.77 $\pm$ 0.13 &  2.56 $\pm$ 0.12 & 6.45 $\pm$ 0.24 & 1.05 $\pm$ 0.11 &2.55 $\pm$ 0.28 \\  
\textbf{LDS-AF} (Ours) & 5, 10 & \multicolumn{1}{c}{--} & \textbf{2.06} $\pm$ 0.09 & \textbf{1.66} $\pm$ 0.02 & \textbf{4.67} $\pm$ 0.25 & \textbf{3.58} $\pm$0.05 & \textbf{3.13} $\pm$ 0.18 & \textbf{8.19} $\pm$ 0.26 & \textbf{2.11}  $\pm$ 0.05 & \textbf{6.22} $\pm$ 0.09\\
\midrule 
\textbf{LDS-AF-TD-P} (Ours) & 5, 10 & \multicolumn{1}{c}{--} & 2.46 $\pm$0.09 & 1.91 $\pm$ 0.04 & 5.21 $\pm$ 0.15 & 3.71 $\pm$ 0.11& 2.39 $\pm$ 0.08 & 7.07 $\pm$ 0.18& 1.60 $\pm$ 0.06& 5.70 $\pm$ 0.10\\  
\textbf{LDS-AF-TD-NN} (Ours) & 5, 10 & \multicolumn{1}{c}{--} & \textbf{2.06} $\pm$ 0.03 & \textbf{1.65} $\pm$ 0.02 & \textbf{4.62} $\pm$ 0.07 & \textbf{3.50} $\pm$ 0.05& \textbf{3.09} $\pm$ 0.07 & \textbf{8.15} $\pm$ 0.17 & \textbf{1.98} $\pm$ 0.03 & \textbf{5.91} $\pm$ 0.04\\
\bottomrule
\end{tabular}}
\caption{NuScenes prediction error results (lower is better) and diversity results (higher is better), including previously reported results (top), and results of LDS variants and newly implemented baselines (bottom). LDS-based models produce the most plausible and diverse predictions throughout.}
\label{table:nuscenes-accuracy}
\end{table*}


\subsection{Quantitative Results} 
\label{sec:quantitative-results}

\para{NuScenes.} In Table \ref{table:nuscenes-accuracy} (Left), we compare the prediction accuracy of LDS-AF, LDS-AF-TD-\{NN, P\}, and the baselines described above. The best method within each sub-category is bolded. \textbf{LDS-AF} and \textbf{LDS-AF-TD-NN} achieve the best overall performance.

Comparing AF and CVAE to prior multi-modal models, we see that although the two ``vanilla" generative models achieve better one-sample prediction (i.e., $\text{minADE}_1$), they perform significantly worse when more predictions are made. This confirms our hypothesis that i.i.d.~samples from a  generative model do not adequately capture diverse modes, causing it to fail to cover the ground truth with good accuracy. Consequently, post-hoc diverse sampling significantly improves the performance of these models. Out of all the post-hoc neural diverse sampling methods, LDS provides the most significant improvement for both AF and CVAE. In particular, the most performant model LDS-AF achieves the best results out of all models in the batch setting, even outperforming the strongest multi-modal model CoverNet. This suggests there is much to gain from a (simple) pre-trained model by applying appropriate post-hoc diverse sampling. We highlight that despite not being designed for CVAE, LDS still outperforms DSF and DLow, both of which are originally intended for CVAE, demonstrating the general merit of our approach. Next, in the transductive setting, LDS-AF-TD-NN is indeed able to tailor its predictions towards each small batch of test instances and achieves the overall best results among all models. Finally, LDS-AF-TD-NN also significantly outperforms the particle variant LDS-AF-TD-P, likely due to the neural variant having the advantage of explicitly conditioning the samples on the observation input $\mathbf{o}$. 

Next, we compare the models in terms of their prediction diversity in Table \ref{table:nuscenes-accuracy} (Right). LDS models consistently outperform the baseline models by a large margin. In particular, they are the only models whose diversity does not collapse when the number of modes increases from $5$ to $10$. This shows that LDS is more ``efficient'' with its samples, since it does not repeat any trajectories. In contrast, all other methods produce pairs of very similar predictions when $K=10$. Given that LDS also produces accurate predictions, these results provide strong evidence that LDS is able to simultaneously optimize accuracy and diversity. Furthermore, LDS also achieves the highest diversity under the mean diversity metrics in Appendix \ref{appendix:nuscenes-additional-results}.

\begin{figure*}
\centering
\resizebox{\textwidth}{!}{
\begin{tabular}{cccc}
&\textbf{LDS} (Ours) & \textbf{AF} & \textbf{MTP} \\ 
\rotatebox[]{90}{$\quad \quad \quad \quad \quad \textbf{Frame 1}$} & \includegraphics[width=.33\textwidth, trim={0 5cm 0 5cm}, clip=True]{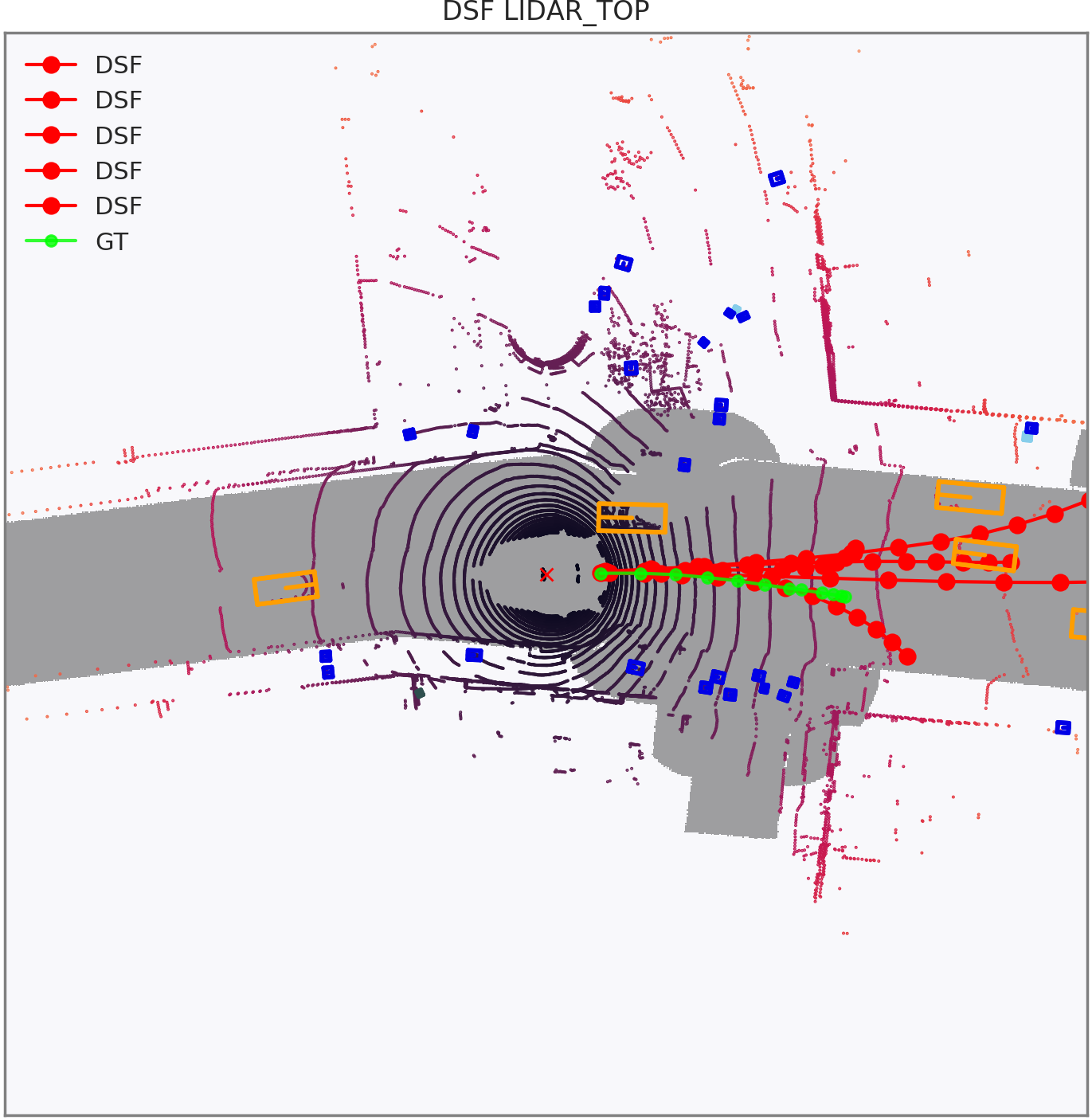}&
\includegraphics[width=.33\textwidth, trim={0 5cm 0 5cm}, clip=True]{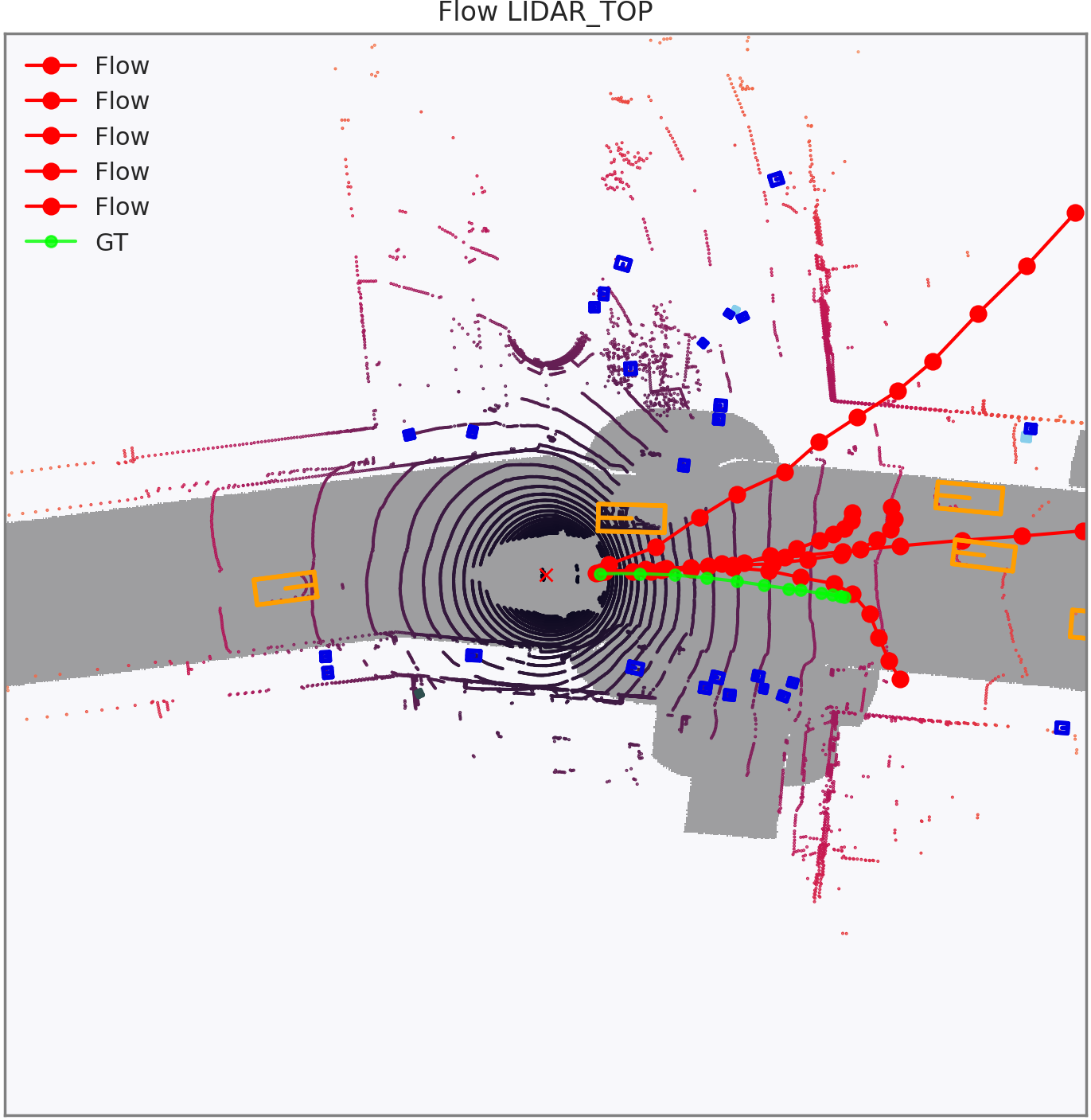}&
\includegraphics[width=.33\textwidth, trim={0 5cm 0 5cm}, clip=True]{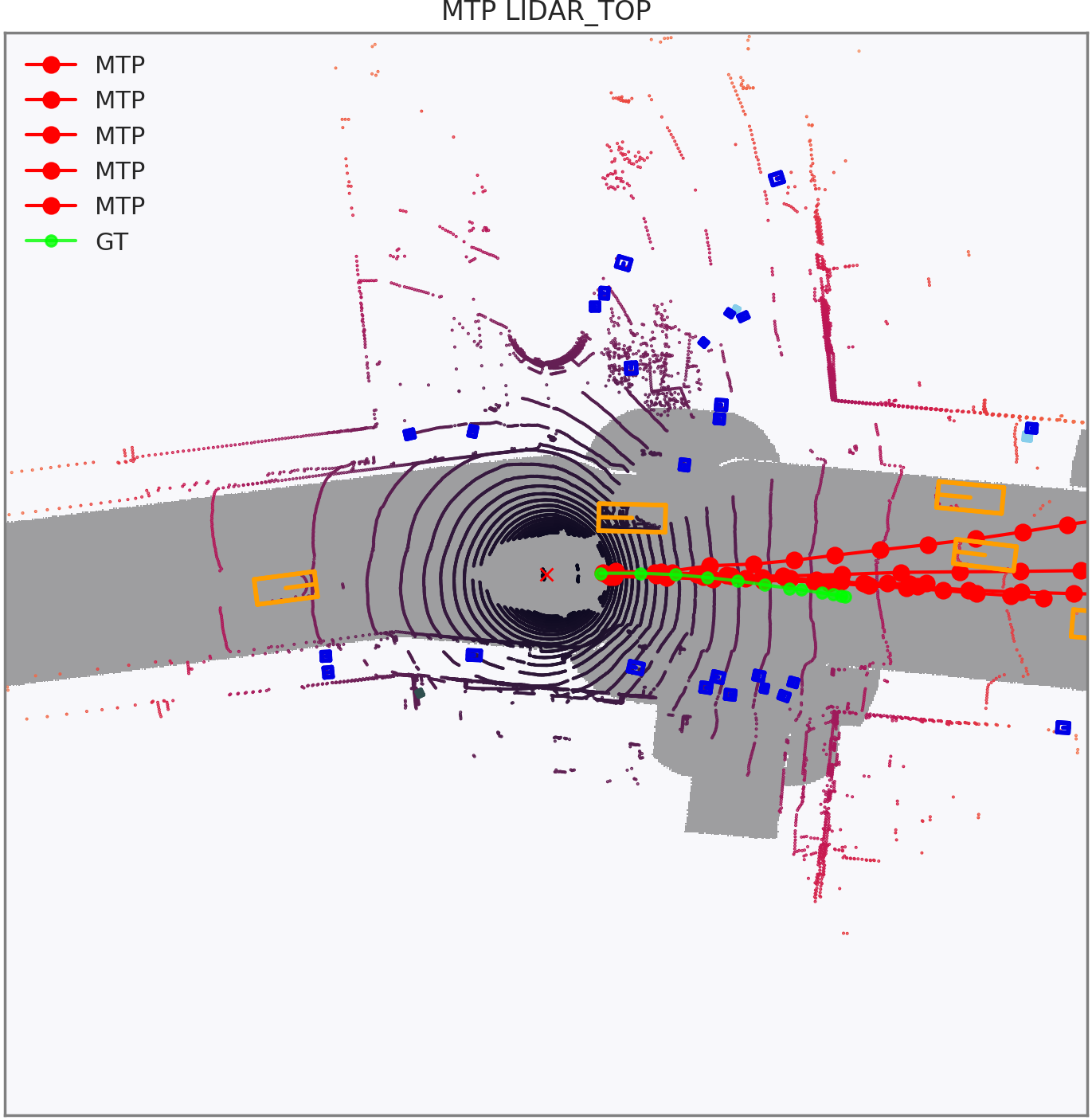}
\\[-1cm]
\rotatebox[]{90}{$\quad \quad \quad \quad \quad \textbf{Frame 2}$}& \includegraphics[width=.33\textwidth, trim={0 5cm 0 5cm}, clip=True]{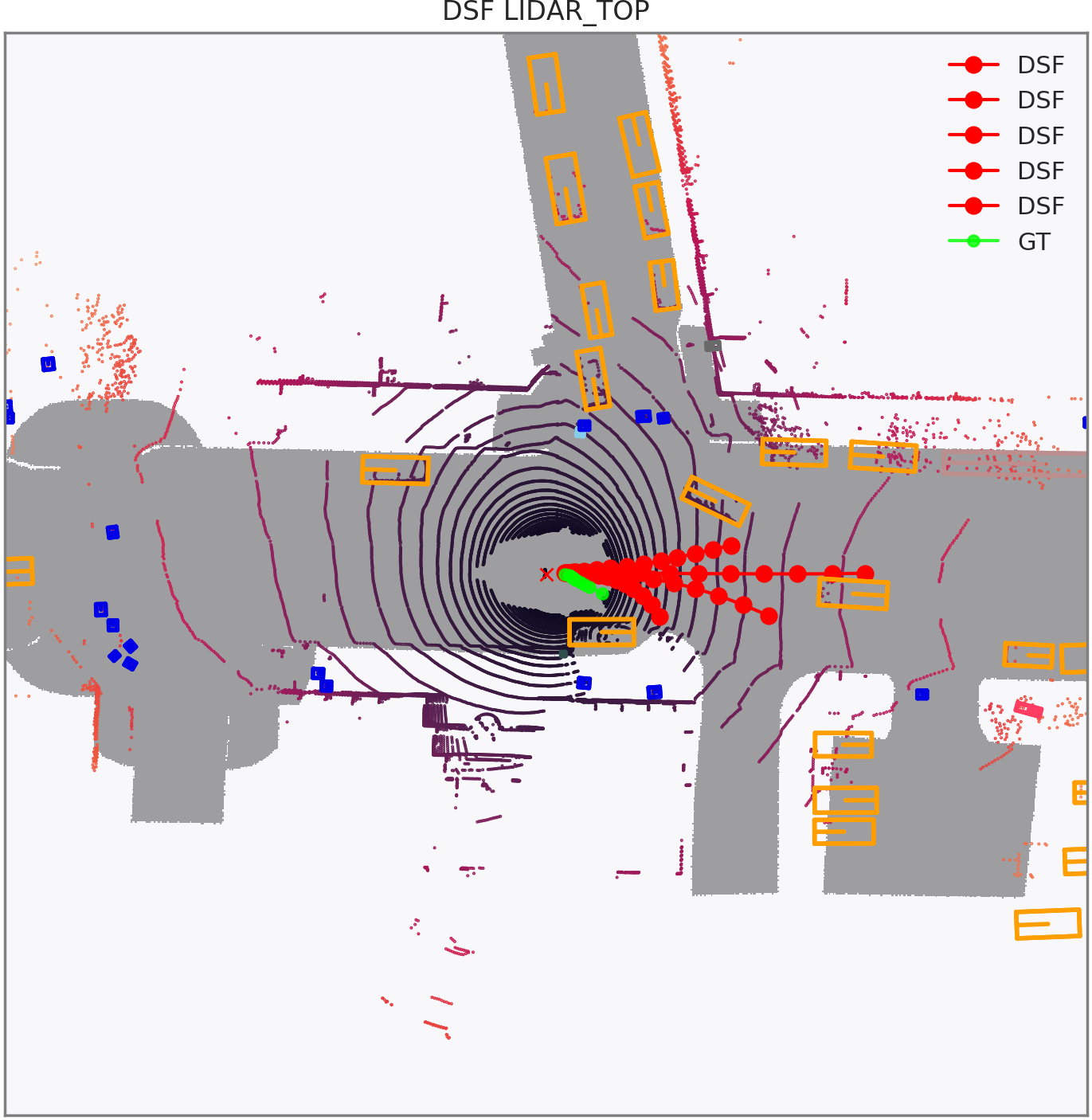}&
\includegraphics[width=.33\textwidth, trim={0 5cm 0 5cm}, clip=True]{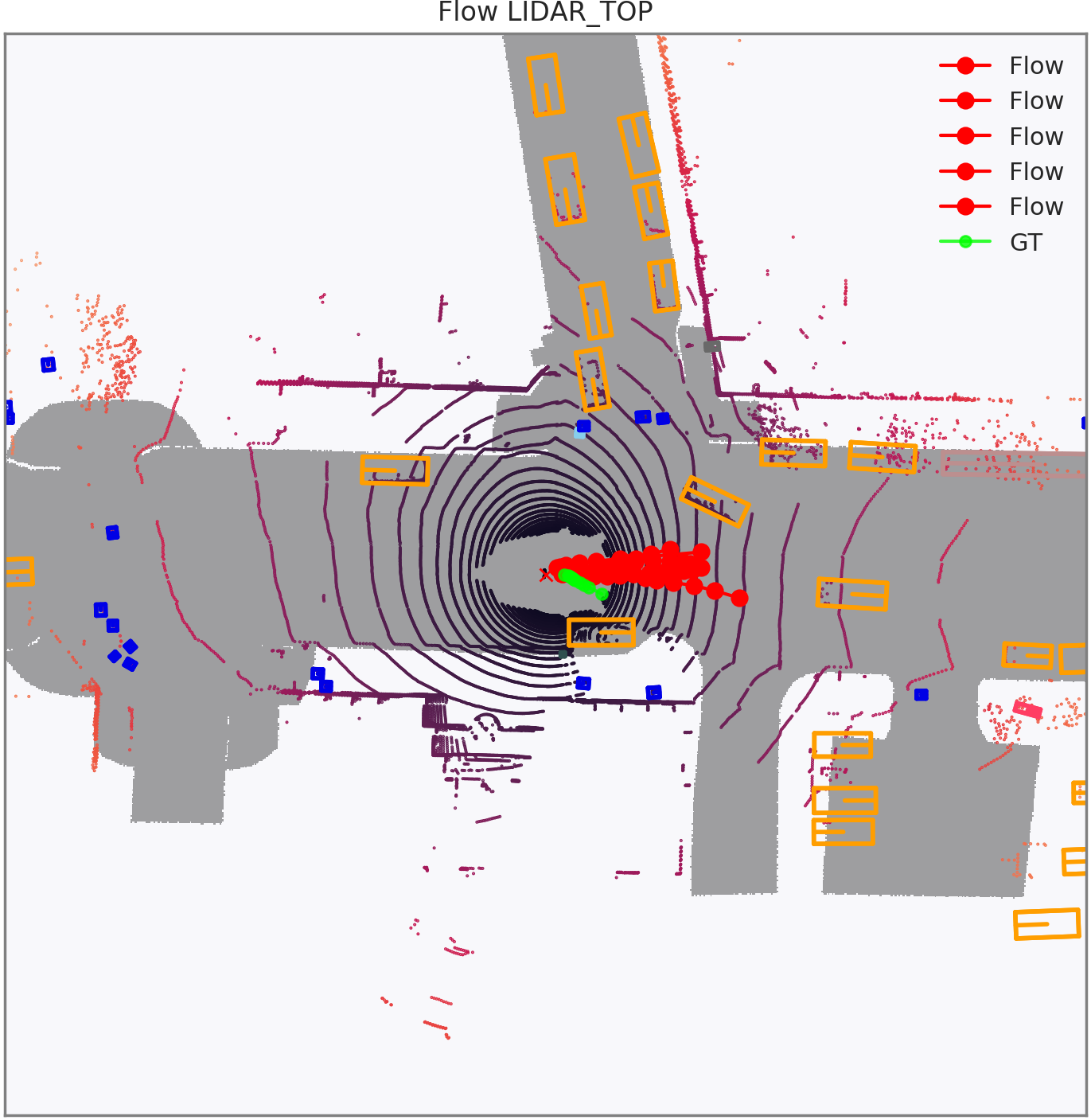}&
\includegraphics[width=.33\textwidth, trim={0 5cm 0 5cm}, clip=True]{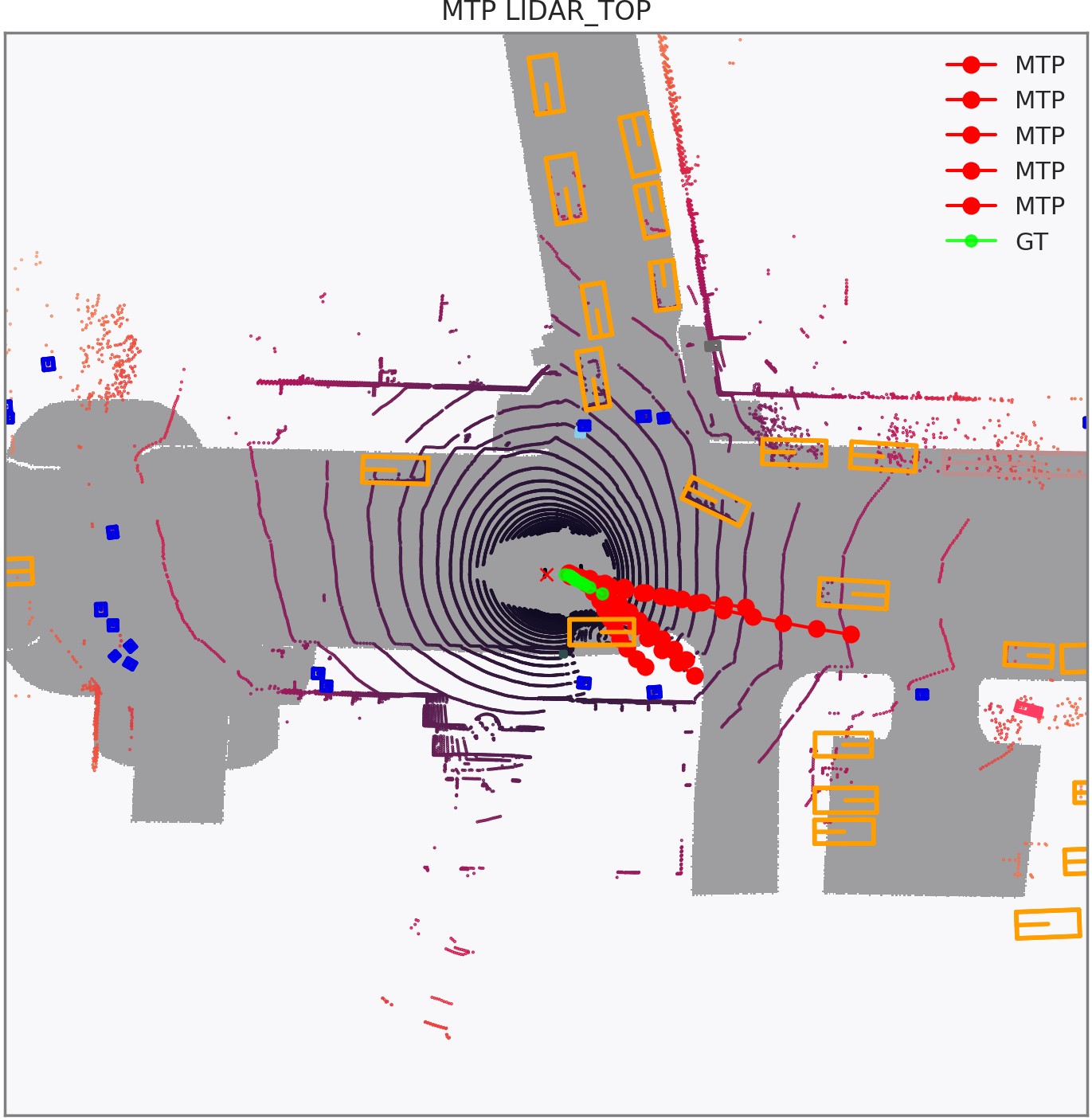}
\end{tabular}}
\vspace{-1cm}
\caption{Model trajectory forecasts at two separate frames in the same scene. $K=5$ predicted trajectories are shown in red, and the true recorded future trajectory from the dataset is shown in green. \textbf{LDS} predicts more diverse and plausible trajectories than both baselines.}
\label{fig:prediction-visualization}
\end{figure*}

\para{Forking Paths.} 
The training set results on ActEV/VIRAT for CAM-NF and various baseline models are included in Appendix \ref{appendix:actev-virat-results}. To summarize the training set results, we find CAM-NF effective on this dataset, only outperformed by the current state-of-art method Multiverse \cite{liang2020garden}, thus satisfying our goal of testing whether LDS can improve a strong backbone model. Now, we show the FP test results in Table \ref{table:LDS-ablation-loss-function} (Left). Note that the FP dataset comes with two different categories ``45-Degree'' and ``Top Down'' depending on the camera angle view of the human annotators. We report the average results between the two views for readability, and leave the full results split over each sub-category to Appendix \ref{appendix:forking-path-split}. We observe that CAM-NF already outperforms all prior methods on all metrics. With LDS, CAM-NF improves even further, outperforming all prior methods by a large margin. In comparison, DSF and DLow are not able to achieve the same level of performance boost, and in the case of DSF, the effect is even detrimental. The transductive variant LDS-TD-NN improves performance further on FDE metrics, while performing on par with LDS on ADE metrics; this results is promising as the transductive variant never observes the training set, and this dataset consists of a clear distributional shift between the training set and the testing set. In Appendix \ref{appendix:additional-analysis-forking-paths}, we provide additional analysis on Forking Paths results.

\begin{figure*}
\centering
\resizebox{\textwidth}{!}{
\begin{tabular}{ccc}
\textbf{LDS} (Ours) & \textbf{CAM-NF} & \textbf{Multiverse} \\ 
\includegraphics[width=.33\textwidth, trim={0 6cm 0 0}, clip=True]{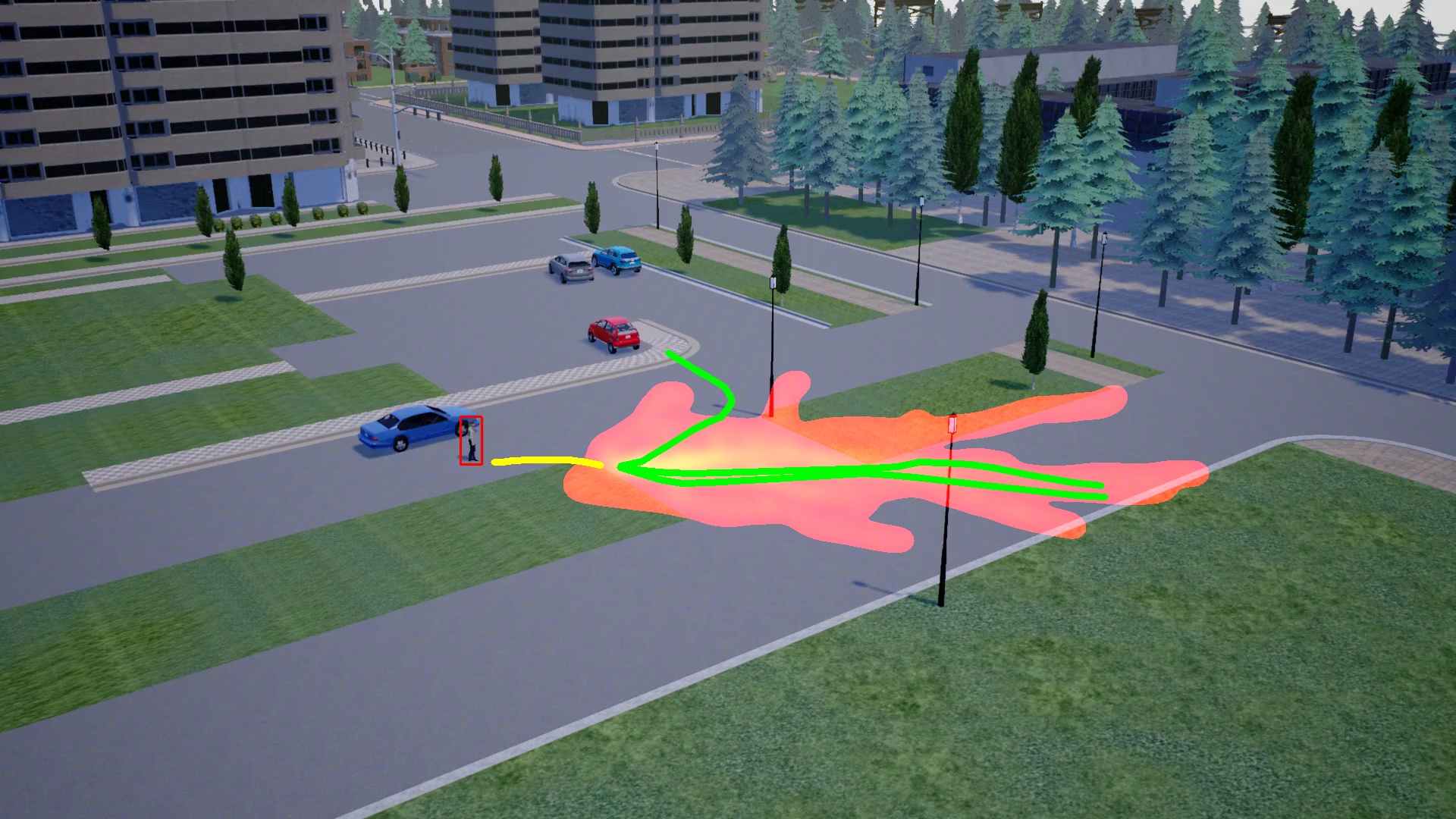}\hfill &
\includegraphics[width=.33\textwidth, trim={0 6cm 0 0}, clip=True]{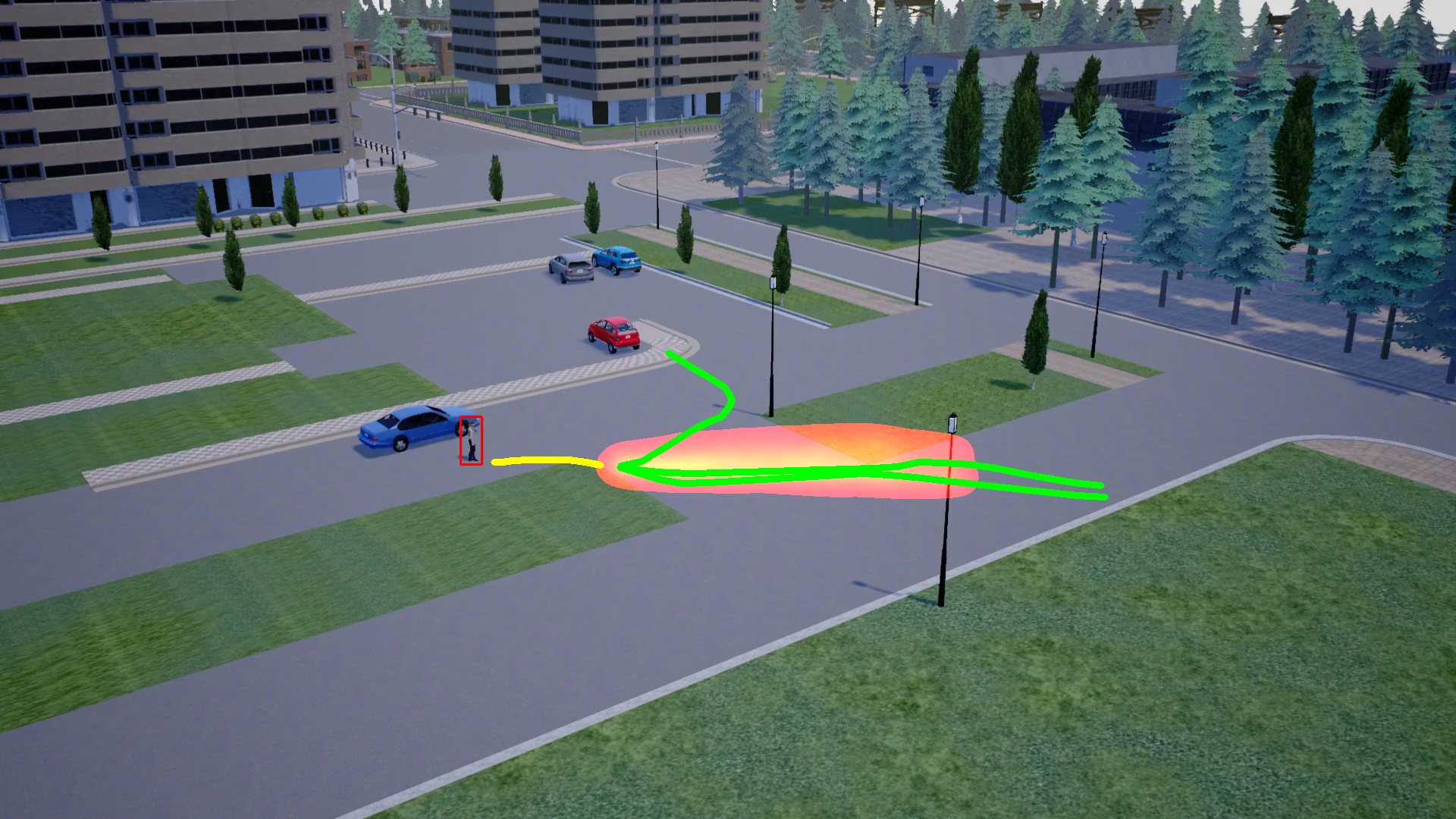}\hfill & 
\includegraphics[width=.33\textwidth, trim={0 6cm 0 0}, clip=True]{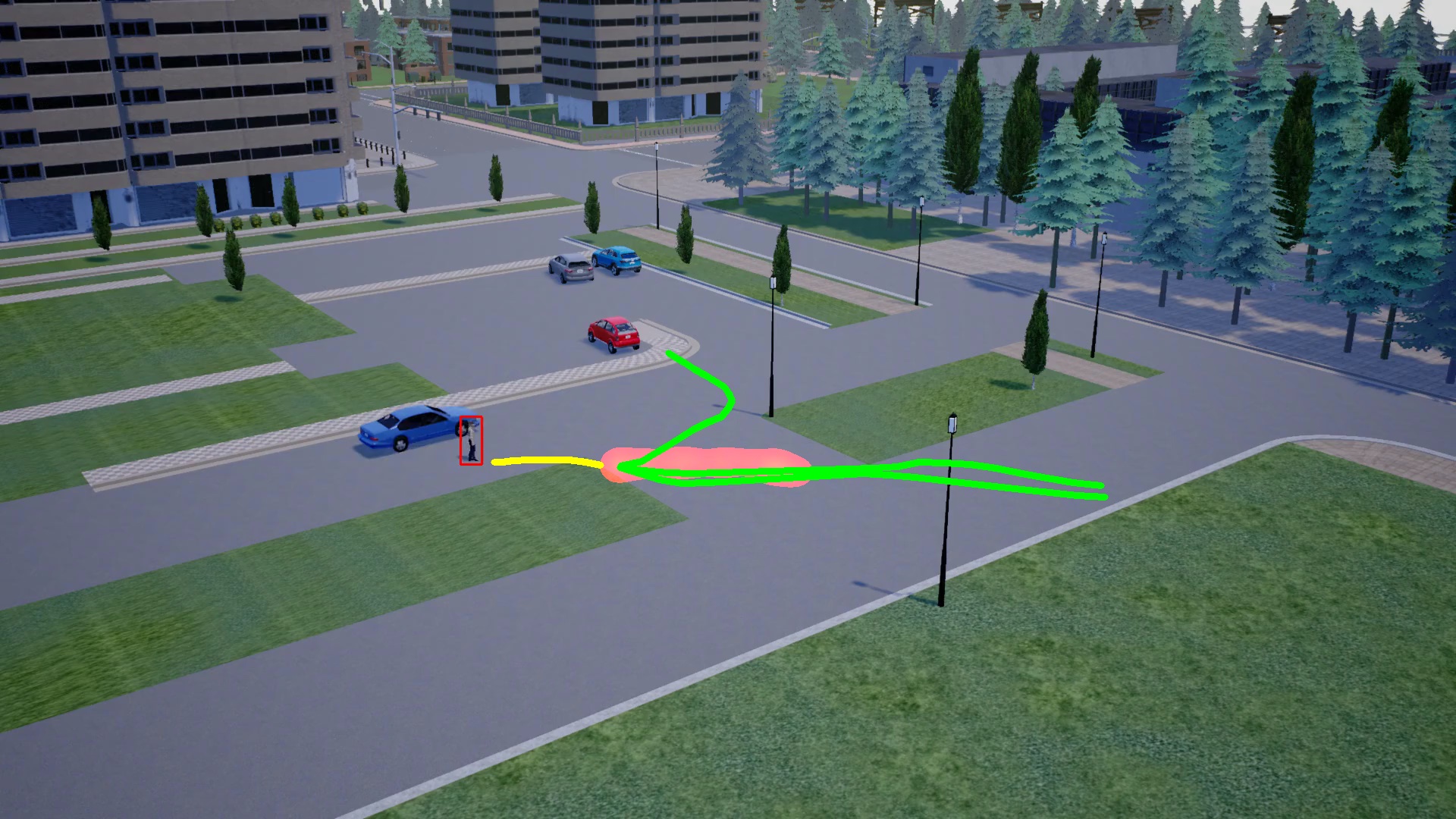}\hfill
\end{tabular}}
\caption{Visualization of predictions from various models on a single scene from the Forking Paths dataset. The red-yellow heatmap corresponds to the visited state density from the 20 predicted trajectories from each model; yellow indicates higher density. The green lines are the ground-truth human annotated futures. \textbf{LDS} produces diverse forecasts that cover the diverse futures, while the other two methods appear to have collapsed to a single output.}
\label{fig:forking-path-visualization}
\end{figure*}

\begin{table*}[]
\resizebox{\textwidth}{!}{
\begin{tabular}{|c|c|c|}
\hline 
{Method}& {$\text{minADE}_{20} (\downarrow)$} & {$\text{minFDE}_{20} (\downarrow)$} \\
\hline
\textbf{Linear}$^*$ &205.0 $\pm$ 0.0 & 388.0 $\pm$ 0.0 \\
\textbf{LSTM}$^*$ & 192.4 $\pm$ 2.2 & 368.3 $\pm$ 3.4 \\
\textbf{Social-LSTM}$^*$ \cite{alahi2016social}& 189.0 $\pm$ 1.7 & 363.7 $\pm$ 3.0 \\
\textbf{Social-GAN}$^*$ \cite{gupta2018social}& 179.9$\pm$ 4.3 & 334.4 $\pm$ 9.0 \\ 
\textbf{Next}$^*$ \cite{liang2019peeking} & 176.8 $\pm$ 2.4 & 343.4 $\pm$ 6.1 \\ 
\textbf{Multiverse}$^*$ \cite{liang2020garden} &
163.3 $\pm$ 2.3 & 325.2 $\pm$ 3.5 \\
\hline
\textbf{CAM-NF} \cite{park2020diverse} & 148.0 $\pm$ 2.3 & 293.6 $\pm$ 4.7 \\
\textbf{DSF} \cite{yuan2019diverse} & 162.8 $\pm$ 2.0 & 320.6 $\pm$ 3.6 \\ 
\textbf{DLow} \cite{yuan2020dlow} & 137.8 $\pm$ 5.9 & 273.4 $\pm$ 14.0 \\ 
\textbf{LDS} (Ours) & \textbf{98.6} $\pm$ 5.8 & \textbf{182.0} $\pm$ 14.5 \\
\textbf{LDS-TD-NN} (Ours) & \textbf{100.0} $\pm$ 3.2 & \textbf{178.1} $\pm$ 7.6 \\ 
\hline 
\end{tabular}
\quad 
\begin{tabular}{|c|cccc|}
\hline
Method/Metric
& $\text{mADE}_{\text{5}} (\downarrow)$ & $\text{mFDE}_{\text{5}} (\downarrow)$ & $\text{minASD}_{\text{5}}$ $(\uparrow)$ & $\text{minFSD}_{\text{5}}$ $(\uparrow)$ \\
\hline
Ours & \textbf{2.06} & \textbf{4.62} & 3.09 & 8.15 \\
\hline
Ours w.o. Diversity  & 5.95 & 14.63 & 0.16 & 0.37\\
Ours w.o. Likelihood  & 8.06 & 19.40 & \textbf{10.16} & \textbf{24.83}\\
\hline
Ours w. Rec & 2.16 & 4.89 & 3.29 & 9.00 \\
Ours w. meanDiv & 4.85 & 11.43 & 0.17 & 0.36 \\
\hline
Method/Metric &
$\text{mADE}_{10} (\downarrow)$ & $\text{mFDE}_{10} (\downarrow)$ & $\text{minASD}_{10}$ $(\uparrow)$ & $\text{minFSD}_{10}$ $(\uparrow)$ \\
\hline 
Ours & \textbf{1.65} & \textbf{3.50}  & 1.98 & 5.91 \\ 
\hline
Ours w.o. Diversity & 4.97 & 12.52 & 0.07 & 0.15\\
Ours w.o. Likelihood & 5.61 & 13.20 & \textbf{4.55} & \textbf{11.02}\\
\hline 
Ours w. Rec & 1.78 & 3.85 & 2.92 & 5.73 \\
Ours w. meanDiv & 4.92 & 11.66 & 0.03 & 0.05 \\ 
\hline
\end{tabular}}
\caption{\textbf{Left}: Forking Paths results. LDS-augmented CAM-NF significantly outperforms all other methods, including Multiverse and DLow-augmented CAM-NF. \textbf{Right}: LDS ablation results on NuScenes. In row 2-3, one of the LDS losses is removed, and performance significantly deteriorates. In row 4-5, one of the LDS losses is replaced by an alternative, and performance again declines sharply.}
\label{table:LDS-ablation-loss-function}
\end{table*}

\subsection{Qualitative Results}
\label{sec:qualitative-results}

Next, we illustrate trajectories from LDS and baselines in the two benchmarks to demonstrate that LDS indeed outputs more diverse and plausible trajectories. 

\para{NuScenes examples.} In Figure \ref{fig:prediction-visualization}, we show visualizations of two separate frames of the same nuScenes instance, overlaid with predictions from LDS-AF, AF, and MTP. Overall, LDS produces the most diverse and plausible set of trajectories in both frames. In the first frame, AF exhibits a failure mode as some of its predictions go far off the road. This provides evidence that sampling i.i.d.~from a vanilla flow model may fail to identify realistic trajectories. 
But when LDS is used to draw samples from the same flow model (i.e., LDS-AF), the trajectories become both more diverse and more realistic. In the second frame, MTP outputs a few trajectories that violate road constraints, while AF trajectories are concentrated in one cluster. Again, LDS-AF is the only model that predicts both diverse and plausible trajectories. In Appendix \ref{appendix:nuscenes-additional-visualizations}, we provide additional visualizations including LDS-AF-TD-NN trajectories (Figure \ref{fig:lds-af-vs-lds-af-td-nn}), as well as visualizations of different sets of trajectories LDS-AF outputs by varying the $\epsilon$ input to the model (Figure \ref{fig:epsilon-figures}). 

\para{Forking Paths examples.} Visualizations of LDS (on top of CAM-NF), CAM-NF, and Multiverse predictions on the FP test set are shown in Figure \ref{fig:forking-path-visualization}.
Additional visualizations are provided in Figure \ref{fig:forking-path-visualization-additional} in Appendix \ref{appendix:forking-paths-details}. Again, LDS outperforms the other two methods and is the only approach that comes close to covering the diverse ground-truth futures.

\subsection{Ablation Study}
\label{sec:ablation-studies}

We conduct an ablation study to understand the impact of the various design choices in LDS---specifically, the importance of the different parts of the LDS loss function to its empirical performance. To this end, we train four ablations of LDS-AF on nuScenes. The first two omit one of the terms in the LDS objective---i.e., one without the diversity loss (\textbf{Ours w.o.~Diversity}), and one without the likelihood loss (\textbf{Ours w.o.~Likelihood}). The latter two modify the two LDS loss terms: one replaces the NLL loss (Equation \ref{eq:LDS-likelihood-loss}) with DLow's Reconstruction$+$KL
losses (see Appendix \ref{appendix:nuscenes-details} for details) (\textbf{Ours w.~Rec}), and the other replaces the minimum in the diversity loss (Equation \ref{eq:LDS-diversity-loss}) with the mean (\textbf{Ours w.~meanDiv}).
All four of these models are trained using the same procedure as LDS. As shown in Table \ref{table:LDS-ablation-loss-function} (Right), the first two ablations significantly deteriorate performance. As expected, \textbf{Ours w.o.~Diversity} records close to zero diversity, and \textbf{Ours w.o.~Likelihood} achieves high diversity but at the cost of plausibility. Note that \textbf{Ours w.o.~Diversity} also performs poorly in terms of accuracy; this result shows that diversity is necessary to achieve good accuracy due to the stochastic nature of future trajectories. Thus, both terms in the LDS objective are integral to its success, and taking away either completely erases its benefits. 

Next, we find that \textbf{Ours w.~Rec} also reduces performance. This result demonstrates that leveraging the generative model's likelihood better captures the ground truth trajectory future. 
Finally, \textbf{Ours w.~meanDiv} significantly reduces overall performance---the prediction error increases two-fold while the diversity metrics collapse. This result demonstrates the importance of using the more robust minimum diversity metric compared to the mean diversity in the objective. In particular, the mean diversity does not penalize the degenerate case where most of the forecasts collapse to one trajectory, but one outlier forecast is very distant from the others. Together, these ablations all validate the key design choices in LDS. We include additional ablation studies assessing LDS's sensitivity to the pre-trained model, dependence on $\epsilon$, and its training stability in Appendix \ref{appendix:additional-ablation-results}.

Finally, we have set the number of modes $K$ based on predefined metrics for each dataset. However, this choice may not always be easy to make in practice. In general, a good strategy is to select a $K$ large enough and then discard samples based on ascending likelihood. Because LDS exhibits mode-seeking behavior, a large $K$ will likely ensure that the modes are included in the samples. Then, we can use likelihood as a reasonable proxy for the plausibility of each sample to guide the discarding process. In Appendix \ref{appendix:additional-ablation-results}, we illustrate an example of selecting a $K$ larger than the number of modes and discuss potential pitfalls.

\section{Conclusion}
We have proposed Diversity Sampling for Flow (LDS), a post-hoc learning-based diverse sampling technique for pre-trained generative trajectory forecasting models. LDS leverages the likelihood under the pre-trained generative model and a robust diversity loss to learn a sampling distribution that induces diverse and plausible trajectory predictions. Though intended for normalizing flow models, LDS is also compatible with VAEs, and independent of this choice, consistently achieves the best results compared to other sampling techniques and multi-modal models on two distinct forecasting benchmarks. Beyond its simple ``plug-in" improvement nature, through extensive ablation studies, we validate our method's design choices responsible for its effectiveness. Finally we introduce the transductive learning problem for trajectory forecasting, and show that LDS can be readily used to adapt to test instances on the fly and present a competitive solution to this novel problem setting.

\section*{Acknowledgements and Disclosure of Funding.} 

This work was supported by gift funding from GE Research and NEC Laboratories America, as well as NSF award CCF 1910769. The U.S. Government is authorized to reproduce and distribute reprints for Government purposes notwithstanding any copyright notation herein.

\clearpage 
{\small
\bibliographystyle{ieee_fullname}
\bibliography{references_iccv}
}

\clearpage 
\appendix 

\section{Normalizing Flow Models for Trajectory Forecasting}
\label{appendix:flow}
 In this section, we review some preliminaries on normalizing flow based trajectory forecasting models. We refer readers to \cite{papamakarios2019normalizing} for a comprehensive review of normalizing flows. 
 
 Normalizing flows learn a bijective mapping between a simple base distribution (e.g. Gaussian) and complex target data distribution through a series of learnable invertible functions. In this work, we denote the flow model as $f_{\theta}$, where $\theta$ represents its learnable parameters. The base distribution is a multivariate Gaussian $\textbf{Z} \sim \mathcal{N}(\textbf{0}, \textbf{I}) \in \mathbb{R}^{T \times 2}$, which factorizes across timesteps and $(x,y)$ coordinates. Then, the bijective relationship between $\textbf{Z}$ and $\textbf{S}$ is captured by the following forward and inverse computations of $f_{\theta}$: 
\begin{equation}
\textbf{S} =  f_{\theta}(\textbf{Z};\textbf{o}) \sim p_{\theta}(\textbf{S} \mid \textbf{o}), \quad \textbf{Z} =  f_{\theta}^{-1}(\textbf{S};\textbf{o}) \sim P_{\textbf{Z}} 
\end{equation}
We further impose the structural dependency between $\textbf{S}$ and $\textbf{Z}$ to be an invertible autoregressive function, $\tau_{\theta}$, between the stepwise relative \textit{offset} of the trajectory and the corresponding $\textbf{z}$ sample \cite{rhinehart2018deep, filos2020can}:
$$
\textbf{s}_t - \textbf{s}_{t-1} = {\tau}_{\theta}(\textbf{z}_t; \textbf{z}_{<t})
$$

The flow model can be trained using maximum likelihood. Because $\tau_{\theta}$ is autoregressive ($\textbf{z}_t$ does not depend on any $\textbf{z}_k$ where $k > t$), its Jacobian is a lower-triangular matrix, which admits a simple log-absolute-determinant form \cite{papamakarios2019normalizing}. The negative log-likelihood (NLL) objective is
\begin{equation} 
\label{eq:flow-objective}
\begin{aligned}
&- \log p_{{\theta}}(\textbf{S}; \textbf{o}) \\
=& - \log \Big( p(\textbf{Z})\big\lvert \text{det} \frac{\text{d} f_{\theta}}{\text{d} \textbf{Z}}\rvert_{\textbf{Z}= f_{\theta}^{-1}(\textbf{S}; \textbf{o})} \big\rvert^{-1} \Big) \\
=& - \Big(\sum^T \sum^D \log p(\textbf{Z}_{t,d}) - \sum^T \sum^D \log \bigg\rvert \frac{\partial \tau_{\theta}}{\partial \textbf{Z}_{t,d}}\bigg\rvert \Big)
\end{aligned}
\end{equation}
Once the model is trained, both sampling and \textit{exact} inference are simple. To draw one trajectory sample $\mathbf{S}$, we sample $\mathbf{Z} \sim P_{\mathbf{Z}}$ and compute $\mathbf{S} = f_{\theta}(\mathbf{Z};\mathbf{o})$
Additionally, the exact likelihood of \textit{any} trajectory $\textbf{S}$ under the model $f_{\theta}$ can be computed by first inverting $\textbf{Z} = f_{\theta}^{-1}(\textbf{S};\textbf{o})$ and then computing its transformed log probability via the change of variable formula, as in the second line of Equation \ref{eq:flow-objective}. 

\section{Toy Example Details}
\label{appendix:toy-example}
In the example given in Figure \ref{fig:toy-example}(a), we simulate a vehicle either going straight or turning right at the intersection. The vehicle obeys the Bicycle dynamics \cite{lavalle2006planning} and uses a MPC-based controller with intermediate waypoints to guide it to its goal. In each simulation, the vehicle starts behind the bottom entrance of the intersection with a fixed initial position perturbed by white noise, and the simulation ends when the vehicle reaches its goal. We collect $100$ simulations where the vehicle's final goal is the top exit of the intersection) and $900$ simulations where the vehicle's final goal is the right exit of the intersection. They combine into a training dataset of $1000$ trajectories. Each trajectory is of $100$ simulation steps. We downsample it by a factor of 10 and use the first two steps as the history input to the flow model and the task is to forecast the next 8 steps. With the first 2 steps ``burned" in, the vehicle is exactly at the bottom entrance of the intersection, creating a bi-modal dataset that proves to be difficult to model for a normalizing flow model. 

After the data collection, we train an autoregressive affine flow as in Appendix \ref{appendix:affine-flow-details} using the entire dataset until the log likelihood converges;  Then, we sample 100 times from the flow model using i.i.d unit Gaussian inputs to create trajectories as in Figure \ref{fig:toy-example}(b). We train a LDS model with $K=2$ on top of this flow model and generate $2$ samples using one Gaussian noise $\epsilon$ to generate the trajectories in Figure \ref{fig:toy-example}(c).

\section{LDS-CVAE Objective}
\label{appendix:LDS-cvae-objective}
As shown in our experiments, LDS can also be applied to CVAE models. Doing so requires changing the $\text{NLL}$ term in the LDS objective (Equation \ref{eq:LDS-full-loss} to $\text{ELBO}$. Because $\text{ELBO}$ is a lower bound of the true likelihood, optimizing for this modified objective would still achieve optimizing for the original objective. Formally, let $q_{\phi}(\mathbf{Z}|\mathbf{o})$ be the approximate latent posterior computed from the encoder and $p_{\theta}(\mathbf{S}|\mathbf{Z})$ be the likelihood computed from the decoder, we have
\begin{equation}
\label{eq:LDS-cvae-objective}
\text{L}_{\text{LDS-CVAE}}(\psi) \coloneqq \text{ELBO}(\psi) - \lambda_d \text{L}_{\text{d}}(\psi)
\end{equation}
where $\text{ELBO}(\psi) = \mathbb{E}_{\mathbf{Z} \sim q_{\phi}(\mathbf{Z}|\mathbf{o})} [\log p_{\theta}(\mathbf{S}|\mathbf{Z})] - \text{KL}(q_{\phi}(\mathbf{Z}|\mathbf{o})|p(\mathbf{Z}))$. The $\mathbf{Z}$s fed into CVAE are generated from $r_{\psi}(\epsilon;\mathbf{o})$, while the prior distribution is $p(\mathbf{Z}) \sim \mathcal{N}(0, \mathbf{I})$.
\section{Transductive LDS Algorithms}
\label{appendix:LDS-pseudocode}
\begin{minipage}{0.46\textwidth}
\begin{algorithm}[H]
\centering
\caption{LDS-TD-NN Training}\label{algo:LDS-td}
\begin{algorithmic}[1]
\STATE {\bfseries Input:} Flow $f_{\theta}$, Context $\mathbf{o}$
\STATE Initialize LDS model $r_{\psi}$
\FOR{i=1,...}
\STATE Sample $\mathbf{\epsilon} \sim \mathcal{N}(0, \mathbf{I})$
\STATE Compute  $\mathbf{Z}_1,..., \mathbf{Z}_K = r_{\psi}(\mathbf{\epsilon};\mathbf{o})$
\STATE Transform $f_{\theta}(\mathbf{Z}_1),.., f_{\theta}(\mathbf{Z}_K)$
\STATE Compute losses using Equations \ref{eq:LDS-likelihood-loss} \& \ref{eq:LDS-diversity-loss}
\STATE Update $\psi$ w.r.t gradient of Equation \ref{eq:LDS-full-loss}
\ENDFOR 
\STATE Compute  $\mathbf{Z}_1,..., \mathbf{Z}_K = r_{\psi}(\mathbf{\epsilon};\mathbf{o})$
\STATE {\bfseries Output:} $\mathcal{S} = f_{\theta}(\mathbf{Z}_1),.., f_{\theta}(\mathbf{Z}_K)$
\end{algorithmic}
\end{algorithm}
\end{minipage}

\begin{minipage}{0.46\textwidth}
\begin{algorithm}[H]
\centering
\caption{LDS-TD-P Training}\label{algo:LDS-td-p}
\begin{algorithmic}[1]
\STATE {\bfseries Input:} Flow $f_{\theta}$, Context $\mathbf{o}$
\STATE Randomly initialize $\mathbf{Z}_1,..., \mathbf{Z}_K \sim \mathcal{N}(0, \mathbf{I})$
\FOR{i=1,...}
\STATE Transform $f_{\theta}(\mathbf{Z}_1),.., f_{\theta}(\mathbf{Z}_K)$
\STATE Compute losses using Equations \ref{eq:LDS-likelihood-loss} \& \ref{eq:LDS-diversity-loss}
\STATE Update $\mathbf{Z}_1,..., \mathbf{Z}_K$ w.r.t gradient of Equation \ref{eq:LDS-full-loss}
\ENDFOR 
\STATE {\bfseries Output:} $\mathcal{S} = f_{\theta}(\mathbf{Z}_1),.., f_{\theta}(\mathbf{Z}_K)$
\end{algorithmic}
\end{algorithm}
\end{minipage}
\section{DLow \& DSF Details} 
\label{appendix:dlow}
Our implementations of DLow and DSF utilizes the same architecture as LDS. The main difference is the loss functions of the two methods. The DLow objective includes three terms:
\begin{equation}
    \label{eq:dlow-loss}
    \begin{aligned}
    &\text{Reconstruction Loss}: E_r(\hat{\textbf{s}}) = \min_{k\in K} \norm{\hat{\textbf{s}}_k-\textbf{s}}^2 \\ 
    &\text{Diversity Loss}: E_d(\hat{\textbf{s}}) = \frac{1}{K(K-1)}\sum_{i\neq j \in K} \text{exp}\Big(-\frac{\norm{\hat{\textbf{s}}_i - \hat{\textbf{s}}_j}^2}{\sigma_d} \Big) \\ 
    &\text{KL Loss}: L_{\text{KL}}(\textbf{z}) = \sum^K_{k=1} \text{KL}(p_{\psi}(\textbf{z}_k|\textbf{o})||p(\textbf{z}_k))
    \end{aligned}
\end{equation}
and the whole objective is:
$$
L_{\text{DLow}}(\psi) = \lambda_r E_r + \lambda_d E_d + \lambda_{\text{KL}} L_{\text{KL}}
$$
We tune the hyperparameters of DLow and find the following setting to work the best: $\lambda_d = 0.5, \lambda_r = 1, \lambda_{\text{KL}}=1$, and $\sigma_d = 1.$

DSF is a bit more involved. First, suppose that given input $\mathbf{o}$, DSF network $\psi$ learns a set of latent samples $\mathbf{z}_1,...,\mathbf{z}_K$, and the flow model decodes them to $\mathbf{s}_1,...,\mathbf{s}_K$. DSF constructs a deterministic point process (DPP) kernel $\mathbf{L} = \text{Diag}(\mathbf{r}) \cdot \mathbf{S} \cdot \textbf{Diag}(\mathbf{r})$, where $\mathbf{S}_{ij} = \text{exp}\big(-k \cdot \norm{\mathbf{s}_i-\mathbf{s}_j}^2\big)$ for some $k>0$. Each entry in the quality vector $\mathbf{r}$ is defined as $r_i = \omega \text{exp}\big(-\mathbf{z}_i^{\top} \mathbf{z}_i + R^2\big)$ if $\norm{\mathbf{z}_i} \leq R$; otherwise, $r_i = \omega$. $R$ is chosen as the $x$-th quantile of the chi-squared distribution with degree of freedom equal to the dimension of $\mathbf{z}_i$. Finally, DSF objective is $L_{\text{DSF}}(\psi) = -\text{trace}(\mathbf{I}-(\mathbf{L}(\psi)+\mathbf{I})^{-1})$, and stochastic gradient is backpropagated with respect to DSF parameters $\psi$. We choose $k=1$ and $x=90$ to be consistent with the original work; however, we find DSF to be sensitive to these hyperparameter choices and fail to scale to large-dimension tasks (e.g. Forking Paths).

\section{NuScenes Experimental Details}
\label{appendix:nuscenes-details}
\subsection{Dataset Details}
NuScenes \cite{caesar2020nuscenes} is a large urban autonomous driving dataset. The dataset consists of instances of vehicle trajectories coupled with their sensor readings, such as front camera images and lidar scans. The instances are further collected from $1000$ distinct traffic scenes, testing forecasting models' ability to generalize. Following the official dataset split provided by the nuScenes development kit, we use $32186$ instance for training, $8560$ instances for validation, and report results on the $9041$ instances in the test set. 

\subsection{Model Inputs}
\label{appendix:nuscenes-model-inputs}
\textbf{Model Inputs}. All models we implement (AF, CVAE baseline models and MTP) accept the same set of contextual information
\begin{align*}
\textbf{o}=\{\text{Lidar scans, velocity, acceleration, yaw}\}
\end{align*}
of the predicting vehicle at time $t=0$. Below we visualize an example Lidar scan and its histogram version \cite{rhinehart2019precog} that is fed into the models. 
\fboxrule=2pt
   \begin{figure}[H]
   \resizebox{\columnwidth}{!}{
 \fbox{\includegraphics[width=4cm]{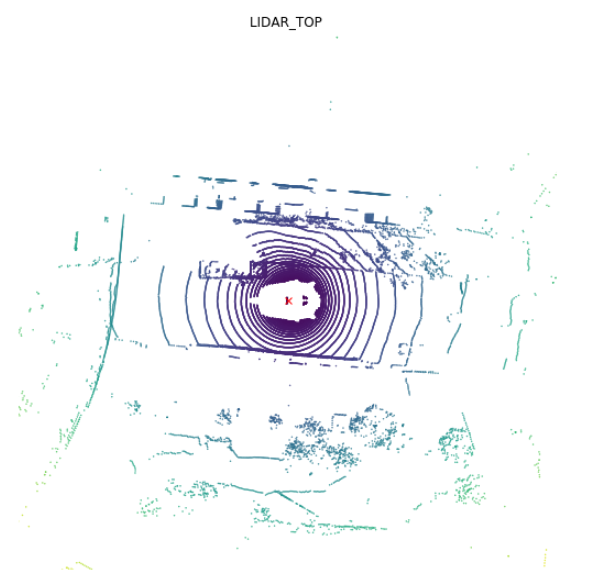}}
 }
  \caption{LiDAR inputs in nuScenes.}
 \label{fig:nuscenes-lidar}
\end{figure}
The Lidar scans are first processed by a pre-trained MobileNet-v128 \cite{howard2017mobilenets} to produce visual features. These features, concatenated with the rest of the raw inputs, are passed through a neural network to produce input features for the models.

\subsection{Autoregressive Affine Flow Details}
\label{appendix:affine-flow-details}
Our architecture is adapted from the implementation\footnote{\url{https://github.com/OATML/oatomobile/blob/alpha/oatomobile/torch/networks/sequence.py}} provided in \cite{filos2020can}. Here, we describe it in high level and leave the details to the architecture table provided below. 
AF consists of first a visual module that transforms the observation information $\textbf{o}$ into a feature vector $\textbf{h}_0$. Then, $\textbf{h}_0$ is processed sequentially through a GRU network \cite{chung2014empirical} to produce the per-step  \textit{conditioner} $\textbf{h}_t$ of the affine transformation: $\textbf{h}_t = \text{GRU}(\textbf{s}_t, \textbf{h}_{t-1})$. Finally, we train a neural network (MLP) on top of $\textbf{h}_t$ to produce the modulators $\mu, \sigma$ of the affine transformation: 
\begin{equation}
    \label{eq:flow-autoregressive}
    \begin{aligned}
    \textbf{s}_t -\textbf{s}_{t-1} &= {\tau}_{\theta}(\textbf{z}_t; \textbf{z}_{<t}) \\ &=\underbrace{\mu_{\theta}(\textbf{s}_{1:t-1}, \phi)}_{\text{MLP}_1(\textbf{h}_t)} + \underbrace{\sigma_{\theta}(\textbf{s}_{1:t-1},\phi)}_{\text{EXP}\Big(\text{MLP}_2(\textbf{h}_t)\Big)} \textbf{z}_t
    \end{aligned}
\end{equation}
\begin{table}[H]
  \caption{AF Architecture Overview}
  \label{table:LDS-architecture}
  \centering
  \begin{tabular}{ll}
    \toprule
    \cmidrule(r){1-2}
    Attributes & Values \\
    \midrule
    Visual Module & MobileNet(200 $\times$ 200 $\times$ 3, 128)  \\
    & Linear(128+3,64) \\
        & Linear(64,64) \\
            & Linear(64,64) \\
            \midrule 
    Autoregressive Module & GRUCell(64) \\ 
    \midrule
    MLP Module & ReLU $\circ$ Linear(64,32) \\ 
    & Linear(32, 4) \\
    \midrule 
    Base Distribution & $\mathcal{N}(0, \textbf{I})$ \\
    \bottomrule
  \end{tabular}
\end{table}

\subsection{CVAE Details}
Our CVAE implementation is adapted from the implementation\footnote{\url{https://github.com/Khrylx/DLow/blob/master/models/motion_pred.py}} in \cite{yuan2020dlow}. It takes the same set of inputs as our AF model except the addition of a one-frame history input. The history is encoded using a GRU network of hidden size $64$ to produce $h_0$, which is then concatenated with the rest of the inputs. This concatenated vector is then encoded through a 2-layer fully-connected network. To encode the future, our CVAE model uses a GRU network of the same architecture as the GRU encoder for the history. Finally, the encoded input and output (i.e. future) is concatenated and passed through another 2-layer network to give the mean and the variance of the approximate posterior distribution. For the decoder, we first sample a latent vector $z$ using the reparameterization trick. Then, $z$ is concatenated with the encoded inputs to condition the per-step GRU roll-out of the reconstructed future. The model is trained to maximize ELBO.
\begin{table}[H]
  \caption{CVAE Architecture Overview}
  \label{table:cvae-architecture}
  \centering
  \begin{tabular}{ll}
    \toprule
    \cmidrule(r){1-2}
    Attributes & Values \\
    \midrule
    History Encoder & GRU(2, 64)\\ 
    Visual Module & MobileNet(200 $\times$ 200 $\times$ 3, 128)  \\
    \midrule 
    Full Input Encoder& Linear(128+64+3, 64) \\
        & Linear(64, 64) \\
            & Linear(64, 64) \\
            \midrule 
    Full Output Encoder & GRU(2, 64) \\
    \midrule 
    Input Output Merger & Linear(64+64, 64) \\
    & Linear(64, 32+32) \\ 
    $\mu, \sigma$ & $\mathbb{R}^{32}, \mathbb{R}^{32}$ \\
    \midrule
    Decoder & GRU(2+32+64, 64) \\ 
    & Linear(64, 64) \\
    & Linear(64, 32, 2) \\
    \bottomrule
  \end{tabular}
\end{table}

\subsection{LDS Architecture Details} 
\label{appendix:nuscenes-LDS-details}
LDS $r_{\psi}$ is a single multi-layer neural network with $K$ heads, the number of modes pre-specified. To ensure stable training, we clip the diversity loss to be between $[0,40]$ for $K=5$ and $[0,30]$ for $K=10$. DSF and DLow use the same architecture as LDS. 
\begin{table}[H]
  \caption{LDS Architecture and Hyperparameters Overview}
  \label{table:LDS-architecture}
  \centering
  \begin{tabular}{ll}
    \toprule
    \cmidrule(r){1-2}
    Attributes & Values \\
    \midrule
    LDS Architecture & Linear(Input Size, 64)  \\
    & Linear(64,32) \\
    & Linear(32, 2 $\times$ T $\times$ K) \\
    \midrule
    Learning Rate & 0.001 \\ 
    $\lambda_d$ & 1 \\ 
    Diversity function clip value & 40/30 \\ 
    \bottomrule
  \end{tabular}
\end{table}

\subsection{Training Details}
\label{appendix:training-details}
  
We train the ``backbone" forecasting models AF, CVAE, MTP-Lidar for $20$ epochs with learning rate $10^{-3}$ using Adam \cite{kingma2014adam} optimizer and batch size $64$.  LDS-AF iterates through the full training set once, while LDS-AF-TD directly optimizes on the test set with a minibatch size of $64$ and $400$ adaptation iterations for every minibatch. For all LDS models, we set $\lambda_d=1$ and do not experiment with further hyperparameter tuning. LDS training also uses Adam. For all models, we train $5$ separate models using random seeds and report the average and standard deviations in our results.

\subsection{Additional Results}
\label{appendix:nuscenes-additional-results}
\para{Comparison with Trajectron++} 
To test the limit of our approach, we additionally compare against Trajectron++ \cite{salzmann2020trajectron++}. Different from other baselines we include in the main paper, Trajectron++ utilizes additional inputs such as spatio-temporal graph and vehicle/pedestrian dynamics and evaluates on a shorter horizon (i.e. 8 frames). Here, we re-train an AF model by truncating all trajectories to $8$ frames, and then train a LDS model on top, giving us LDS-AF-8. Trajectron++ only reports $\text{minFDE}_{10}$ and obtains $2.24$. LDS-AF-8 obtains $\text{minFDE}_{10}:2.28, \text{minADE}_{10}:1.09, \text{minASD}_{10}:2.18, \text{minFSD}_{10}:5.80$, where we include the other three metrics for completeness. As shown, LDS-AF-8 is slightly worse than Trajectron++ on $\text{minFDE}_{10}$; however, giving that the backbone model LDS utilizes here is a much weaker model than Trajectron++. this result is encouraging and suggests that even a weak model can produce competitive predictions when it is augmented with an effective sampling mechanism. 

\para{minASD vs. meanASD.} Here, we illustrate the robustness of minASD compared to meanASD. Consider the following two sets of 10 predictions. In the first set, the pairwise distance between all pairs is exactly 1. In the second set, 9 predictions are identical, but their distance to the remaining one is 100. The first set achieves identical diversity value 1 under the two metrics, whereas the second set achieves 0 minASD but $\frac{20}{9}$ meanASD. Therefore, we might incorrectly conclude that the second set is more diverse if we were to solely rely on mean metrics for diversity.

\para{Mean diversity results.} Here, we report the meanASD and meanFSD metrics for diversity. As shown, LDS still achieves the highest diversity on these metrics and provide greater diversity boost than DLow; however, the relative differences among models are much smaller. Additionally, by examining the tables from $K=5$ to $K=10$, we no longer find the pattern that LDS being the only model whose diversity does not deteriorate as we did in Table \ref{table:nuscenes-accuracy} (Right). This is because the mean metric only captures the average behavior and not the worst case behavior. Thus, our hypothesis that the min metrics are more informative than the mean metrics are supported by the following results. 
\begin{table}[H]
\resizebox{\columnwidth}{!}{
\begin{tabular}{llll}
\toprule
Method & Samples & meanASD $(\uparrow)$ & meanFSD $(\uparrow)$\\
\midrule 
MTP-Lidar-5 & 5 & 5.74 $\pm$ 0.79 & 13.80 $\pm$ 2.00\\
CVAE & 5 & 5.38 $\pm$ 0.09 & 12.28 $\pm$ 0.16 \\
DLow-CVAE & 5 & 6.66 $\pm$ 0.21 & 15.43 $\pm$ 0.44 \\
AF & 5 & 6.21 $\pm$ 0.02 & 14.48 $\pm$ 0.04\\
DLow-AF-5 & 5 & \textbf{7.41} $\pm$ 0.29 & \textbf{17.90} $\pm$ 0.67 \\
LDS-AF-5 & 5 & \textbf{7.89} $\pm$ 0.29 & \textbf{19.06} $\pm$ 0.58 \\
\bottomrule 
\end{tabular}}
\end{table}
\begin{table}[H]
\resizebox{\columnwidth}{!}{
\begin{tabular}{llll}
\toprule 
Method & Samples & meanASD $(\uparrow)$ & meanFSD $(\uparrow)$ \\
\midrule
MTP-Lidar-10 & 10 & 5.24 $\pm$ 0.30 & 12.71 $\pm$ 0.59 \\
CVAE & 10 & 5.38 $\pm$ 0.10 & 12.28 $\pm$ 0.18 \\ 
DLow-CVAE & 10 &  6.96 $\pm$ 0.20 & 16.16 $\pm$ 0.45 \\
AF & 10 & 6.21 $\pm$ 0.01 & 14.48 $\pm$ 0.04 \\
DLow-AF-10 & 10 & \textbf{7.88} $\pm$ 0.57 & \textbf{19.49} $\pm$ 1.36  \\ 
LDS-AF-10 & 10 & \textbf{7.90}  $\pm$ 0.28 & \textbf{19.71} $\pm$ 0.74 \\
\bottomrule 
\end{tabular}}
\caption{NuScenes prediction mean diversity results.}
\label{table:nuscenes-mean-diversity}
\end{table}

\subsection{Additional Ablation Results}
\label{appendix:additional-ablation-results}

In this section, we provide some additional ablation studies to further understand the effectiveness of LDS. First,
we aim to understand: \textbf{how sensitive is LDS to the quality of the underlying flow model?} To answer this question, we train an additional AF model with half the number of epochs as the original one ($\textbf{AF}^{-}$), and then train LDS as before. The comparisons are shown in Table \ref{table:LDS-ablation-backbone}. With only half the training time, $\text{AF}^{-}$ performs considerably worse than $\text{AF}$, yet LDS is still able to provide a significant performance boost, achieving $\textbf{33}\%$ reduction in both $\text{minADE}$ and $\text{minFDE}$. This reduction is greater than that of LDS applied to the stronger AF model ($27$\%), suggesting the utility of LDS is greater for weaker pre-trained model and highlighting that its overall effectiveness is robust to the quality of the underlying flow.

\begin{table}[H]
\resizebox{\columnwidth}{!}{
\begin{tabular}{llllllllll}
\toprule
Method & S & $\text{mADE}_5$ &  $\downarrow$\% & $\text{mFDE}_5$ & $\downarrow$\% & minASD & $\uparrow$\% & minFSD & $\uparrow$\%\\
\midrule 
$\text{AF}^{-}$ & 5 & 3.49 $\pm$ 0.16 & - & 7.79 $\pm$ 0.41 & - & 1.99 $\pm$ 0.15 & - & 4.58 $\pm$ 0.46& - \\
LDS-$\text{AF}^{-}$ & 5 & 2.31 $\pm$ 0.19 & $\textbf{34}$\% & 5.17 $\pm$ 0.39 & $\textbf{34}$\% & 2.91 $\pm$ 0.05 & 146\%& 8.25 $\pm$ 0.32 & 180\%\\ 
LDS-$\text{AF}^{-}$-TD & 5 & 2.35 $\pm$ 0.16 & $\textbf{33}$\% & 5.24 $\pm$ 0.33 & $\textbf{33}$\% & 3.00 $\pm$ 0.16 & 151\%& 8.36 $\pm$ 0.14 & 183\%\\
\midrule
AF & 5 & 2.86 $\pm$ 0.01 & -&  6.26 $\pm$ 0.05 & - & 1.58 $\pm$ 0.02 & - & 3.75 $\pm$ 0.04 & -\\ 
LDS-AF & 5 & 2.06 $\pm$ 0.09 & 28\%& 4.67 $\pm$ 0.25 & 25\% & 3.13 $\pm$ 0.18 & 98\% & 8.19 $\pm$ 0.26 & 118\%\\
LDS-AF-TD & 5 & 2.06 $\pm$ 0.02 & 28\%& 4.62 $\pm$ 0.07&  26\% & 3.09 $\pm$ 0.07 & 95\%& 8.15 $\pm$ 0.17 & 117\% \\ 
\bottomrule 
\end{tabular}}
\caption{LDS ablations on the pre-trained flow models.}
\label{table:LDS-ablation-backbone}
\end{table}

\begin{figure*}
  \centering
 \includegraphics[width=.33\textwidth]{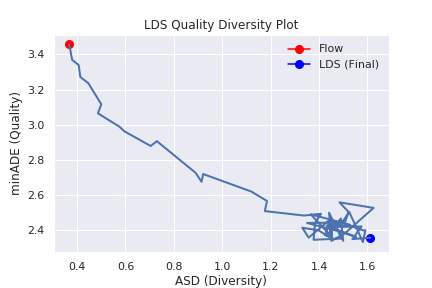}\hfill 
 \includegraphics[width=.33\textwidth]{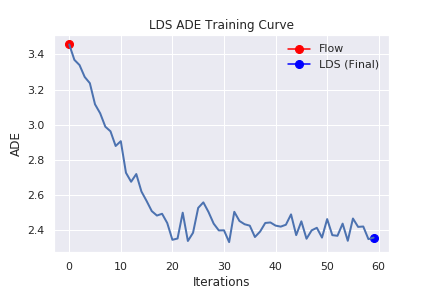}\hfill 
 \includegraphics[width=.33\textwidth]{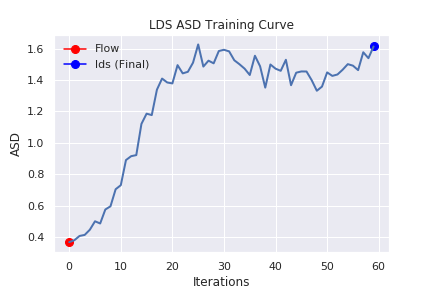}\hfill 
  \caption{\textbf{Left}: LDS (ASD, ADE) plot on nuScenes mini. LDS offers stable and fast improvements to flow outputs in both accuracy and diversity during its training. \textbf{Middle}: LDS ADE over the course of training. \textbf{Right}: LDS minASD over the course of training.}
 \label{fig:quality-diversity-plot}
\end{figure*}

\paragraph{LDS training stability.} We track the state of minADE and minASD on the mini nuScenes validation set over the course of LDS training. On the mini version of the nuScenes dataset, we train a LDS-AF ($K=5$) model, and for every training iteration, we compute the current LDS-AF model's minADE and minASD on the mini validation set. The mini version is a much smaller dataset with only $~1000$ instances, making this procedure manageable. The sequence of (ASD, ADE) pairs are traced as a trajectory on a 2D plane, where the x-axis corresponds to $\text{minASD}_5$ and the y-axis corresponds to $\text{minADE}_5$.

The entire trajectory is visualized in Figure \ref{fig:quality-diversity-plot}. We additionally visualize both minADE and minASD individually over the course of training in the middle and the right panels. In all three plots, the initial AF model's (ASD, ADE) point is colored in red, and the final LDS-AF's (ASD,ADE) point is colored in blue. Focusing on the left panel, we note that the initial point at the top left corner represents the ADE/ASD of the pre-trained flow backbone, which has relatively high error and low diversity. However, LDS quickly discovers good sampling distribution and offers fast and near ``monotonic" improvements along both axis. This provides evidence that the joint objective is effective and helps avoiding potential local minima in the loss landscape. The rapid improvement early on in training also explains why LDS's transductive adaptation procedure can be effective. Finally, an early stopping mechanism may be effective given the fast convergence to the optimum (i.e. the bottom right corner); we leave it to future work to investigate the precise stopping criterion. 

\paragraph{Dependence on $\epsilon$.} One may eliminate the dependence on $\epsilon$ by making the mapping procedure deterministic: $\{\mathbf{Z}_1,...,\mathbf{Z}_K\} = r_{\psi}(\mathbf{o})$. In theory, this simplified formulation should optimize for the same objective as the objective itself does not explicitly depend on $\epsilon$. However, in practice, we find this ablation reduces performance as the trained LDS model may be overfitting to some deterministic set of trajectories to multiple different inputs. We report the average results over $5$ seeds on LDS-AF for $K=10$ where we do not utilize the $\epsilon$ sampling procedure (Step 3 and 4 in Algorithm \ref{algo:LDS-full}): $\text{minADE}_{10}: 1.74, \text{minFDE}_{10}: 3.73, \text{minASD}_{10}: 2.06, \text{minFSD}_{10}: 6.03$. These results are slightly worse than the original LDS-AF results reported in Table \ref{table:nuscenes-accuracy}. Therefore, we confirm that adding innate stochasticity to the training procedure with $\epsilon$ boosts performance, validating our original formulation.

\begin{figure}[H]
   \resizebox{\columnwidth}{!}{
 \includegraphics[width=5cm]{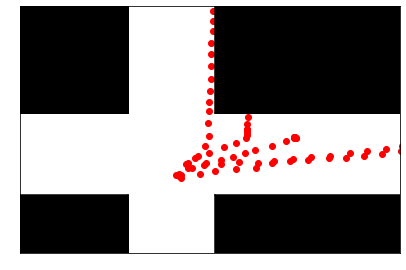}}
  \caption{LDS outputs for $K=5$.}
 \label{fig:toy-example-failure}
\end{figure}

\paragraph{Mismatched K.} Here, we perform a controlled experiment analyzing LDS outputs when the number of samples $K$ it learns to output mismatches the number of modes in the underlying distribution. To do this, we return to our toy intersection environment as in Figure \ref{fig:toy-example}(a). Instead of training a LDS with $K=2$ as done in Figure \ref{fig:toy-example}(c), we train one with $K=5$. This LDS's set of 5 samples are shown in Figure \ref{fig:toy-example-failure}. Among the 5 trajectories, the two modes are still captured. Importantly, the major mode (turning right) is captured twice. However, the remaining two trajectories represent the average of the two modes and correspond to potentially unrealistic behaviors. Note that this set of five trajectories would achieve very low minADE errors in both single and multiple-future evaluations, since at least one trajectory in the predictions is close to both modes. However, for planning settings, this set of predictions may not be optimal. Hence, choosing $K$ carefully is important in practice, and we leave it to future work for further investigations. 

\subsection{Additional Visualizations} 
\label{appendix:nuscenes-additional-visualizations} 
\begin{figure*}
\centering
\resizebox{\textwidth}{!}{
\begin{tabular}{cccc}
&\textbf{LDS} (Ours) & \textbf{AF} & \textbf{MTP} \\ 
\rotatebox[]{90}{$\quad \quad \quad \quad \quad \textbf{Scene 1}$} &
\includegraphics[width=.33\textwidth, trim={0 4cm 0 4cm}, clip=True]{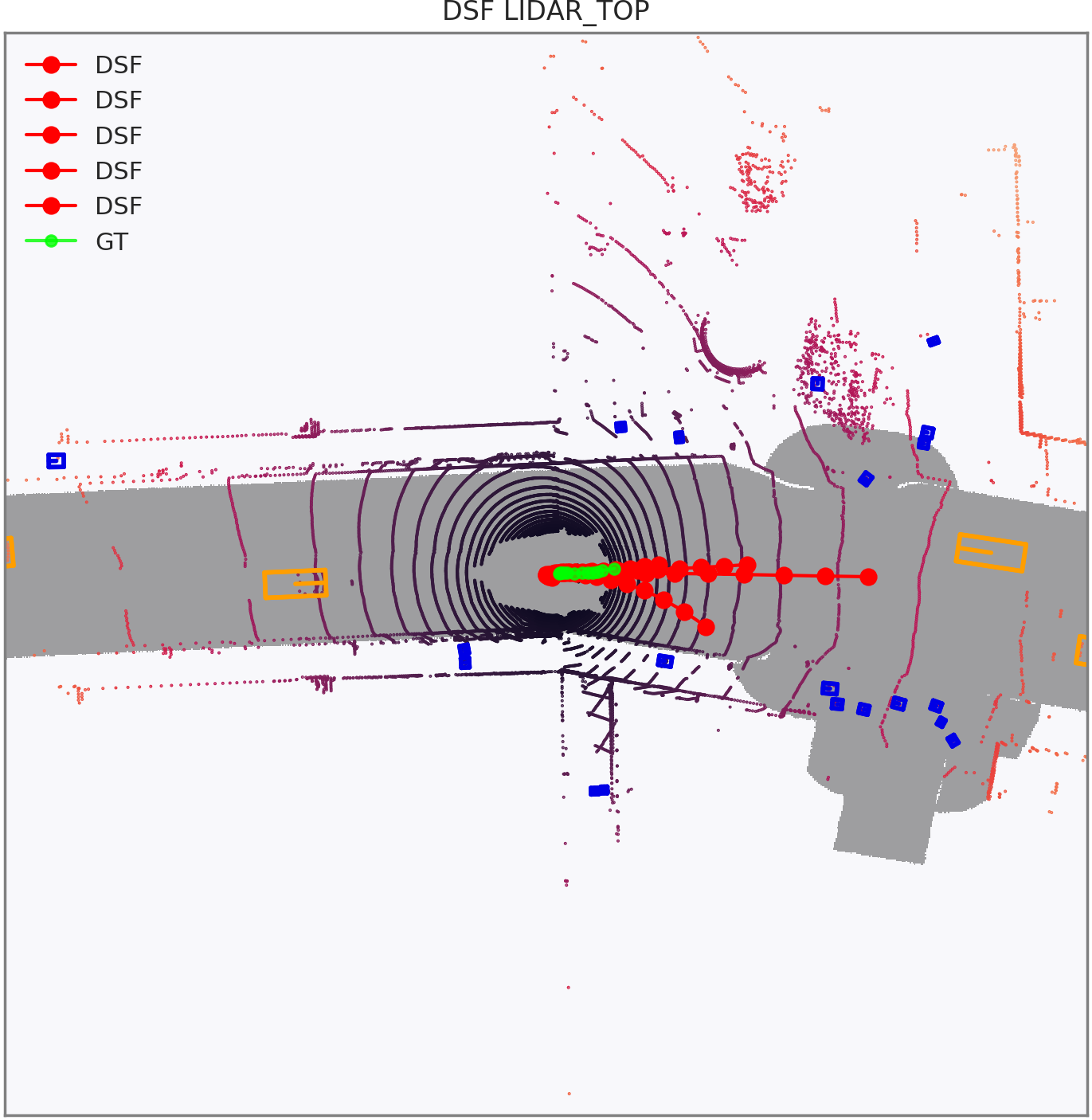} &
\includegraphics[width=.33\textwidth, trim={0 4cm 0 4cm}, clip=True]{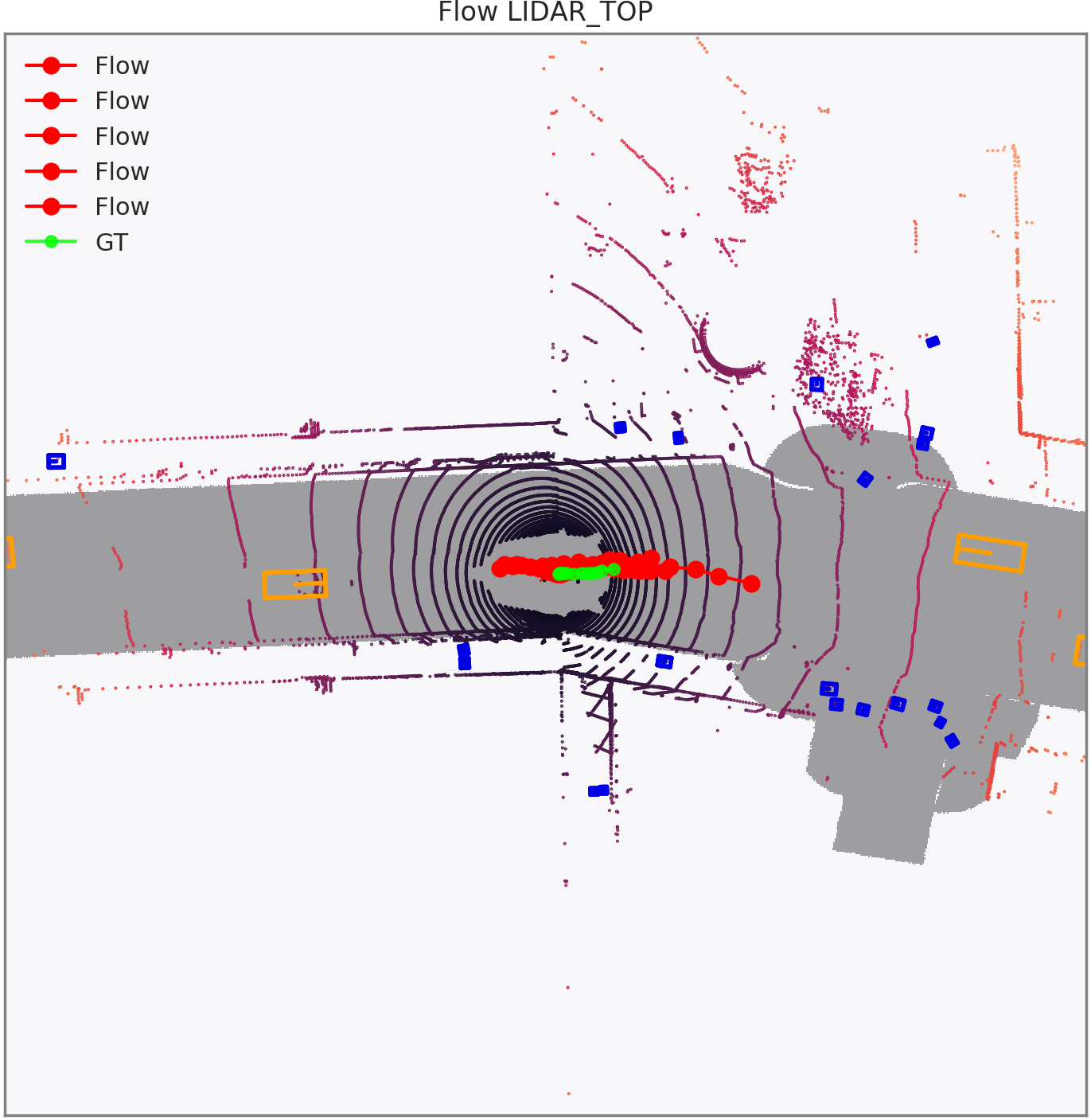} &
\includegraphics[width=.33\textwidth, trim={0 4cm 0 4cm}, clip=True]{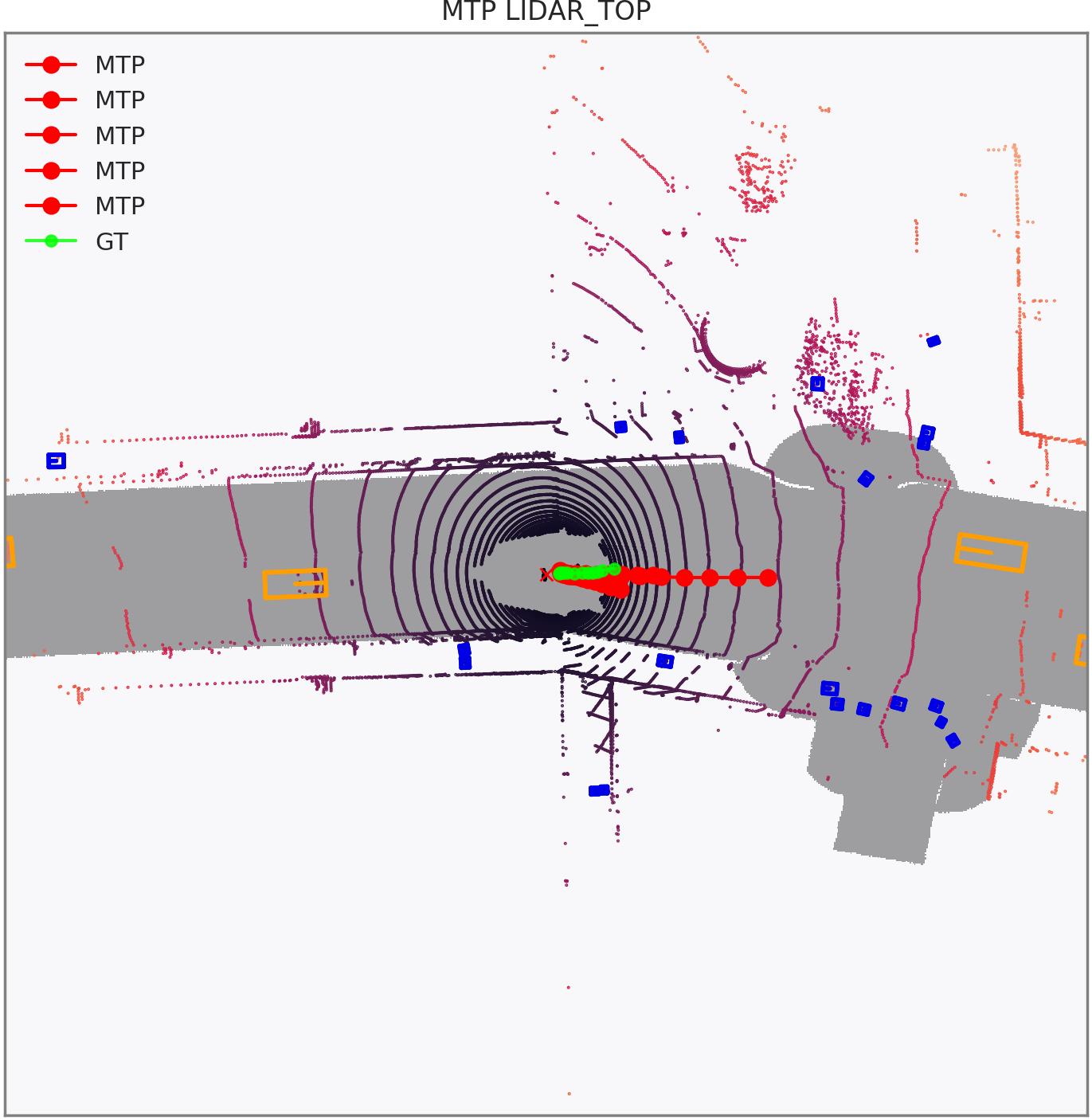} 
\\[-1cm]
\rotatebox[]{90}{$\quad \quad \quad \quad \quad \textbf{Scene 2}$} &
\includegraphics[width=.33\textwidth, trim={0 4cm 0 4cm}, clip=True]{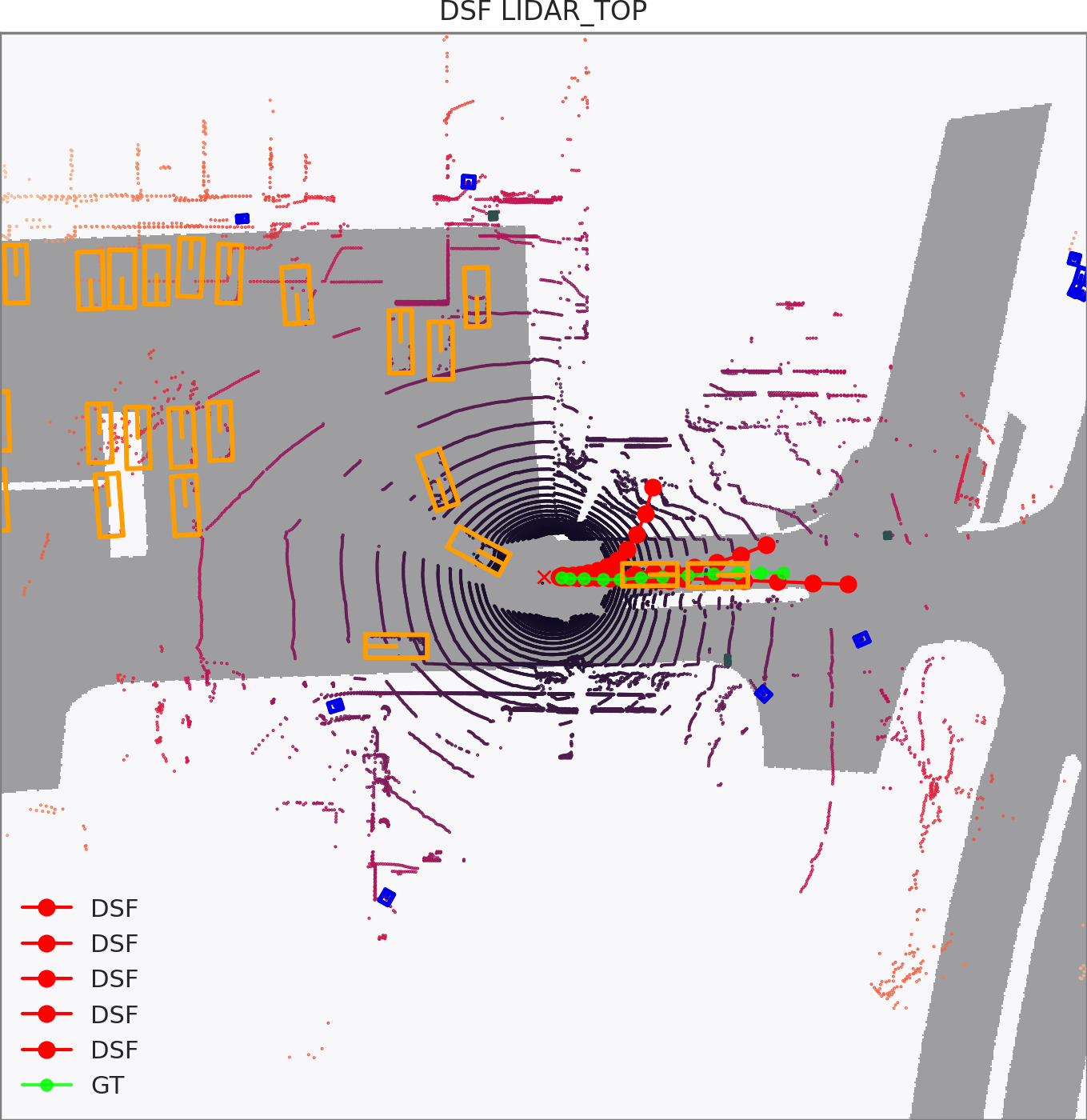} &
\includegraphics[width=.33\textwidth, trim={0 4cm 0 4cm}, clip=True]{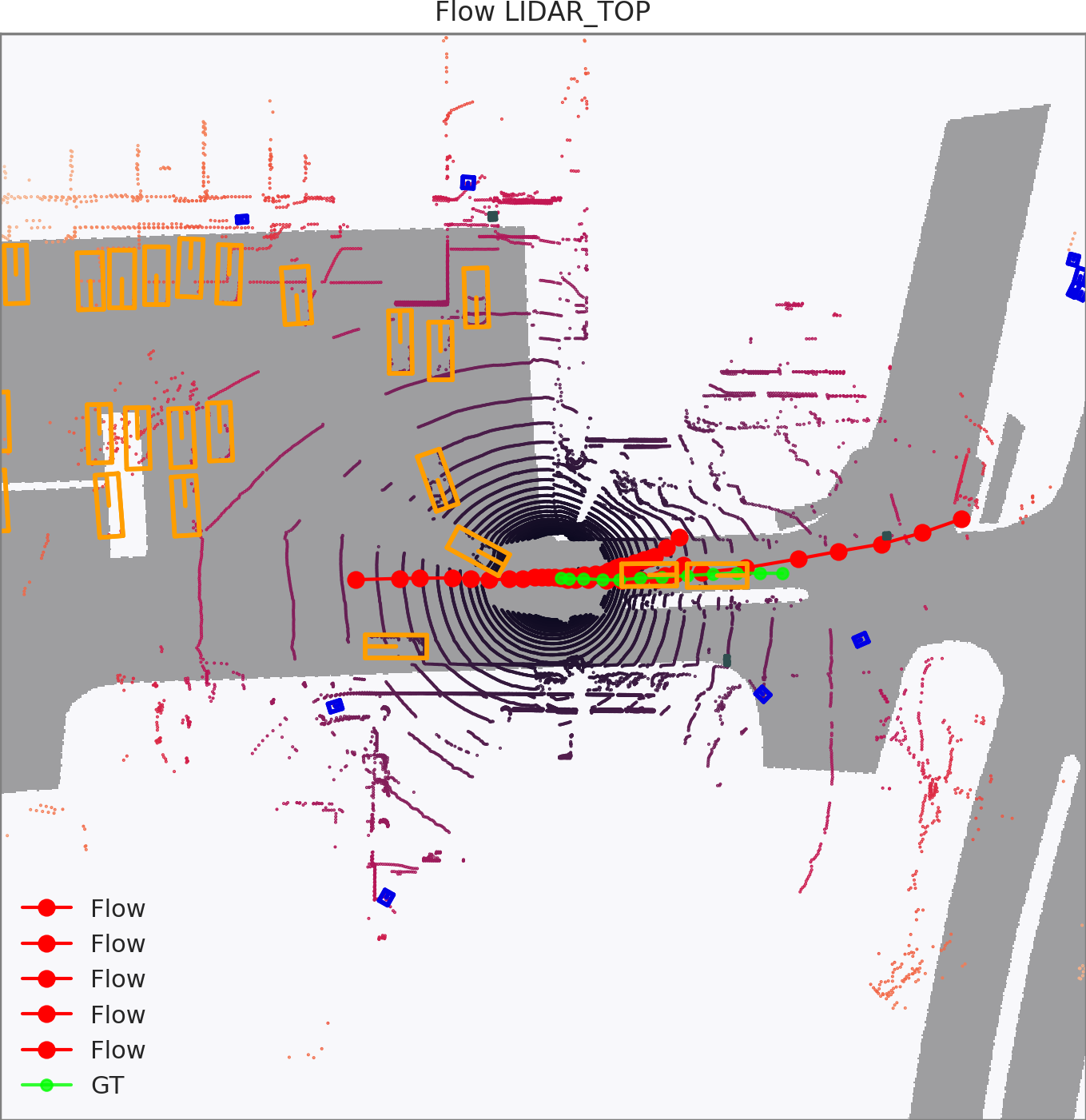} &
\includegraphics[width=.33\textwidth, trim={0 4cm 0 4cm}, clip=True]{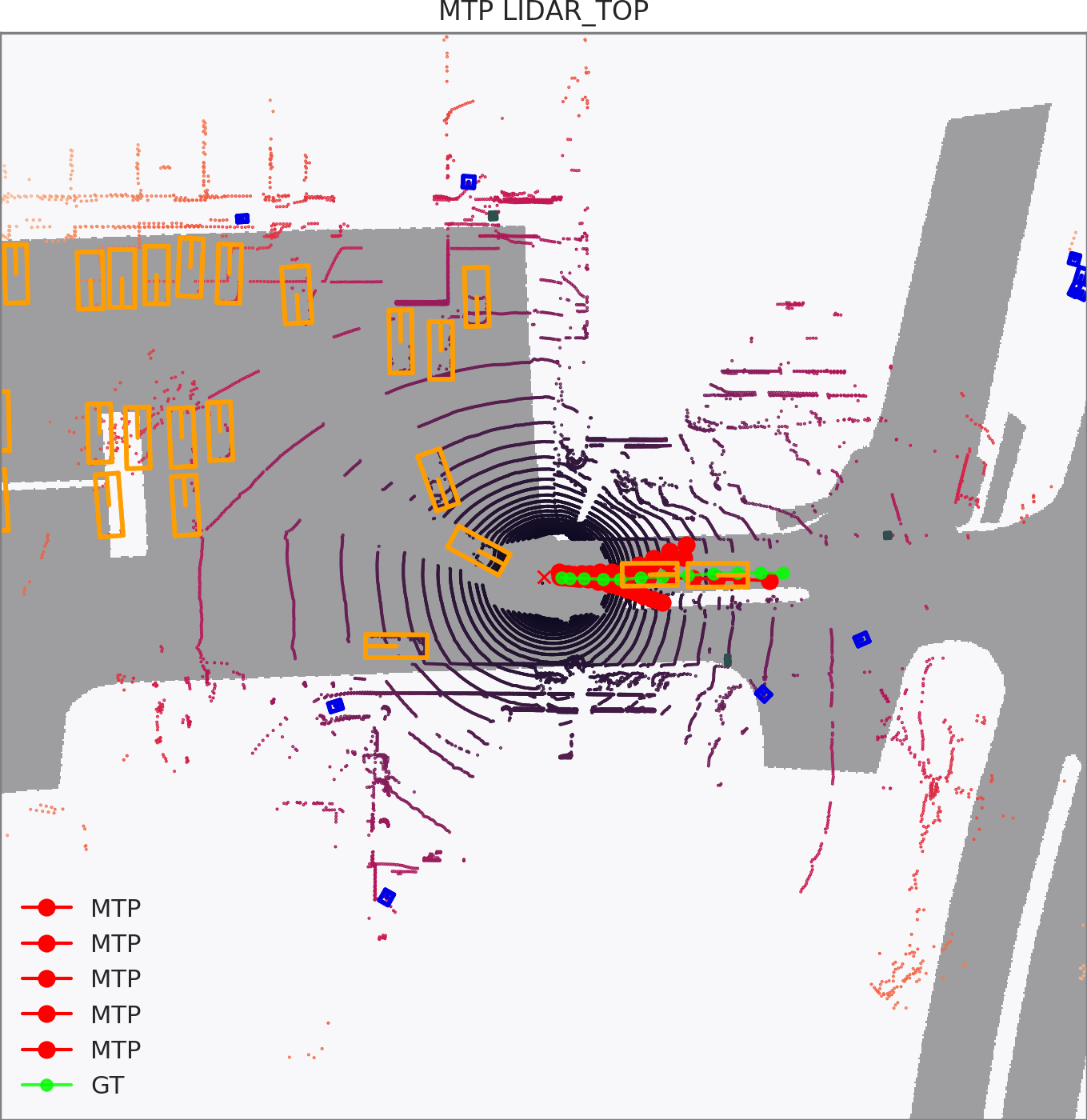} 
\\[-1cm]
\rotatebox[]{90}{$\quad \quad \quad \quad \quad \textbf{Scene 3}$} &
\includegraphics[width=.33\textwidth, trim={0 4cm 0 4cm}, clip=True]{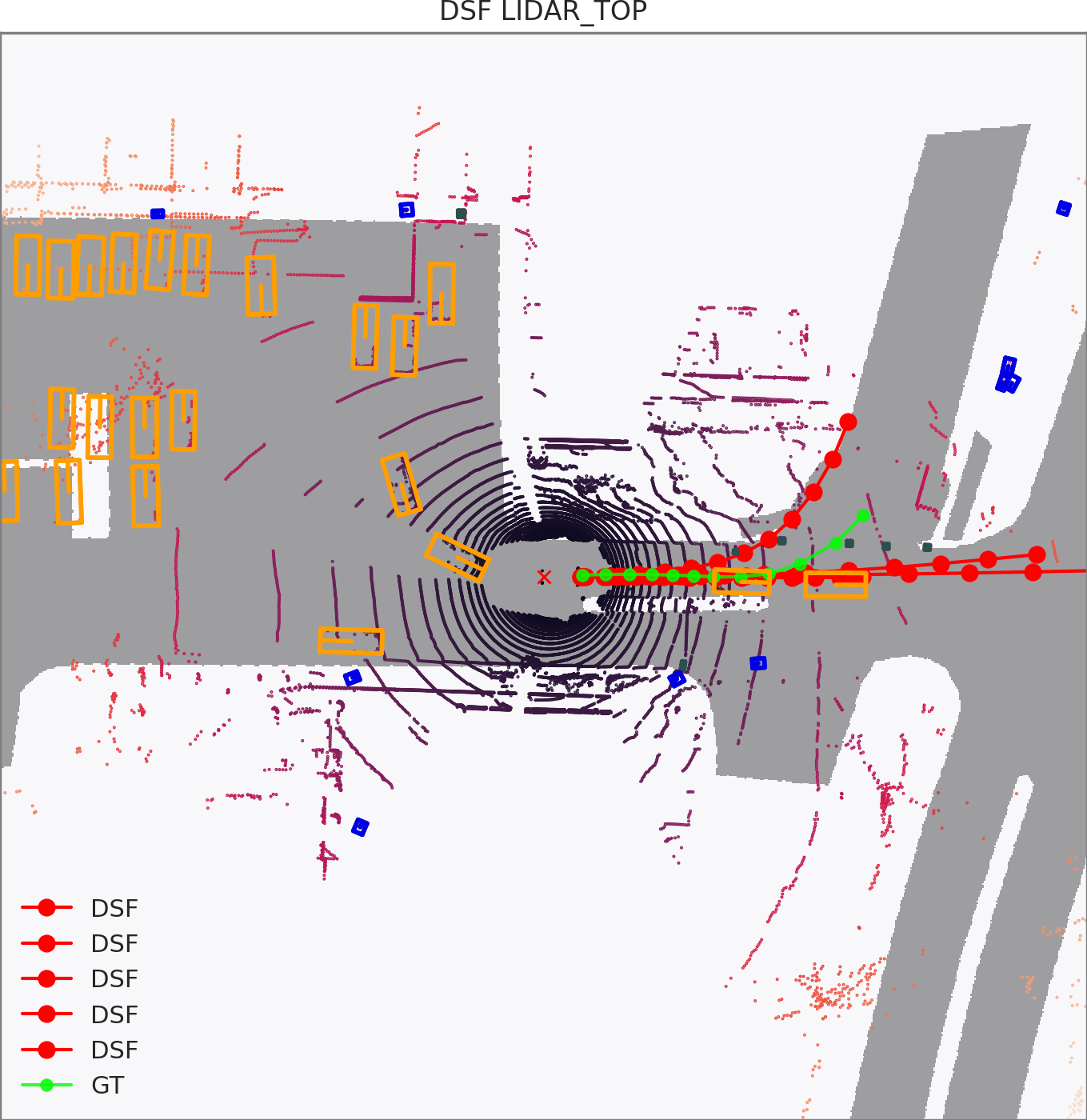}&
\includegraphics[width=.33\textwidth, trim={0 4cm 0 4cm}, clip=True]{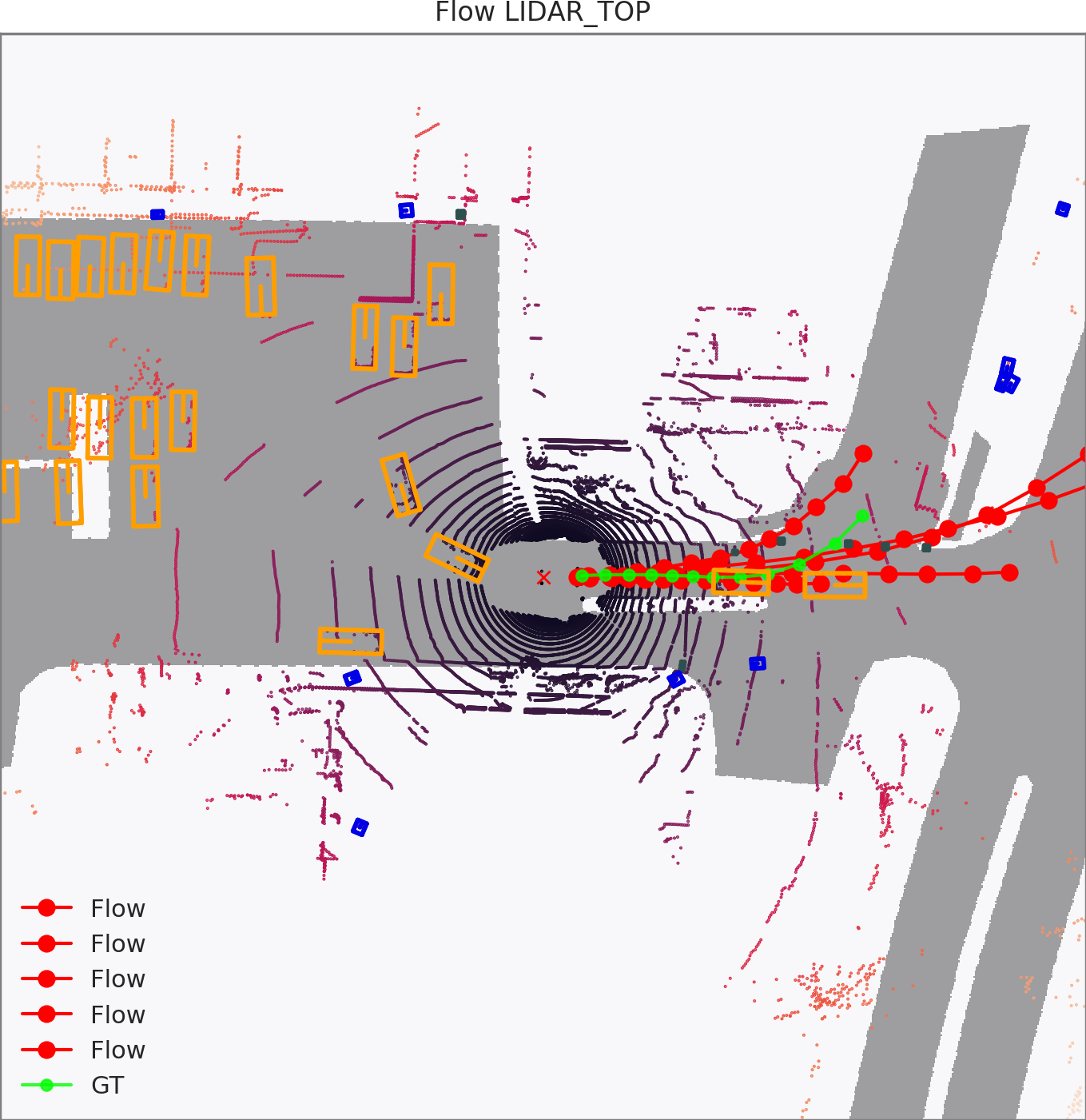}&
\includegraphics[width=.33\textwidth, trim={0 4cm 0 4cm}, clip=True]{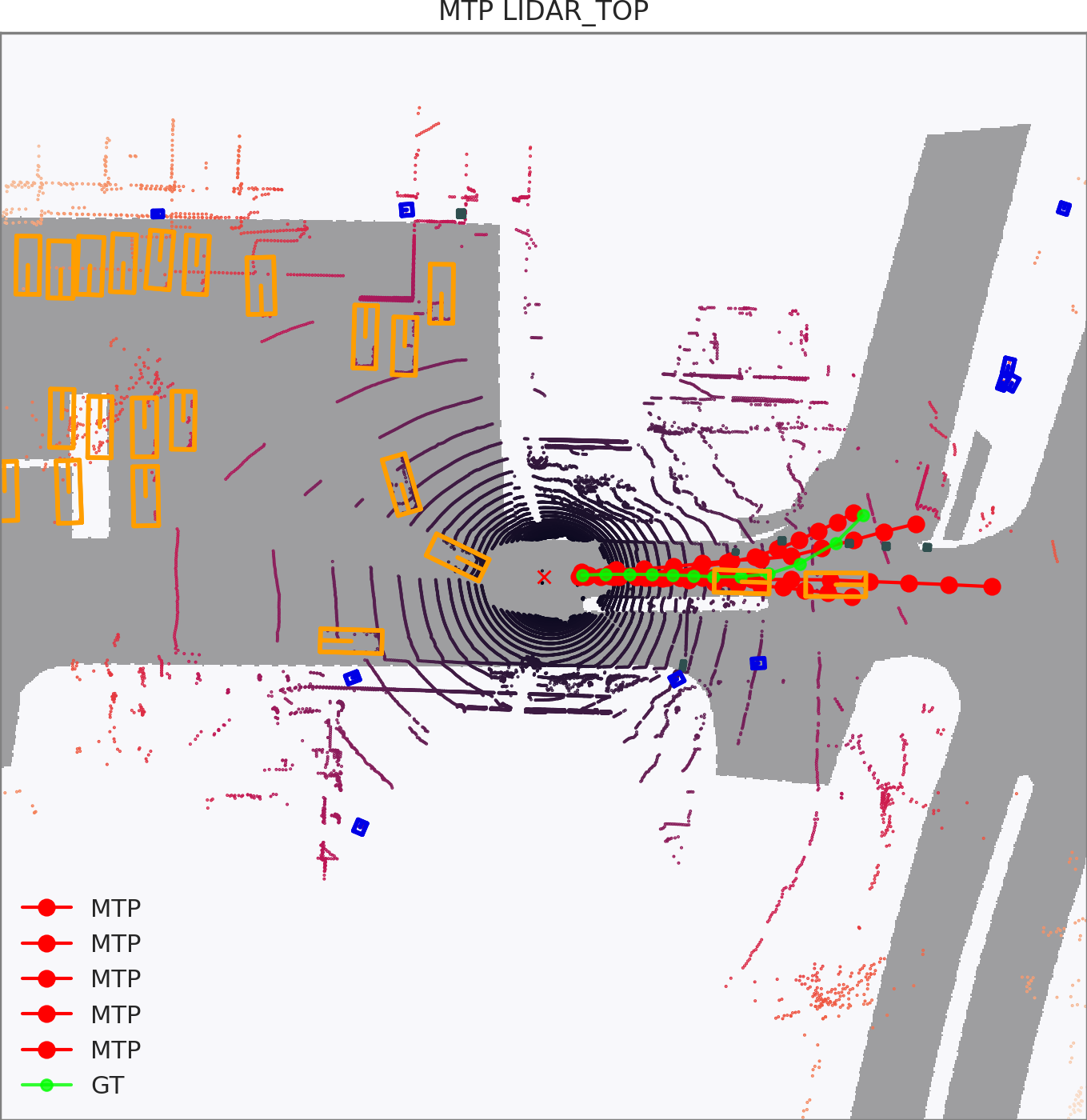}
\\[-1cm]
\rotatebox[]{90}{$\quad \quad \quad \quad \quad \textbf{Scene 4}$} &
\includegraphics[width=.33\textwidth, trim={0 4cm 0 4cm}, clip=True]{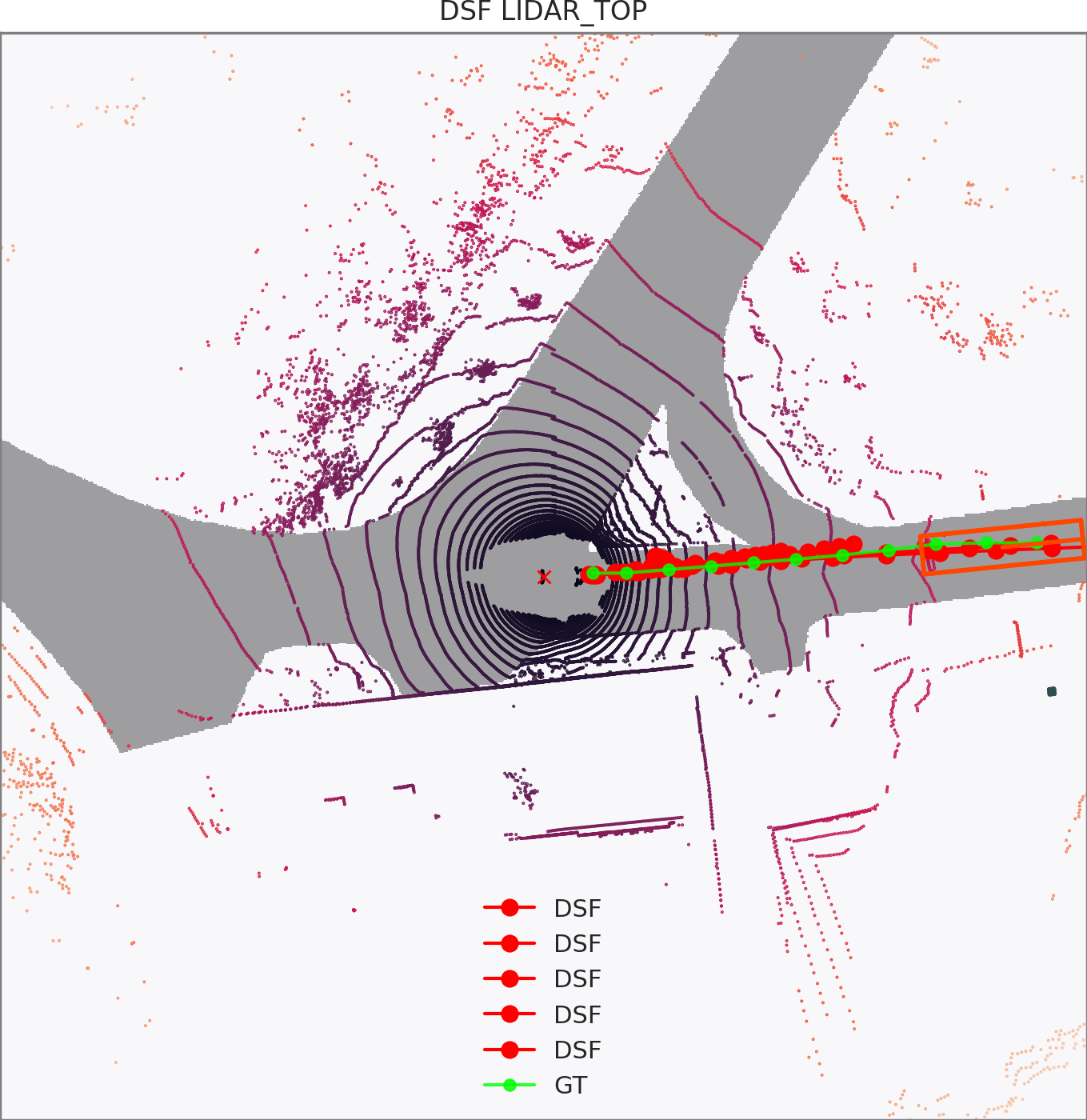}&
\includegraphics[width=.33\textwidth, trim={0 4cm 0 4cm}, clip=True]{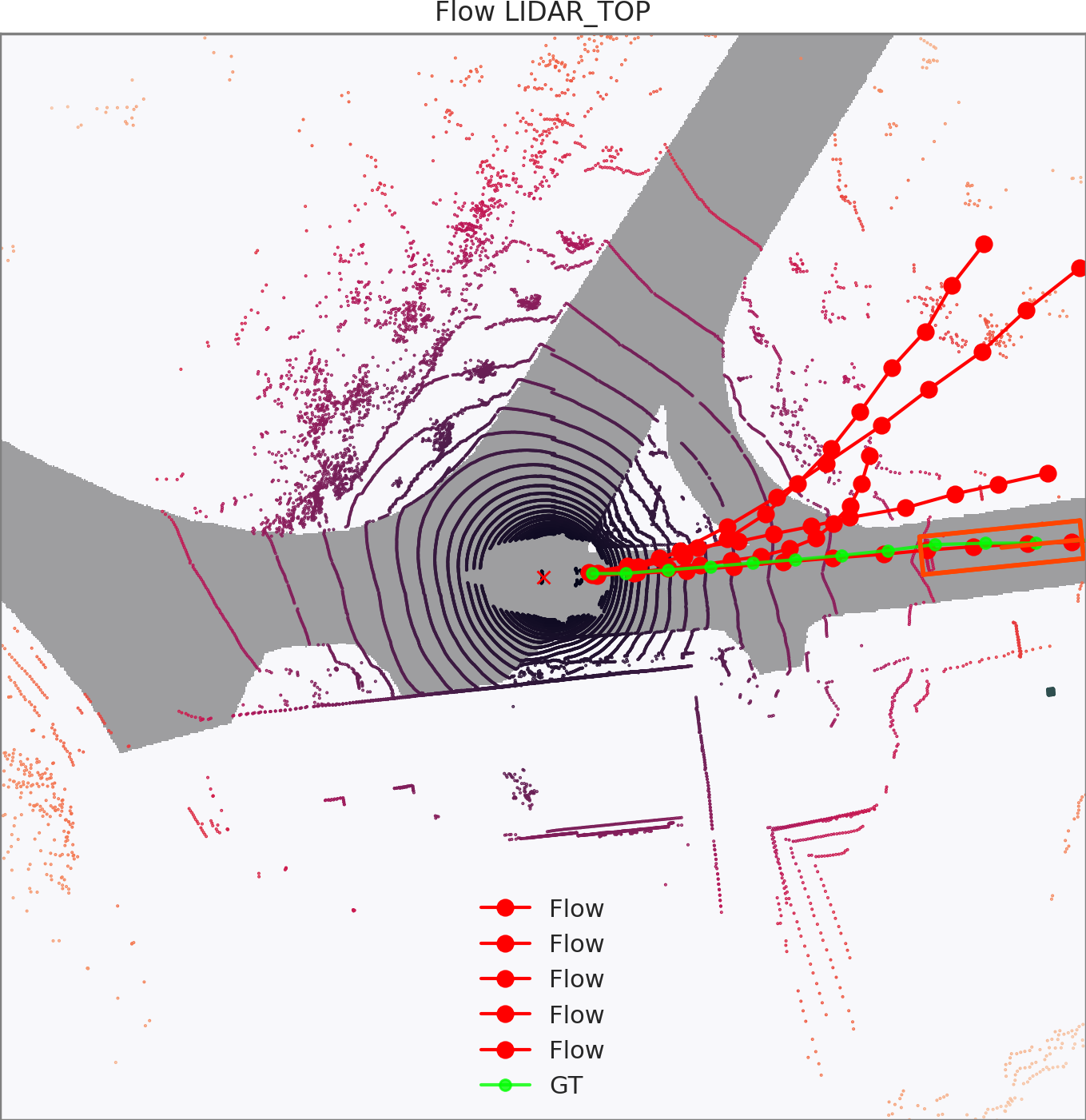}&
\includegraphics[width=.33\textwidth, trim={0 4cm 0 4cm}, clip=True]{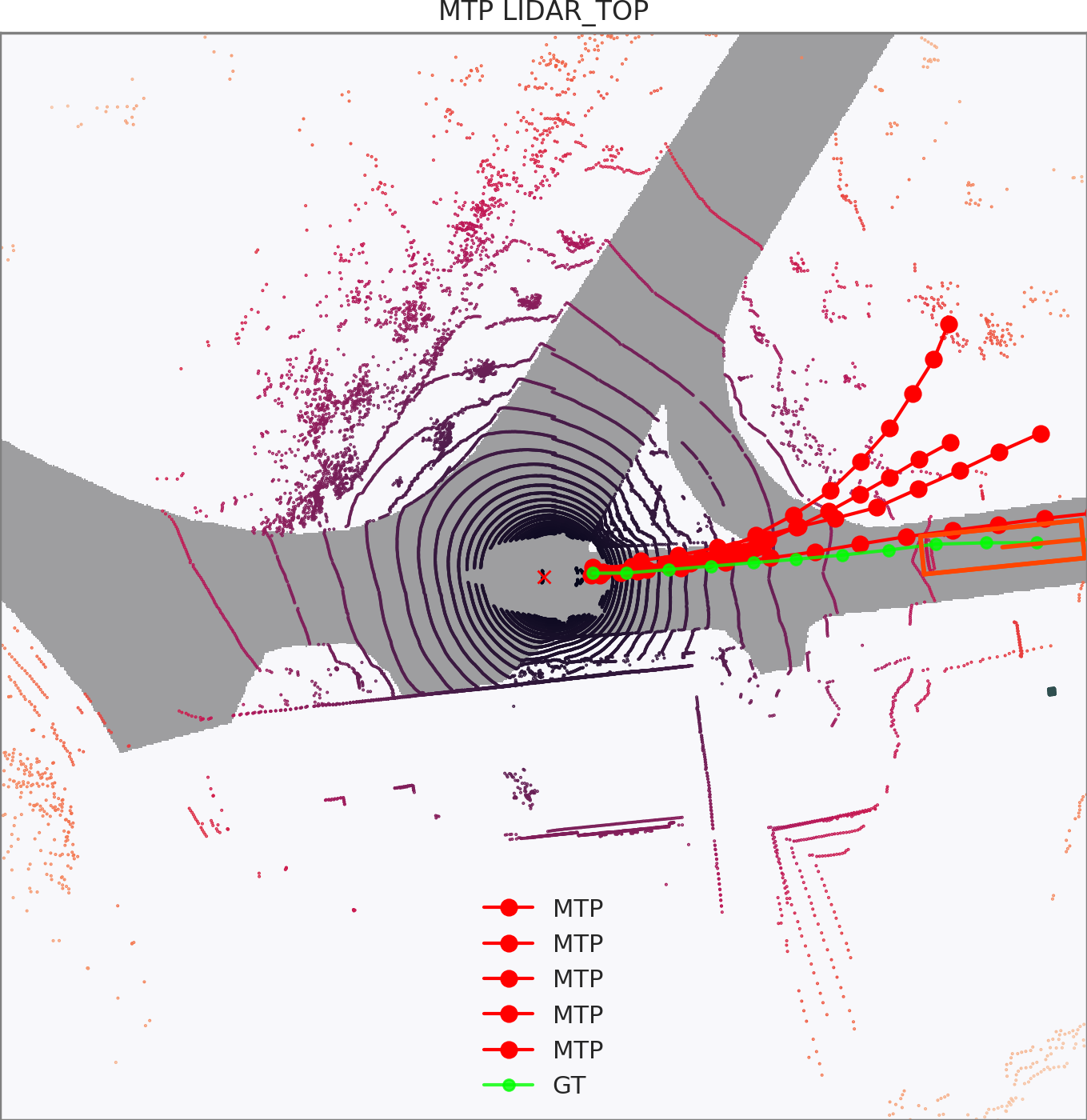}
\end{tabular}}
\caption{Additional model visualizations. The models from left to right: \textbf{LDS}, \textbf{AF}, and \textbf{MTP-Lidar}.}
\label{fig:additional-nuscenes-prediction-visualization}
\end{figure*}

\para{LDS-AF vs. AF vs. MTP-Lidar.} Additional visualizations of LDS-AF, AF, and MTP-Lidar outputs (Figure \ref{fig:additional-nuscenes-prediction-visualization}).

\begin{figure*}
\centering
\includegraphics[width=.33\textwidth]{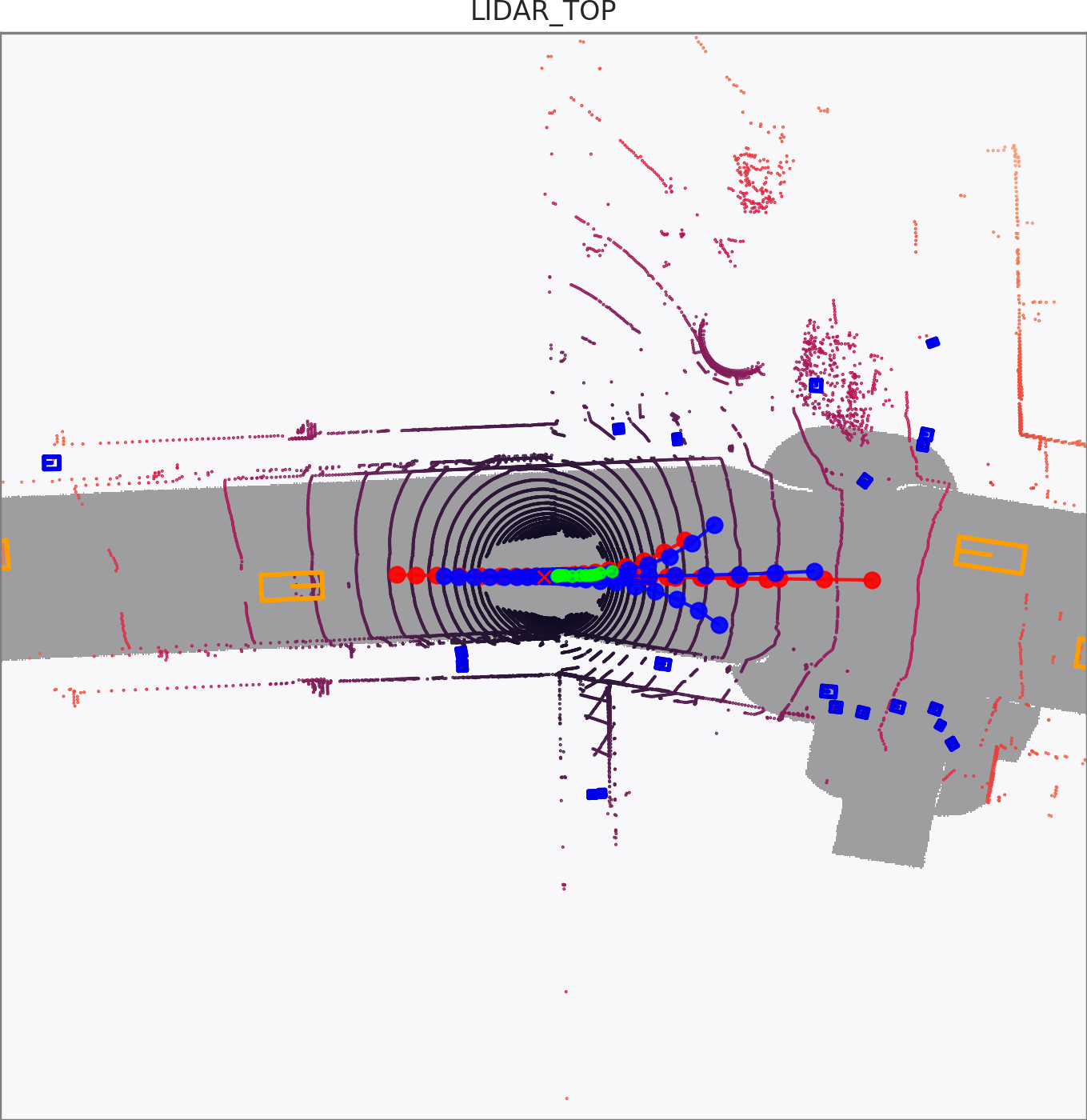}\hfill
\includegraphics[width=.33\textwidth]{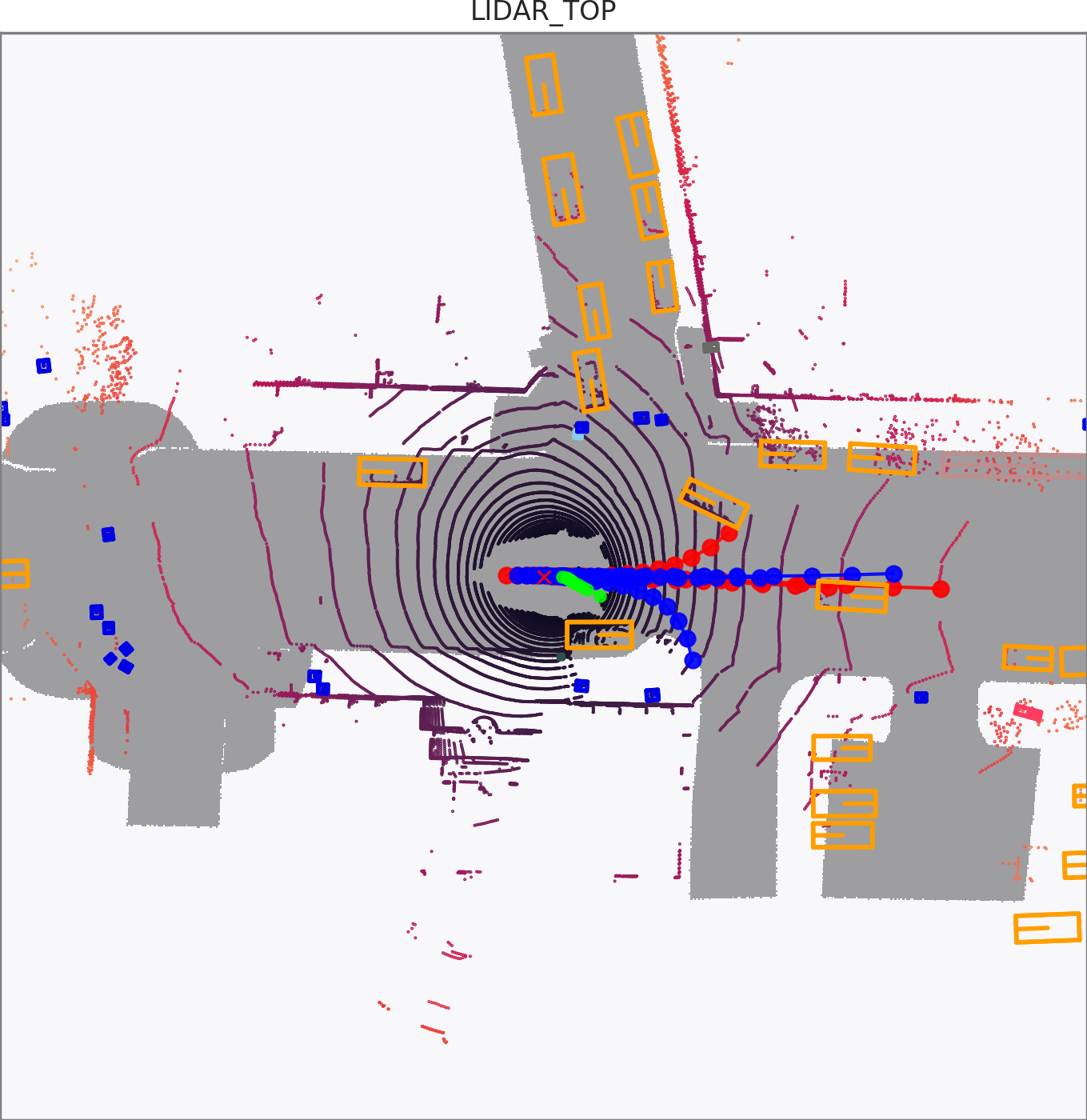}\hfill
\includegraphics[width=.33\textwidth]{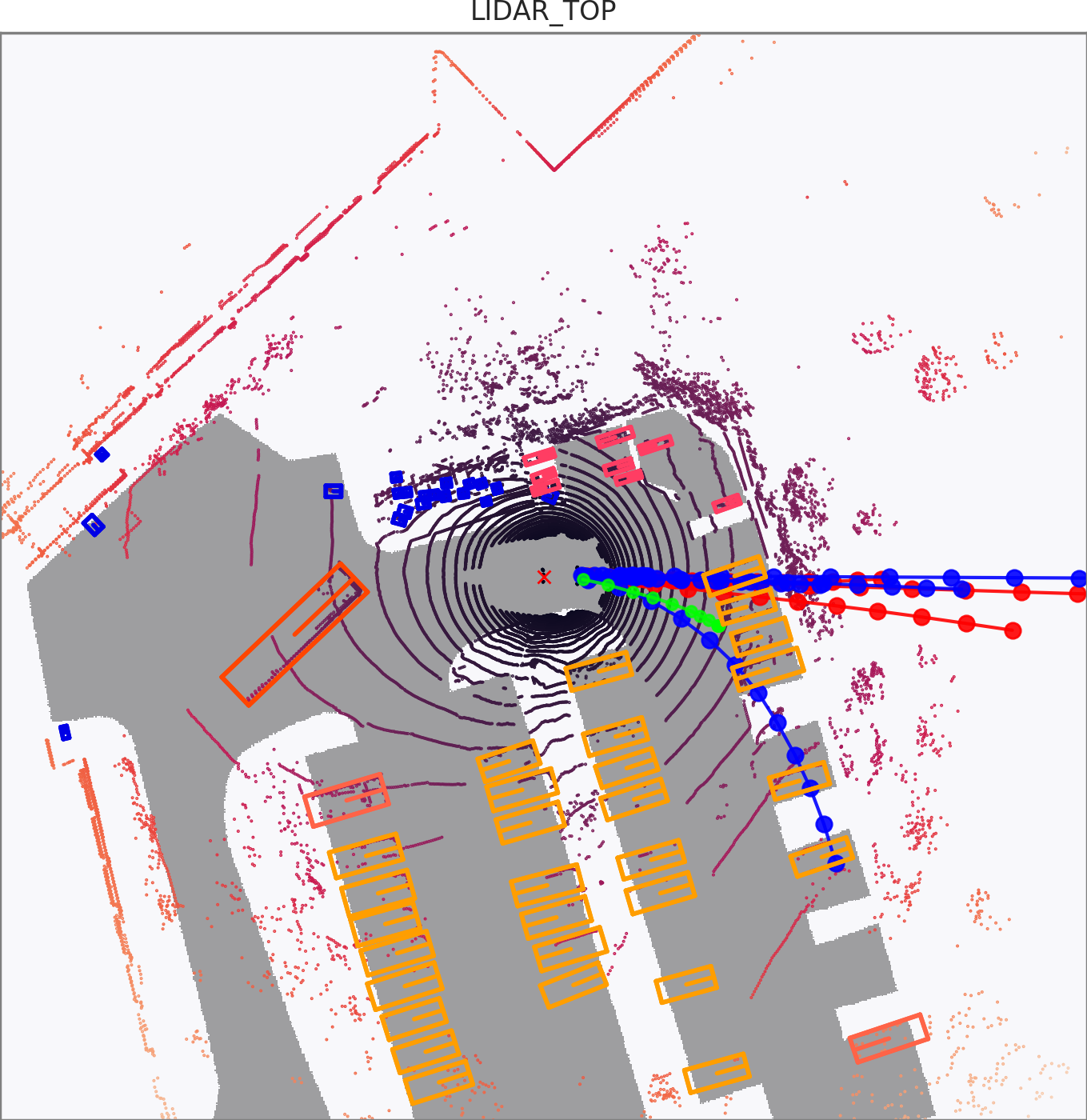}\hfill
\caption{\textbf{LDS-AF} (red) vs. \textbf{LDS-AF-TD-NN} (blue) in NuScenes.}
\label{fig:lds-af-vs-lds-af-td-nn}
\end{figure*}

\para{LDS-AF vs. LDS-AF-TD-NN.} Visualizations of LDS-AF (red) vs. LDS-AF-TD-NN (blue) are provided in Figure \ref{fig:lds-af-vs-lds-af-td-nn}. As shown, LDS-AF-TD-NN generally adapt its predictions to the current observation better and produce more realistic trajectories with respect to the ground truth. 

\begin{figure*}
\centering
\includegraphics[width=.33\textwidth, trim={0 0cm 0 1cm}, clip=True]{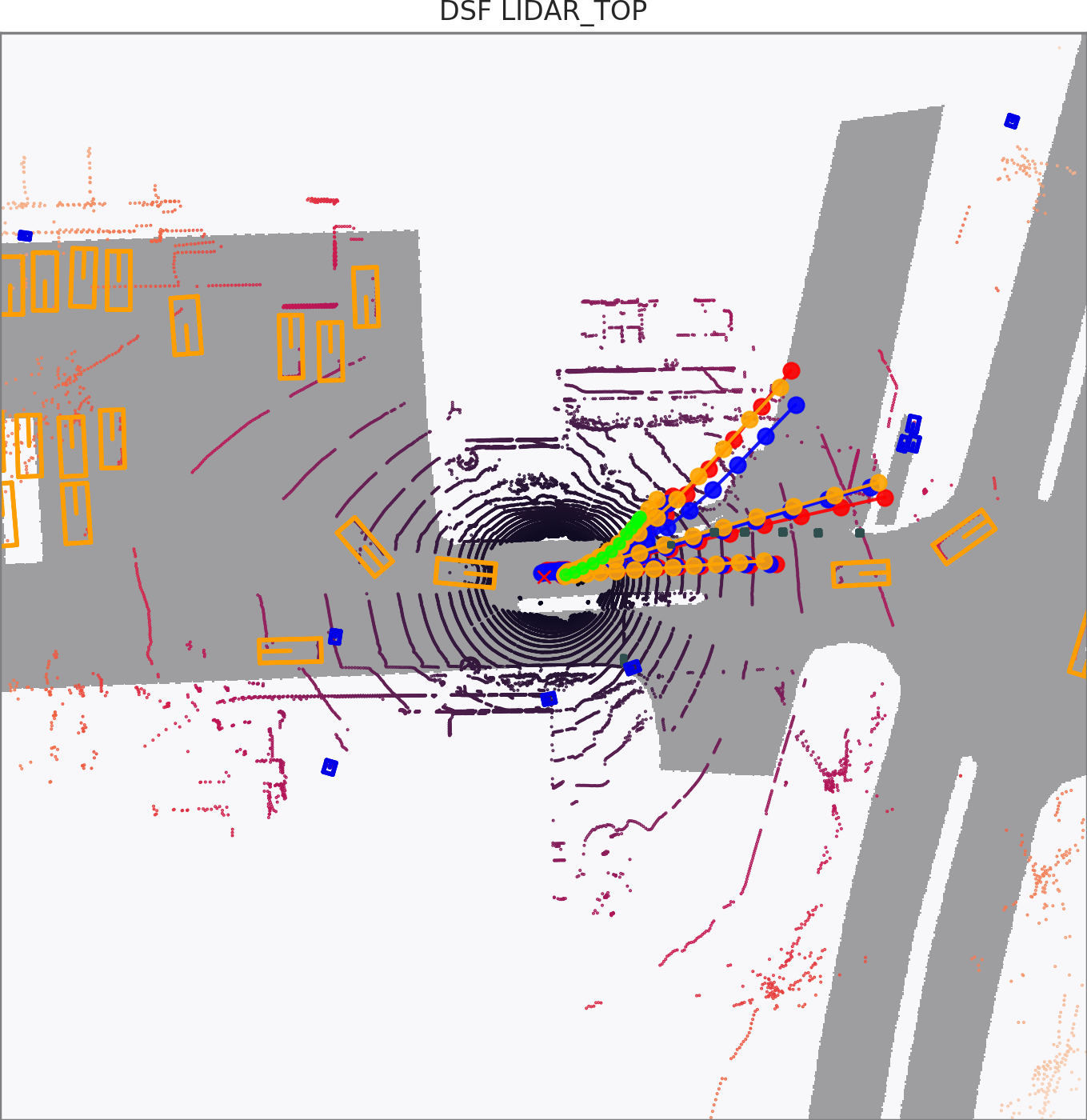}\hfill
\includegraphics[width=.33\textwidth, trim={0 0cm 0 1cm}, clip=True]{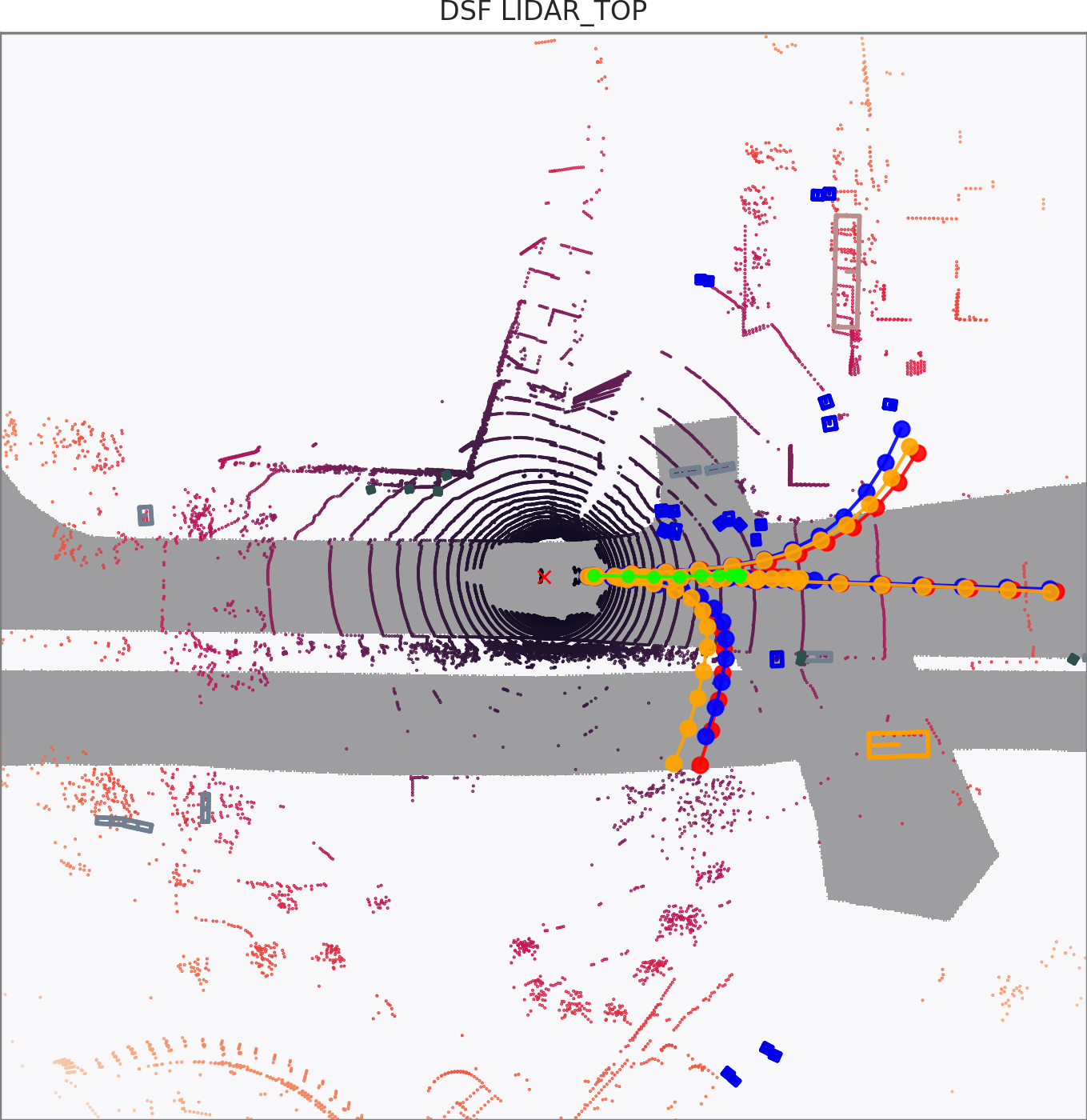}\hfill
\includegraphics[width=.33\textwidth, trim={0 0cm 0 1cm}, clip=True]{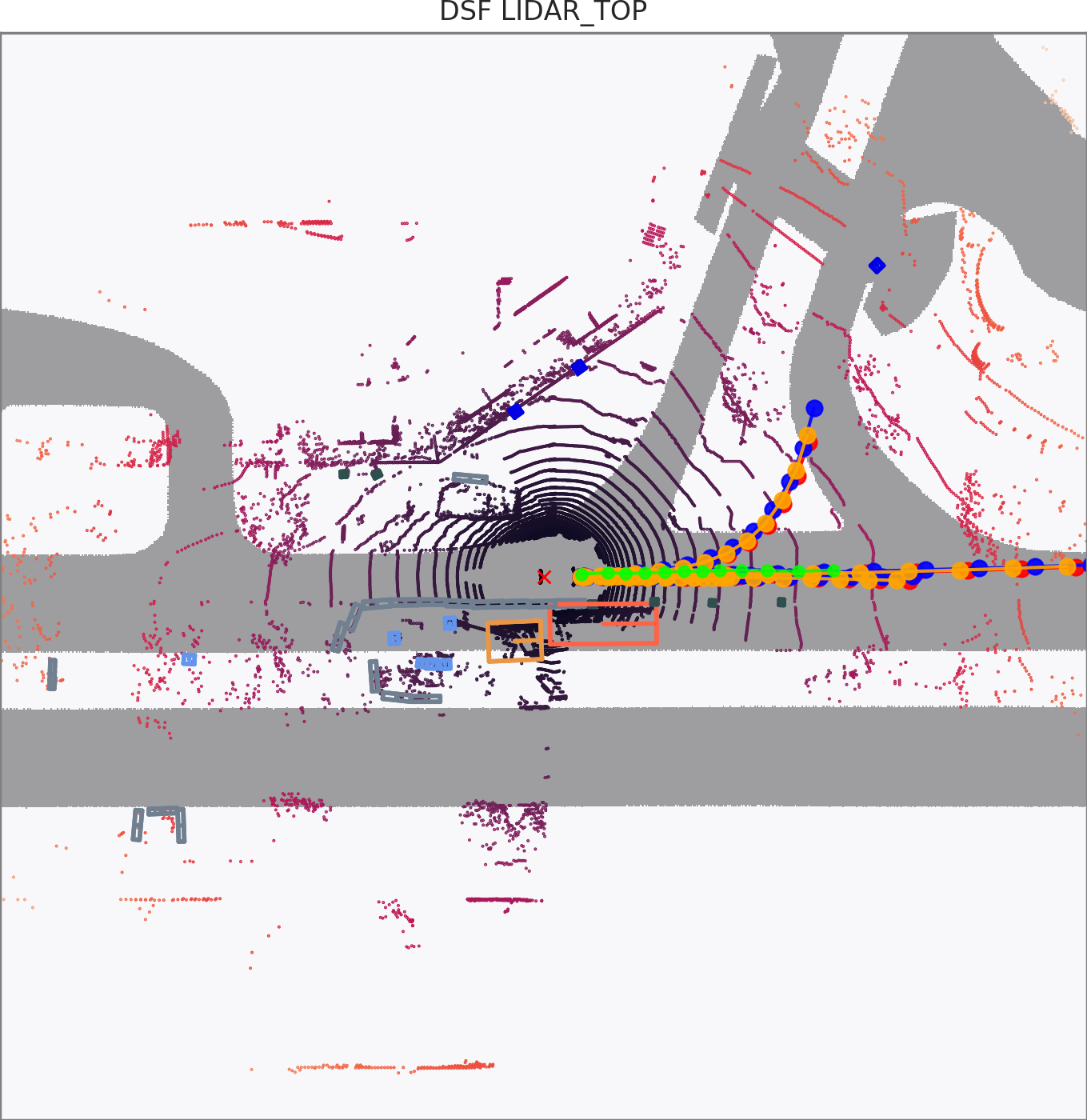}\hfill
\caption{Effect of varying $\epsilon$ on \textbf{LDS-AF} in NuScenes. The three colors (red, blue, orange) denote three different sets of trajectories by randomly sampling $\epsilon$.}
\label{fig:epsilon-figures}
\end{figure*}

\para{Varying $\epsilon$.} Finally, we visualize different sets of LDS-AF trajectories by randomly sampling different $\epsilon \sim \mathcal{N}(0, \mathbf{I})$. The visualizations are in \ref{fig:epsilon-figures}.

\section{Forking Paths Experimental Details}
\label{appendix:forking-paths-details}
\subsection{Dataset Details}
Here, we briefly describe the Forking Paths evaluation dataset \cite{liang2020garden}. FP semi-automatically reconstruct static scenes and their dynamic elements (e.g. pedestrians) from real-world videos in ActEV/VIRAT and ETH/UCY in the CARLA simulator. To do so, it converts the ground truth trajectory annotations from the real-world videos to coordinates in the simulator using the provided homography matrices of the datasets. Then, 127 ``controlled agents" (CA) are selected from the 7 reconstructed scenes. For each CA, there are on average $5.9$ human annotators to control the pedestrian to the pre-defined destinations in a ``natural" way that mimics real pedestrian behaviors. The scenes are also rendered to the annotators in different camera angles ranging from `Top-Down' to `45-Degree'. The annotation can last up to 10.4 seconds, which is far longer than the $4.8$ seconds prediction window in the original datasets, making the forecasting problem considerably harder. In total, there are $750$ trajectories after data cleaning. 

\subsection{CAM-NF Model Details}
Compared to the perception-based flow model we use in nuScenes, CAM-NF only uses the historical trajectory of the agent and other surrounding agents in the scene, consistent with various prior methods \cite{gupta2018social, alahi2016social} in pedestrian forecasting; we leave exploring the utility of added perception modules as in \cite{liang2019peeking, liang2020garden} to future work. 
Hence, $\mathbf{o} = \{\mathbf{S}^a_{-T_{\text{hist}}:0}\}_{a=1}^A$, where $A$ is the number of pedestrians in the scene and $T_{\text{hist}}$ the length of history. CAM-NF first encodes the history of all pedestrians in the scene using a LSTM encoder. Then, it computes cross-pedestrian attention feature vectors using self-attention \cite{vaswani2017attention} to model the influences of nearby pedestrians, and uses these attention vectors as features for a normalizing flow decoder, which outputs future trajectories of the predicted pedestrian. The decoder is an autoregressive affine flow similar to the one used for nuScenes. Here, we briefly describe the architectures and refer interested readers to the original work.

The encoder first uses an LSTM of hidden dimension $512$ \cite{hochreiter1997long} to extract a history embedding for every agent up to the most recent timestep $t=0$:
$$
h_t^a = \text{LSTM}(\textbf{s}_{t-1}^a, h_{t-1}^a) \quad t=T_{\text{hist}}-1,...,0
$$
Then, the history embedding of all agents $h_t^0,..., h_t^A$ are aggregated to compute a corresponding cross-agent attention embedding using self-attention. This embedding is then combined with the history embedding to form the inputs to the normalizing flow decoder:
$$
\tilde{h}^a = h_0^a + \text{SELF-ATTENTION}(Q^a, \textbf{K}, \textbf{V})
$$
where
$(Q^a, K^a, V^a)$ is the query-key-value triple for each agent, and the bold versions are their all-agent aggregated counterparts. Note that $\tilde{h}^a$ is passed through a linear layer of hidden dimension $256$ before passed to the decoder.

The decoder is similar to the autoregressive affine flow used for nuScenes with a few minor changes. First, $\sigma_{\theta} \in \mathbb{R}^{2 \times 2}$ to model correlation between the two dimensions (i.e. $x, y$) of the states. Second, a velocity smoothing term $\alpha(\textbf{s}_{t-1}-\textbf{s}_{t-2})$ is added to the step-wise update. That is,
\begin{equation}
    \label{eq:flow-autoregressive}
    \begin{aligned}
    \textbf{s}_t -\textbf{s}_{t-1} &= {\tau}_{\theta}(\textbf{z}_t; \textbf{z}_{<t}) \\ &=\alpha(\textbf{s}_{t-1}-\textbf{s}_{t-2}) + \underbrace{\mu_{\theta}(\textbf{s}_{1:t-1}, \phi)}_{\text{MLP}_1(\textbf{h}_t)} + \underbrace{\sigma_{\theta}(\textbf{s}_{1:t-1},\phi)}_{\text{EXP}\Big(\text{MLP}_2(\textbf{h}_t)\Big)} \textbf{z}_t
    \end{aligned}
\end{equation}
We set $\alpha=0.5$ as in the original work. Our implementation is adapted from the original implementation\footnote{\url{https://github.com/kami93/CMU-DATF}}.

\subsection{LDS Architecture Details}
\label{appendix:forking-paths-LDS-details}
Since trajectories in FP have different lengths, we set $T=25$ in the LDS architecture to ensure that a bijective mapping between $\textbf{Z} \in \mathbb{R}^{T \times D}$ exists for all samples in FP. This also means that during training, we optimize LDS for up to $25$ steps. We also slightly increase the size of LDS neural network and set the maximum diversity loss to be $80$, which we find to work well empirically. As before, DLow and DSF use the same architecture as LDS. The whole architecture is as follows:
\begin{table}[H]
  \caption{LDS Architecture and Hyperparameters Overview}
  \label{table:LDS-architecture}
  \centering
  \begin{tabular}{ll}
    \toprule
    \cmidrule(r){1-2}
    Attributes & Values \\
    \midrule
    LDS Architecture & Linear(Input Size, 128)  \\
    & Linear(128,64) \\
    & Linear(64, 2 $\times$ 25 $\times$ K) \\
    \midrule
    Learning Rate & 0.001 \\ 
    $\lambda_d$ & 10 \\ 
    Diversity function clip value & 80\\ 
    \bottomrule
  \end{tabular}
\end{table}
\subsection{Training Details}
We train CAM-NF for $200$ epochs with learning rate $10^{-3}$ using Adam optimizer and batch size $64$. We train LDS, LDS-TD, and DLow models using mode hyperparameter $K=20$.  LDS and DLow iterate through the full training set once, while LDS-AF-TD directly optimizes on the test set with a minibatch size of $64$ and $200$ adaptation iterations for every minibatch. For all LDS models, through a hyperparameter search , we set $\lambda_d = 10$. For all models, we train $5$ separate models using random seeds and report the average and standard deviations.

\subsection{ActEV/VIRAT Results}
\label{appendix:actev-virat-results}
\begin{table}[H]
\resizebox{\columnwidth}{!}{
\begin{tabular}{lll}
\toprule
Method & $\text{minADE}_1 (\downarrow)$ & $\text{minFDE}_1 (\downarrow)$ \\
\midrule
\textbf{Linear}$^*$ & 32.19 & 60.92 \\
\textbf{LSTM}$^*$ & 23.98 & 44.97\\
\textbf{Social-LSTM}$^*$ & 23.10 & 44.27 \\
\textbf{Social-GAN}$^*$ &  23.10 & 44.27\\ 
\textbf{Next}$^*$ &  19.78 & 42.43\\ 
\textbf{Multiverse}$^*$ & \textbf{18.51} & \textbf{35.84}\\
\midrule 
\textbf{CAM-NF} & 19.69 $\pm$ 0.15 & 39.12 $\pm$ 0.31\\
\bottomrule
\end{tabular}}
\caption{Training Results on ActEV/VIRAT. Our backbone flow model CAM-NF achieves comparable performance to the current state-of-art Multiverse.}
\end{table}

\subsection{Forking Paths Full Sub-Category Split Results}
\label{appendix:forking-path-split}
\begin{table}[H]
\resizebox{\columnwidth}{!}{
\begin{tabular}{c|cc|cc}
\hline 
\multirow{2}{*}{Method}& \multicolumn{2}{c|}{$\text{minADE}_{20} (\downarrow)$} & \multicolumn{2}{c}{$\text{minFDE}_{20} (\downarrow)$} \\
\cline{2-5}
& 45-Degree & Top Down & 45-Degree & Top Down \\
\hline
\textbf{Linear}$^*$ &213.2 & 197.6 & 403.2 & 372.9 \\
\textbf{LSTM}$^*$ & 201.0 $\pm$ 2.2 & 183.7 $\pm$ 2.1 & 381.5 $\pm$ 3.2 & 355.0 $\pm$ 3.6 \\
\textbf{Social-LSTM}$^*$ \cite{alahi2016social}& 197.5 $\pm$ 2.5 & 180.4 $\pm$ 1.0 & 377.0 $\pm$ 3.6 & 350.3 $\pm$ 2.3 \\
\textbf{Social-GAN}$^*$ \cite{gupta2018social}& 187.1 $\pm$ 4.7 & 172.7 $\pm$ 3.9 & 342.1 $\pm$ 10.2 & 326.7 $\pm$ 7.7 \\ 
\textbf{Next}$^*$ \cite{liang2019peeking} & 186.6 $\pm$ 2.7 & 166.9 $\pm$ 2.2 & 360.0 $\pm$ 7.2 & 326.6 $\pm$ 5.0 \\ 
\textbf{Multiverse}$^*$ \cite{liang2020garden} &
168.9 $\pm$ 2.1 & 157.7 $\pm$ 2.5 & 333.8$\pm$ 3.7 & 316.5 $\pm$ 3.4 \\
\hline
\textbf{CAM-NF} \cite{park2020diverse} & 155.2 $\pm$ 2.4 & 140.8 $\pm$ 2.2 & 305.0 $\pm$ 4.6 & 282.2 $\pm$ 4.9 \\
\textbf{DSF} \cite{yuan2019diverse} & 169.7 $\pm$ 1.8 & 155.77 $\pm$ 2.1 & 331.7 $\pm$ 3.7 & 309.5 $\pm$ 3.5 \\ 
\textbf{DLow} \cite{yuan2020dlow} & 144.5 $\pm$ 3.8 & 131.0 $\pm$ 8.1 & 284.6 $\pm$ 8.4 & 262.1 $\pm$ 20.5 \\ 
\textbf{LDS} (Ours) & \textbf{103.8} $\pm$ 6.9 & \textbf{93.4} $\pm$ 4.8 & \textbf{190.6} $\pm$ 16.3 & \textbf{173.4} $\pm$ 12.8 \\
\textbf{LDS-TD-NN} (Ours) & \textbf{105.1} $\pm$ 4.3 & \textbf{94.9} $\pm$ 2.1 & \textbf{188.7} $\pm$ 10.4 & \textbf{167.4} $\pm$ 4.8 \\ 
\hline 
\end{tabular}}
\caption{Evaluation results on Forking Paths. LDS-augmented CAM-NF significantly outperforms all other methods, including Multiverse and DLow-augmented CAM-NF.}
\label{table:forking-path-accuracy-split}
\end{table}

\subsection{Additional Analysis on Forking Paths}
\label{appendix:additional-analysis-forking-paths}
As shown in Table \ref{table:LDS-ablation-loss-function}, compared to nuScenes experiments, LDS exhibits larger improvement over DLow on the FP dataset. We believe that this is because the ground truth futures in the FP test set tend to be longer than in the ActEV/VIRAT training set. Since DLow optimizes for the $L_2$ reconstruction loss between its forecasts and the ground-truth future trajectories in the \textit{training} set, it is limited to improving diversity over the horizon of these training trajectories. Thus, it is unable to produce diverse predictions for longer horizon trajectories, such as those in the test set. In contrast, since LDS directly optimizes the likelihood of future trajectory according to the flow model, it does not rely on ground truth futures. Thus, it can improve diversity over much longer time horizons than in the training data. This contrast further highlights the flexibility of LDS. Finally, we note that we were not able to achieve positive results for DSF; this is likely due to the much larger latent sample dimension $(2\times 20=40)$ for this dataset, which as stated in Section \ref{sec:related-work}, would be an issue for DSF.

\subsection{Additional Visualizations}
\begin{figure*}
\centering
\resizebox{\textwidth}{!}{
\begin{tabular}{cccc}
&\textbf{LDS} (Ours) & \textbf{CAM-NF} & \textbf{Multiverse} \\ 
\rotatebox[]{90}{$\quad \quad \quad \quad \quad \textbf{Scene 1}$} &
\includegraphics[width=.33\textwidth]{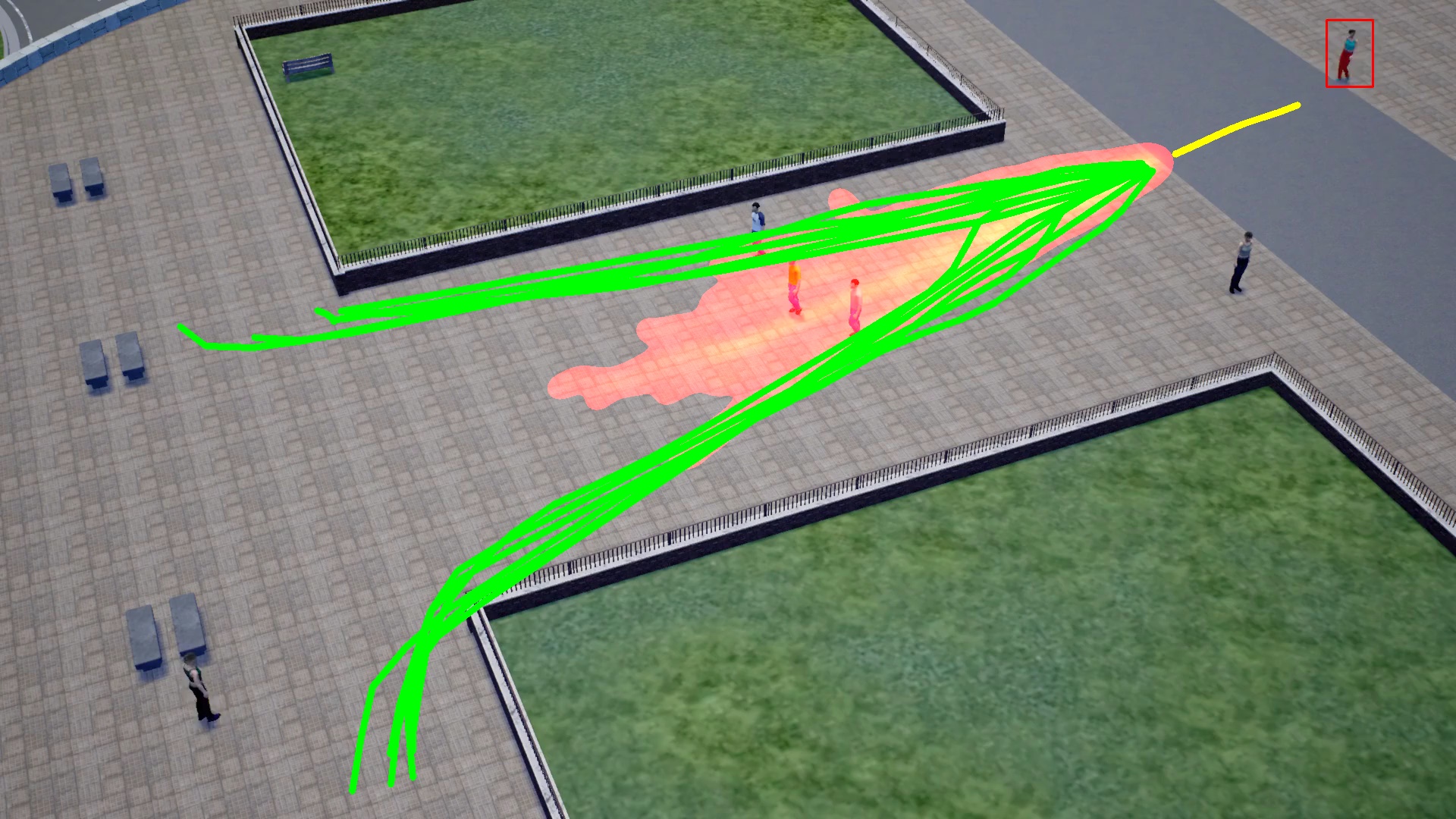}&
\includegraphics[width=.33\textwidth]{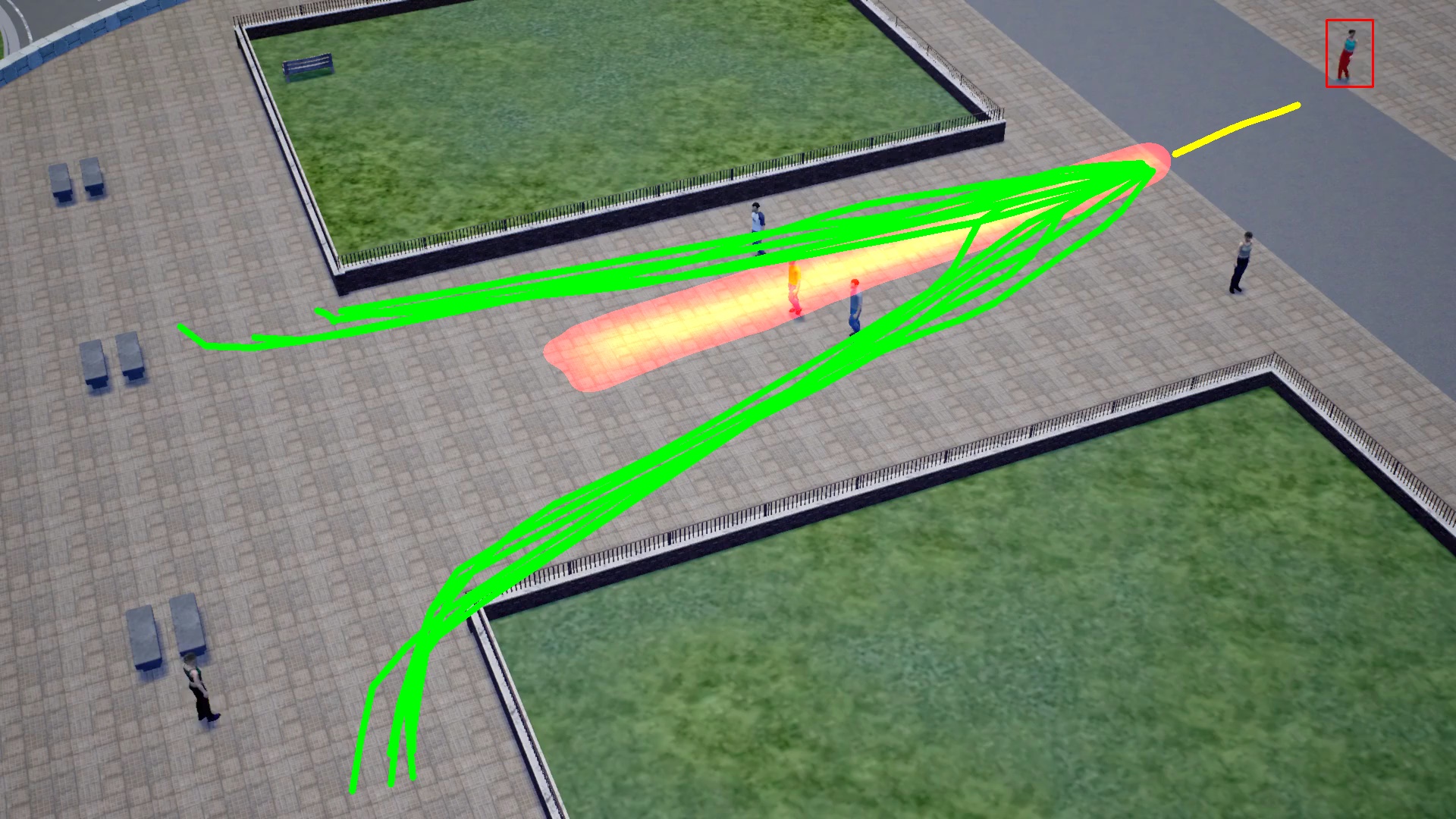}&
\includegraphics[width=.33\textwidth]{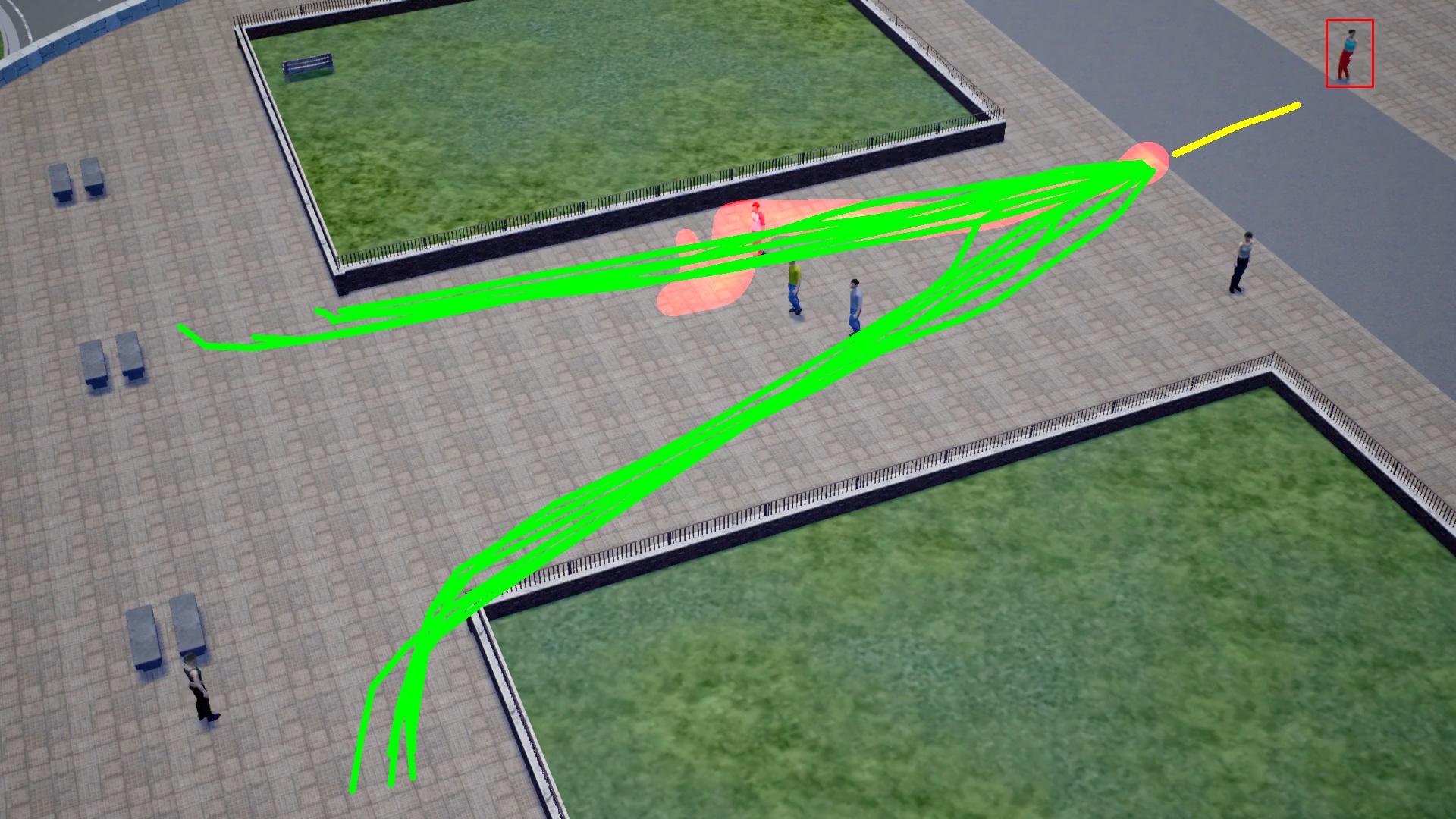}
\\[-1cm]
\rotatebox[]{90}{$\quad \quad \quad \quad \quad \textbf{Scene 2}$} &
\includegraphics[width=.33\textwidth]{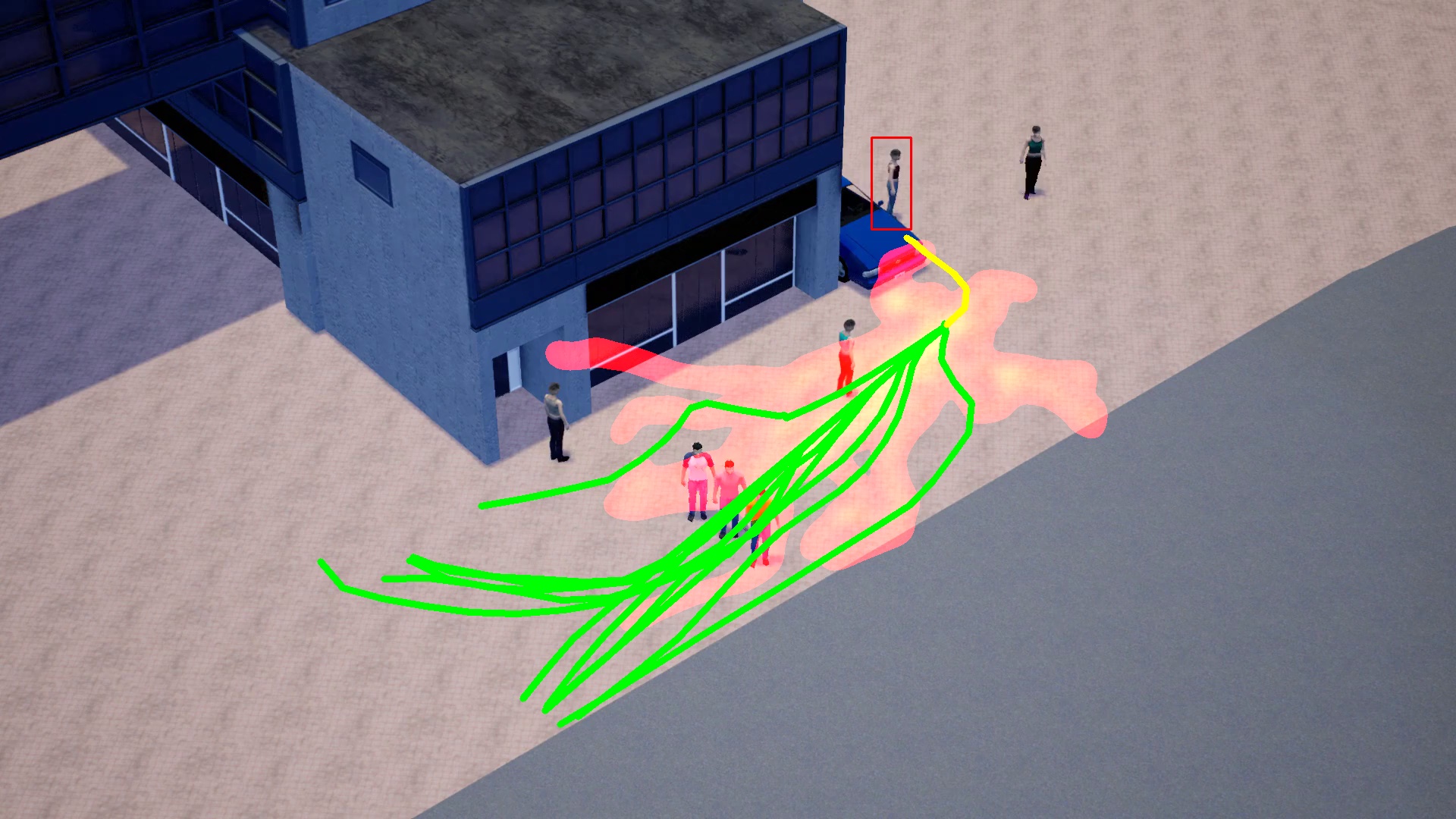}&
\includegraphics[width=.33\textwidth]{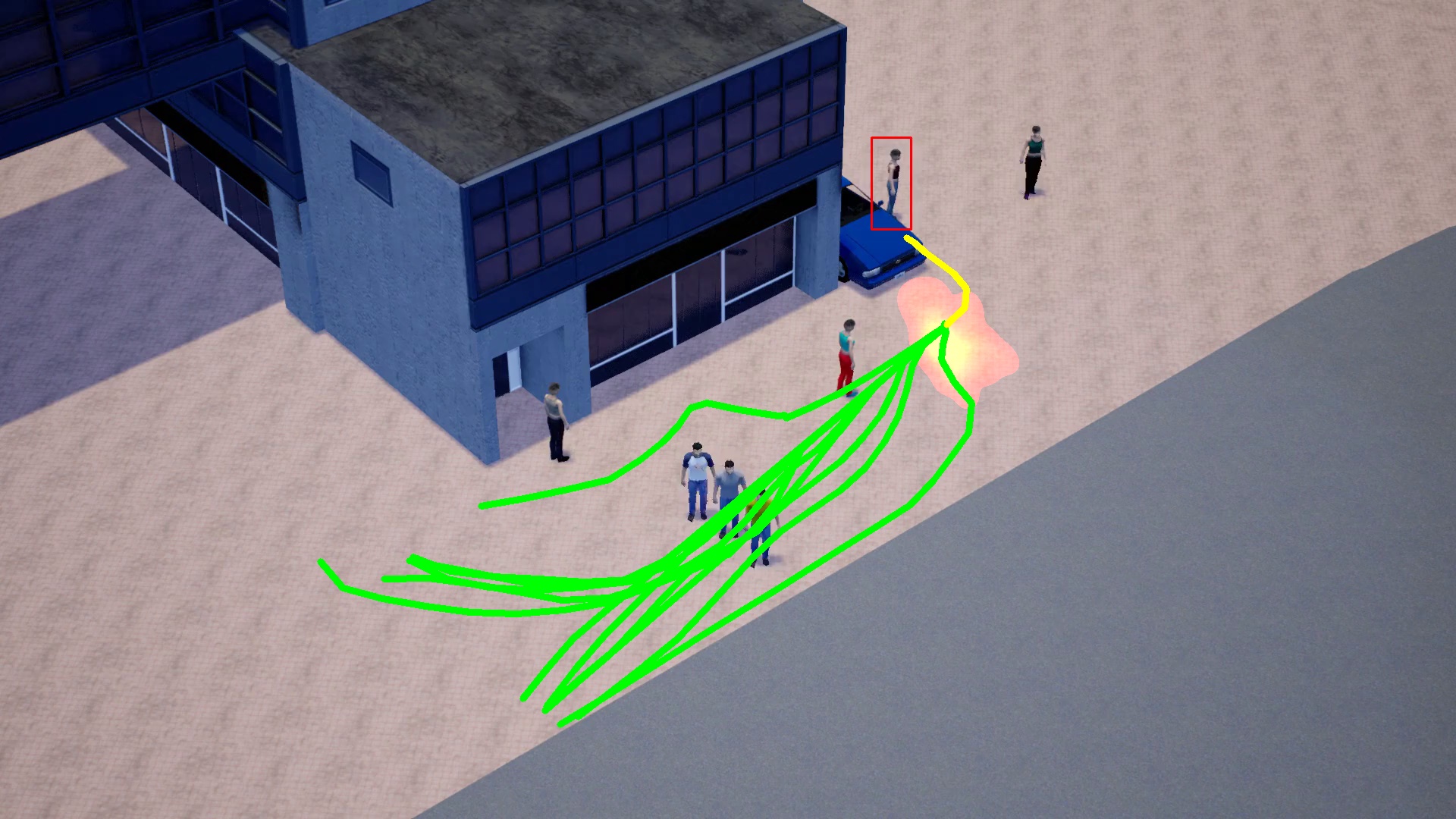}&
\includegraphics[width=.33\textwidth]{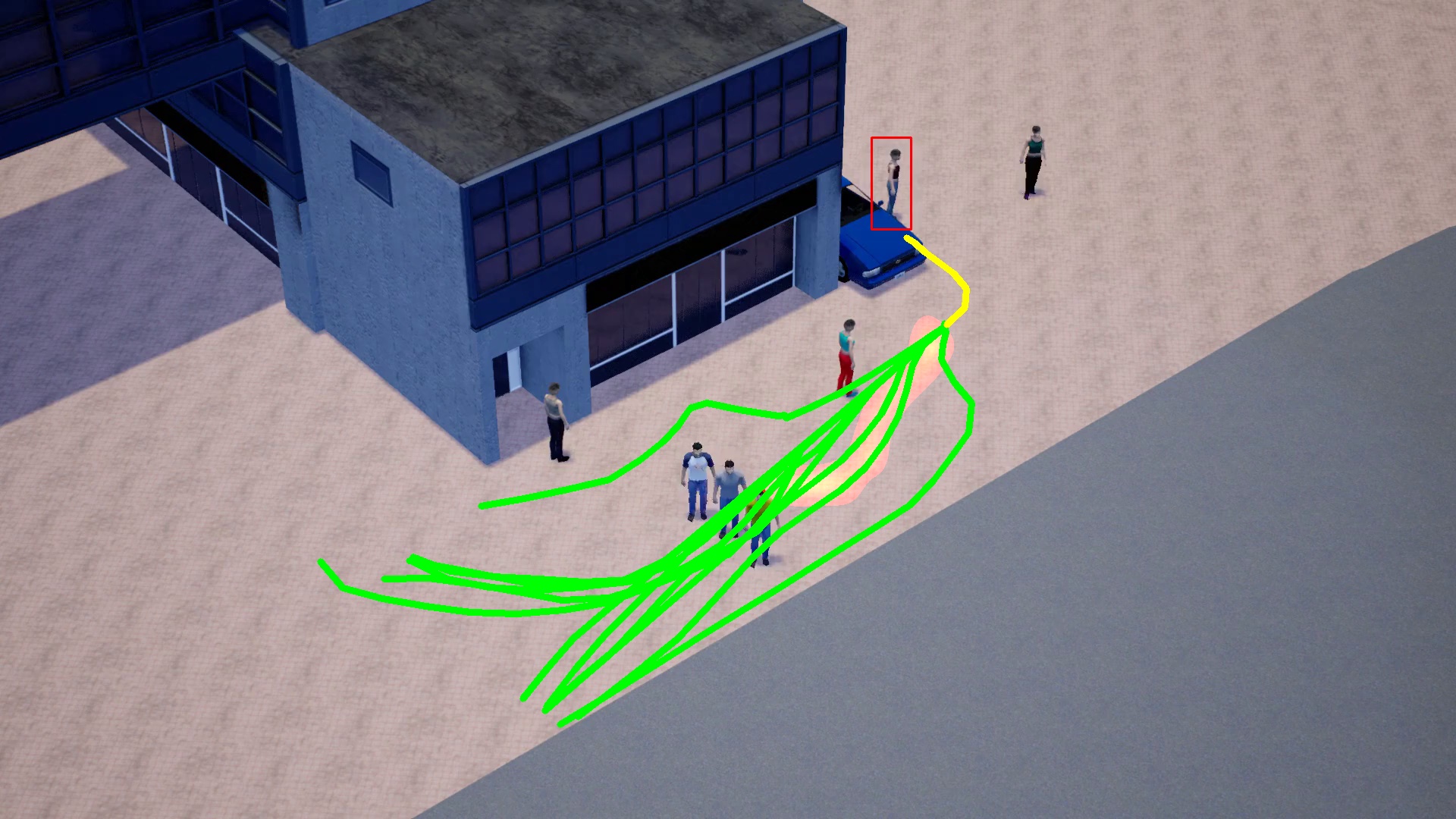}\\[-1cm]
\rotatebox[]{90}{$\quad \quad \quad \quad \quad \textbf{Scene 3}$} &
\includegraphics[width=.33\textwidth]{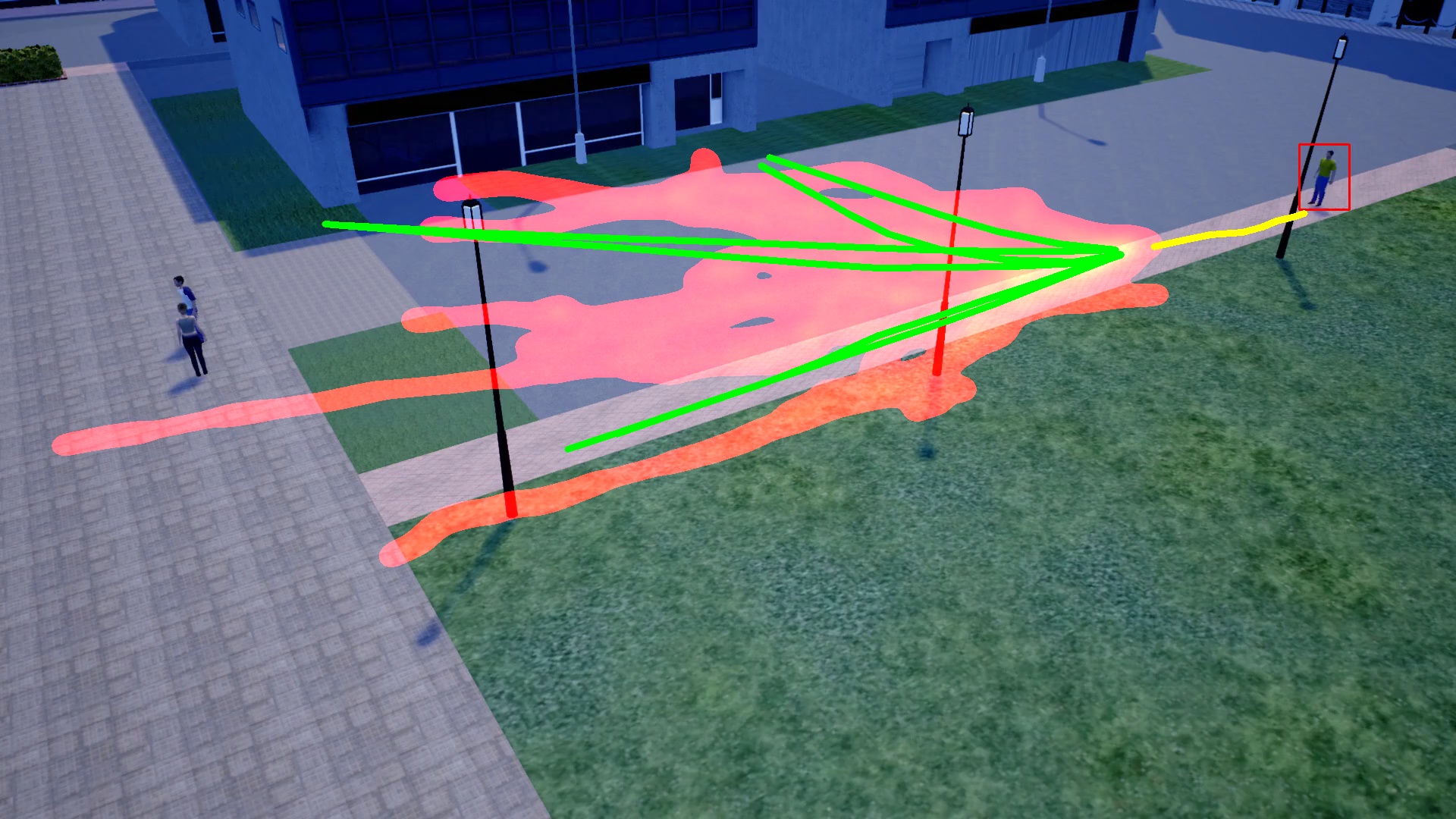}&
\includegraphics[width=.33\textwidth]{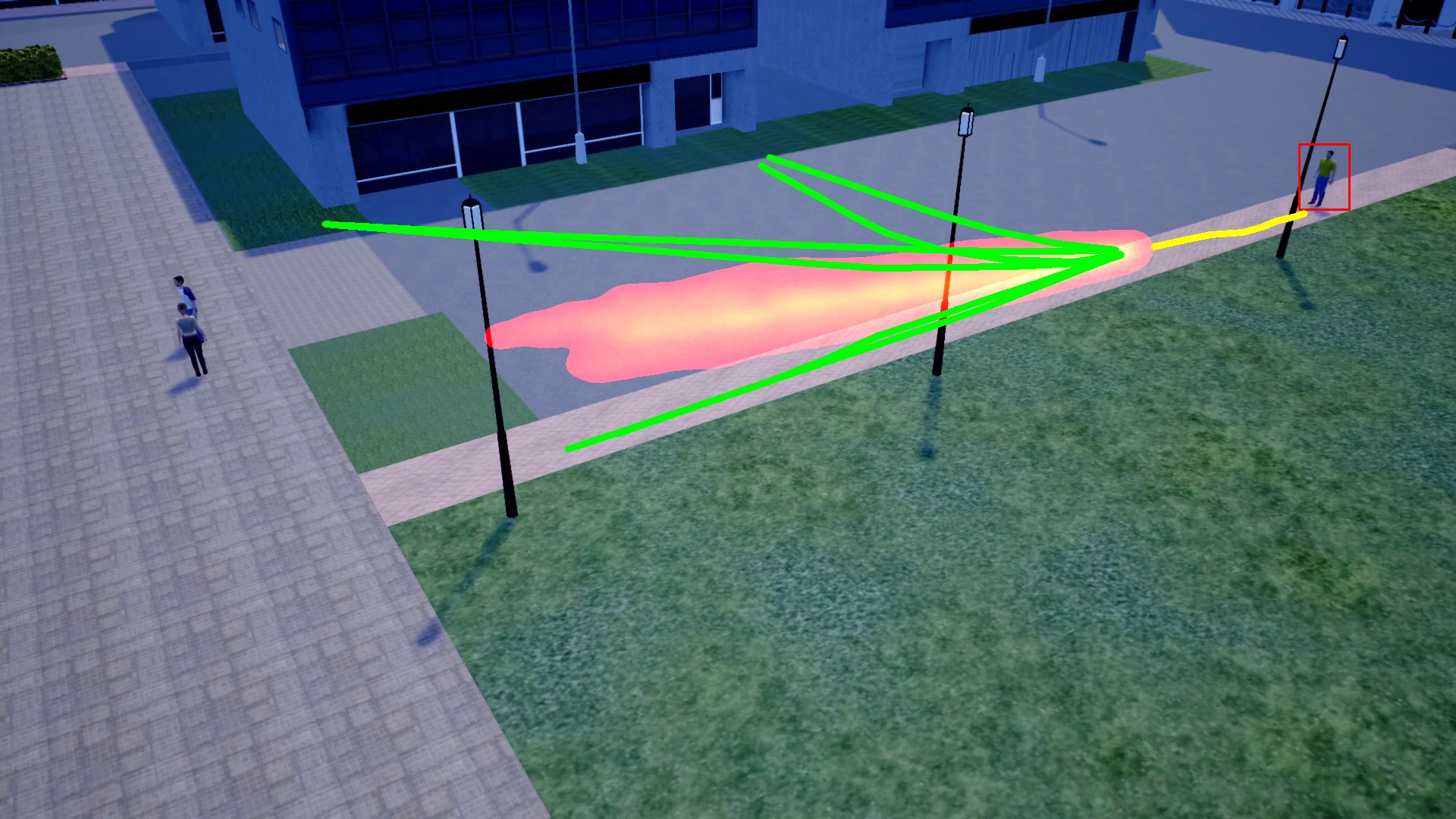}&
\includegraphics[width=.33\textwidth]{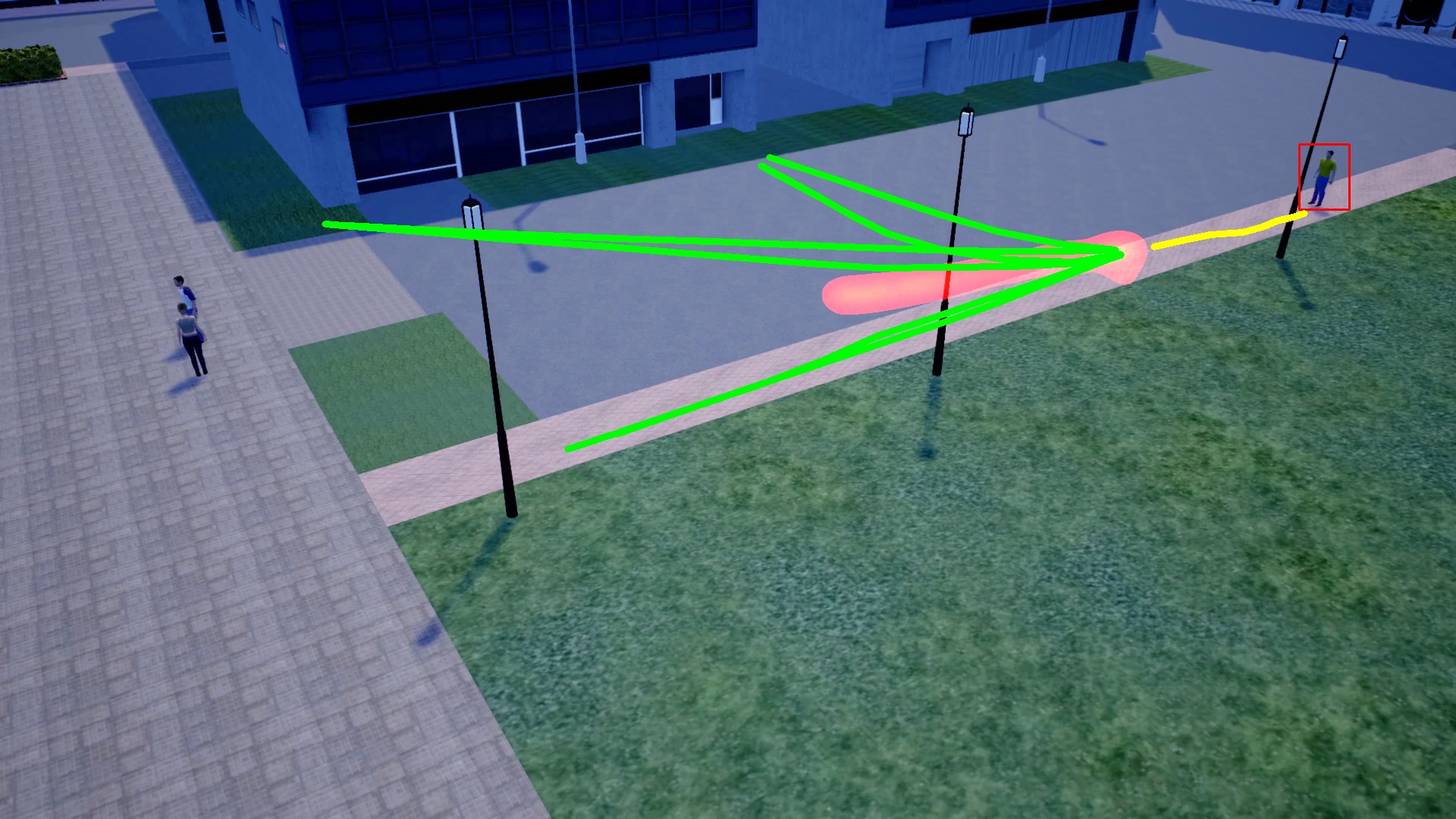}
\\[-1cm]
\rotatebox[]{90}{$\quad \quad \quad \quad \quad \textbf{Scene 4}$} &
\includegraphics[width=.33\textwidth]{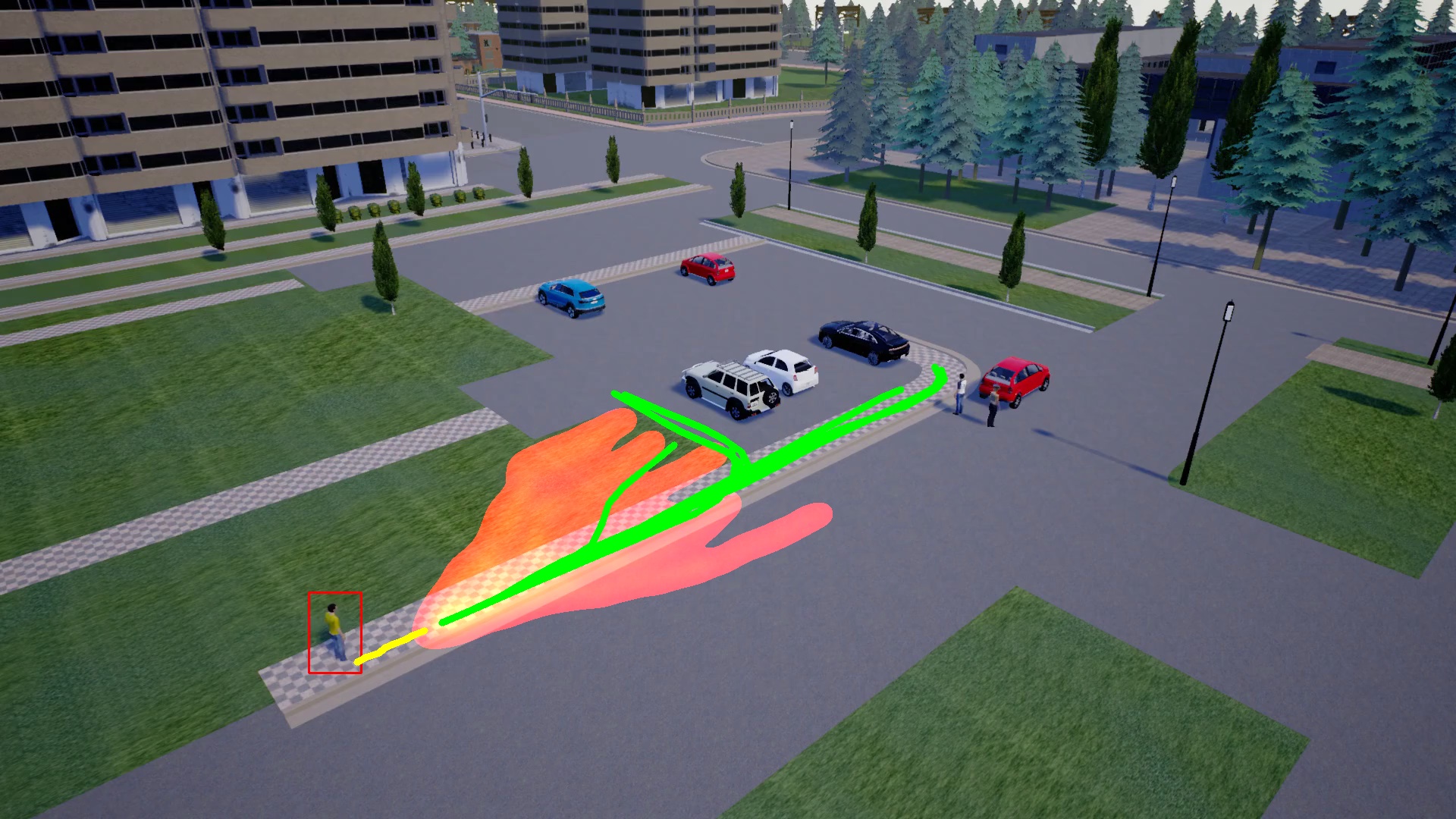}&
\includegraphics[width=.33\textwidth]{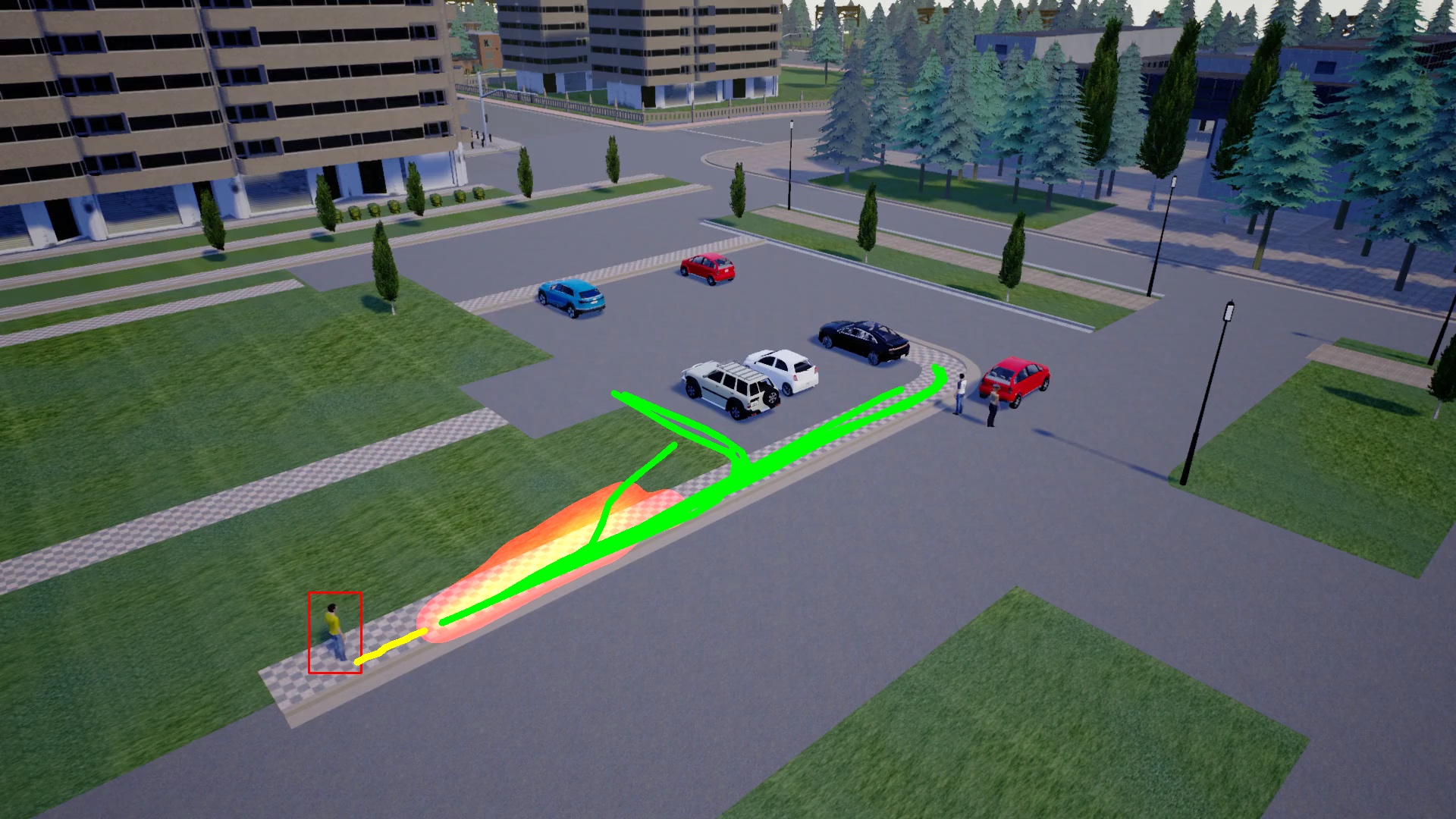}&
\includegraphics[width=.33\textwidth]{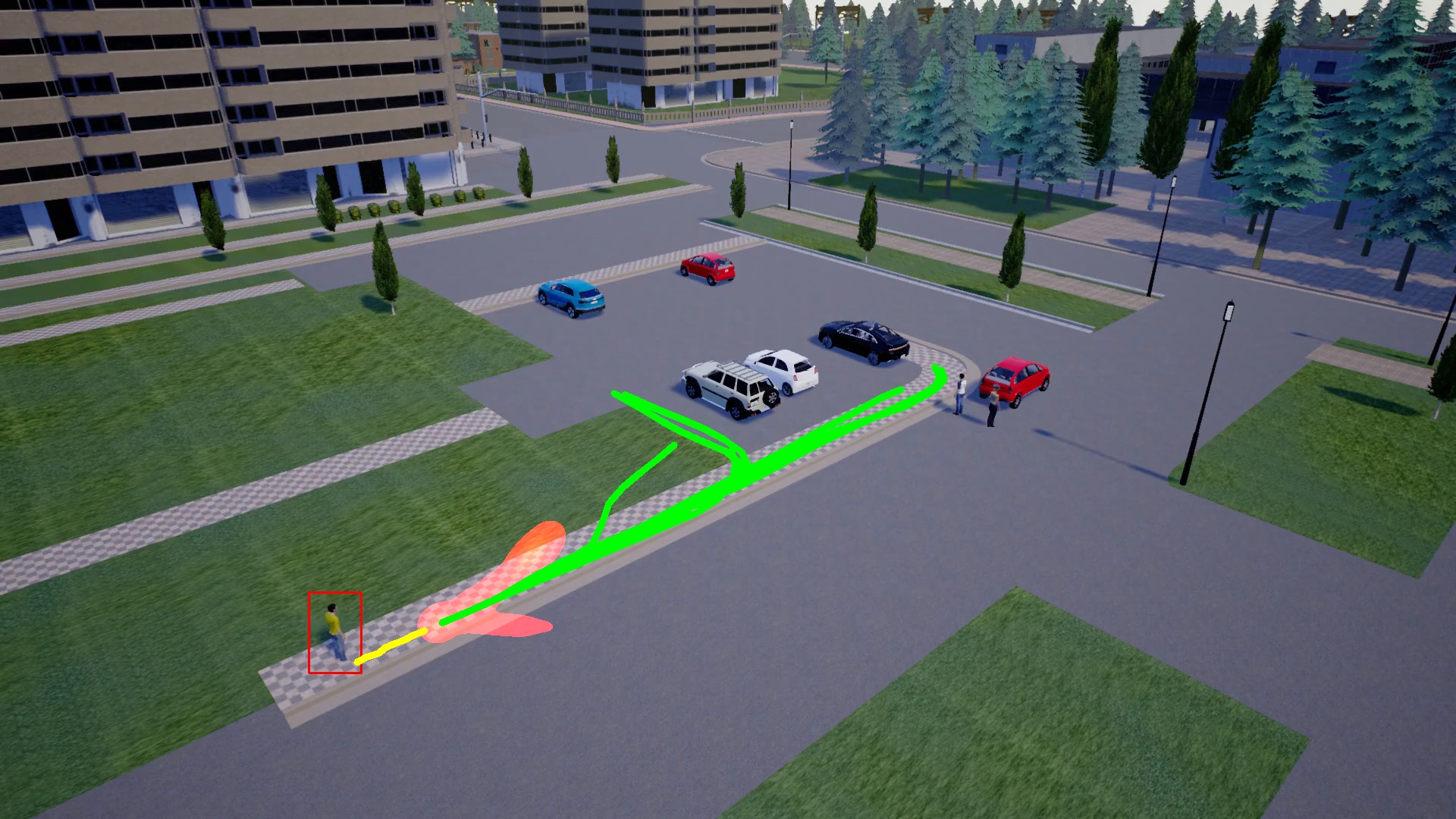}
\end{tabular}}
\caption{Additional model visualizations. The models from left to right: \textbf{LDS}, \textbf{CAM-NF}, and \textbf{Multiverse}.}
\label{fig:forking-path-visualization-additional}
\end{figure*}

In this section, we provide some additional visualizations of LDS, CAM-NF, and Multiverse outputs (Figure \ref{fig:forking-path-visualization-additional}). 

\end{document}